\documentclass[11pt, a4paper]{article}

\usepackage[pages=all, color=black, position={current page.south}, placement=bottom, scale=1, opacity=0, vshift=5mm]{background}

\usepackage[margin=1in]{geometry} 

\usepackage{amsmath}
\usepackage{amsthm}
\usepackage{amssymb}

\usepackage[utf8]{inputenc}
\usepackage{hyperref}
\hypersetup{
	unicode,
	pdfauthor={Manuel Sch\"urch, Dario Azzimonti, Alessio Benavoli, Marco Zaffalon},
	pdftitle={Correlated Product of Experts},
	pdfsubject={Correlated Product of Experts},
	pdfkeywords={Gaussian Process, non-parametric regression, prediction aggregation},
	pdfproducer={LaTeX},
	pdfcreator={pdflatex}
}

\theoremstyle{plain}
\newtheorem{theorem}{Theorem}

\newtheorem{proposition}[theorem]{Proposition}

\theoremstyle{definition}
\newtheorem{definition}[theorem]{Definition}

\usepackage{graphicx, color}
\graphicspath{{fig/}}

\usepackage{algorithm, algpseudocode} 
\usepackage{mathrsfs} 

\usepackage{lipsum}

\theoremstyle{thmstylethree}%
\newtheorem{proofII}[theorem]{Proof}%

\usepackage{wrapfig}
\usepackage{subfig}
\usepackage{rotating} 
\usepackage{lscape}

\usepackage{setspace}
\usepackage{etoolbox}

\makeatletter
\patchcmd{\@maketitle}{\LARGE}{\fontsize{30}{25}\selectfont}{}{}
\makeatother

\usepackage{booktabs}

\newcommand{\RR}{\mathbb{R}}
\newcommand{\II}{\mathbb{I}}
\newcommand{\bO}{\bs{0}} 

\newcommand{\pp}[1]{p\left(#1\right)}
\newcommand{\pc}[2]{\pp{#1\vert #2}}
\newcommand{\q}[1]{q\left(#1\right)}
\newcommand{\qc}[2]{\q{#1\vert #2}}

\newcommand{\bs}[1]{\boldsymbol{#1}}

\newcommand{\ba}{\bs{a}}
\newcommand{\bb}{\bs{b}}

\newcommand{\bx}{\bs{x}}
\newcommand{\by}{\bs{y}}
\newcommand{\bz}{\bs{z}}
\newcommand{\bff}{\bs{f}} 

\newcommand{\bv}{\bs{v}}
\newcommand{\bh}{\bs{h}}

\newcommand{\bthe}{\bs{\theta}}

\newcommand{\bmu}{\bs{\mu}}
\newcommand{\bSig}{\bs{\Sigma}}

\newcommand{\bga}{\bs{\gamma}}
\newcommand{\bnu}{\bs{\nu}}

\newcommand{\bA}{\bs{A}}
\newcommand{\bB}{\bs{B}}
\newcommand{\bX}{\bs{X}}

\newcommand{\bH}{\bs{H}}
\newcommand{\bV}{\bs{V}}
\newcommand{\bQ}{\bs{Q}}

\newcommand{\bP}{\bs{P}}
\newcommand{\bS}{\bs{S}}
\newcommand{\bK}{\bs{K}}
\newcommand{\bL}{\bs{L}}
\newcommand{\bD}{\bs{D}}
\newcommand{\bF}{\bs{F}}
\newcommand{\bT}{\bs{T}}
\newcommand{\bM}{\bs{M}}
\newcommand{\bW}{\bs{W}}
\newcommand{\bZ}{\bs{Z}}

\newcommand{\pr}{\bs{\pi}}
\newcommand{\prB}[1]{\pr \left(#1\right)}

\newcommand{\ps}{\bs{\psi}}
\newcommand{\psB}[1]{\ps \left(#1\right)}

\newcommand{\NNo}[1]{\mathcal{N}\left(#1\right)}
\newcommand{\NN}[2]{\mathcal{N}\left(#1\vert #2\right)}

\newcommand{\DIAG}[1]{\text{Diag}\left[#1\right]}

\newcommand{\sign}{\sigma_n^2}

\newcommand{\Kab}[2]{\bs{K}_{#1#2}}

\newcommand{\deT}[1]{\vert #1 \vert}

\newcommand*\diff{\mathop{}\!\mathrm{d}}

\usepackage{ulem}

\title{Correlated Product of Experts \\
for Sparse Gaussian Process Regression}
\author{Manuel Sch\"urch$^{1,2}$
\and Dario Azzimonti$^1$ 
\and Alessio Benavoli$^{3,1}$,
\and Marco Zaffalon$^1$}

\date{
	$^1$Istituto Dalle Molle di Studi sull’Intelligenza Artificiale (IDSIA), Lugano, Switzerland. \\ \texttt{\{manuel.schuerch, dario.azzimonti, marco.zaffalon\}@idsia.ch}\\%
	$^2$Università della Svizzera italiana (USI), Lugano, Switzerland. \\
	$^3$	University of Limerick (UL), Limerick, Ireland.\\ 
\texttt{alessio.benavoli@tcd.ie}\\[2ex]%
}

\begin{document}
\begin{Huge}
	\maketitle
	\end{Huge}
	
	\begin{abstract}
\noindent Gaussian processes (GPs) are an important tool in  machine
learning and statistics
with applications
 ranging from social and natural science through engineering.
 They constitute
a powerful kernelized non-parametric method with well-calibrated uncertainty estimates, however, off-the-shelf GP inference procedures are
limited to datasets with 
several
thousand data points because of their cubic
computational complexity. 
 For this reason, many sparse GPs techniques
have been developed over the past years. In this paper, we focus on GP regression tasks and propose a new approach based on aggregating predictions from several local and correlated experts.  
 Thereby, the degree of correlation between the experts 
 can vary between independent up to fully correlated experts.
 The individual predictions of the experts are 
 aggregated taking into account 
their correlation 
 resulting in 
 consistent
  uncertainty estimates.
  Our method recovers independent Product of Experts, sparse GP and full GP in the limiting cases.
  The presented framework can deal with a general kernel function 
and multiple variables, 
 and
 has a time and space complexity which is linear in the number of experts 
 and data samples,
  which makes our approach highly scalable.
We demonstrate superior performance, in a time vs. accuracy sense, of our proposed method against state-of-the-art GP approximation methods for synthetic as well as several real-world datasets with deterministic and stochastic optimization.
	\end{abstract}


\section{Introduction}

\textit{Gaussian processes} (GPs) 
 are
 a
class of 
powerful probabilistic method used in many statistical models 
due to their modelling flexibility, robustness to overfitting
and availability of well-calibrated predictive uncertainty estimates
with many applications
in
machine learning
and statistics.
However, off-the-shelf GP inference procedures are limited to datasets with a few thousand data points $N$, because of their 
computational  complexity $\mathcal{O}(N^3)$ and memory complexity $\mathcal{O}(N^2)$ 
 due to the inversion of a $N \times N$ kernel matrix
 \cite{ rasmussen2006gaussian}.
For this reason, many GP approximation techniques have been developed over the past years. 
There are at least two different approaches to circumvent the computational limitation of full GP.
 On the one hand, there are \textit{sparse and global} methods 
 \cite{csato2002sparse, quinonero2005unifying, rasmussen2006gaussian, seeger2003fast} based on  $M_g \ll N$ so-called (global) inducing points, 
 which 
 cover sparsely the input space and 
 optimally summarizing the dependencies of the training points. 
  This results in a low-rank approximation of the kernel matrix of size $M_g\times M_g$ which is less expensive to invert.
These methods consistently approximate full GP, for instance the authors in \cite{titsias2009variational} have shown that it converges to full GP as $M_g \rightarrow N$. However, all these methods are still cubic in the number of global inducing points $M_g$ and for many applications - in particular in higher dimensions - the amount of inducing points has to be rather large to capture the pattern of the function properly. A lot of work has been done to optimize the locations of the inducing inputs 
e.g.\ \cite{bui2016unifying, snelson2006sparse, titsias2009variational},
which allows to have less inducing points but more optimization parameters.
This optimization procedures were further improved by stochastic optimization   e.g.\ \cite{bui2017streaming, hensman2013gaussian, kaniasparse, schurch2020recursive}, which allows to update the parameters in mini-batches and thus speed up the inference.
Optimization of these (variational) parameters helps to scale GP approximations, 
however,
 the large number of optimization parameters makes these methods
  hard to train 
  and they are still limited to  $M_g$ global inducing points.

On the other hand, there are \textit{independent and local } models based on averaging predictions from $J$ independent local experts/models resulting in a block-diagonal approximation of the kernel matrix. The final probabilistic aggregation is then based on a product of the individual predictive densities, thus they are called \textit{Product of Experts (PoEs)}, see \cite{fleet2014generalized, deisenroth2015distributed,  hinton2002training, rulliere2018nested, tresp2000bayesian, liu2018generalized}.
%
%
%
PoE methods provide fast and rather accurate predictions,
 because they have fewer hyperparameters than inducing point methods and are locally exact. However, the predictive aggregation of complete independent experts leads  to unreliable uncertainty estimates and less accurate predictions in regions between experts.
 Further, also 
 a rigorous connection to full GP is missing.
%
Beside the mentioned local and global methods, there are  
also numerical approaches for the inversion
exploiting parallelism in specialized hardware \cite{wang2019exact}.
For a more thorough overview of GP approximations we refer to \cite{liu2020gaussian, rasmussen2006gaussian}.

%
%
%


\begin{figure}[htp!]
\centering
\subfloat[Precision Matrices.]{
    \includegraphics[width=0.55\linewidth]{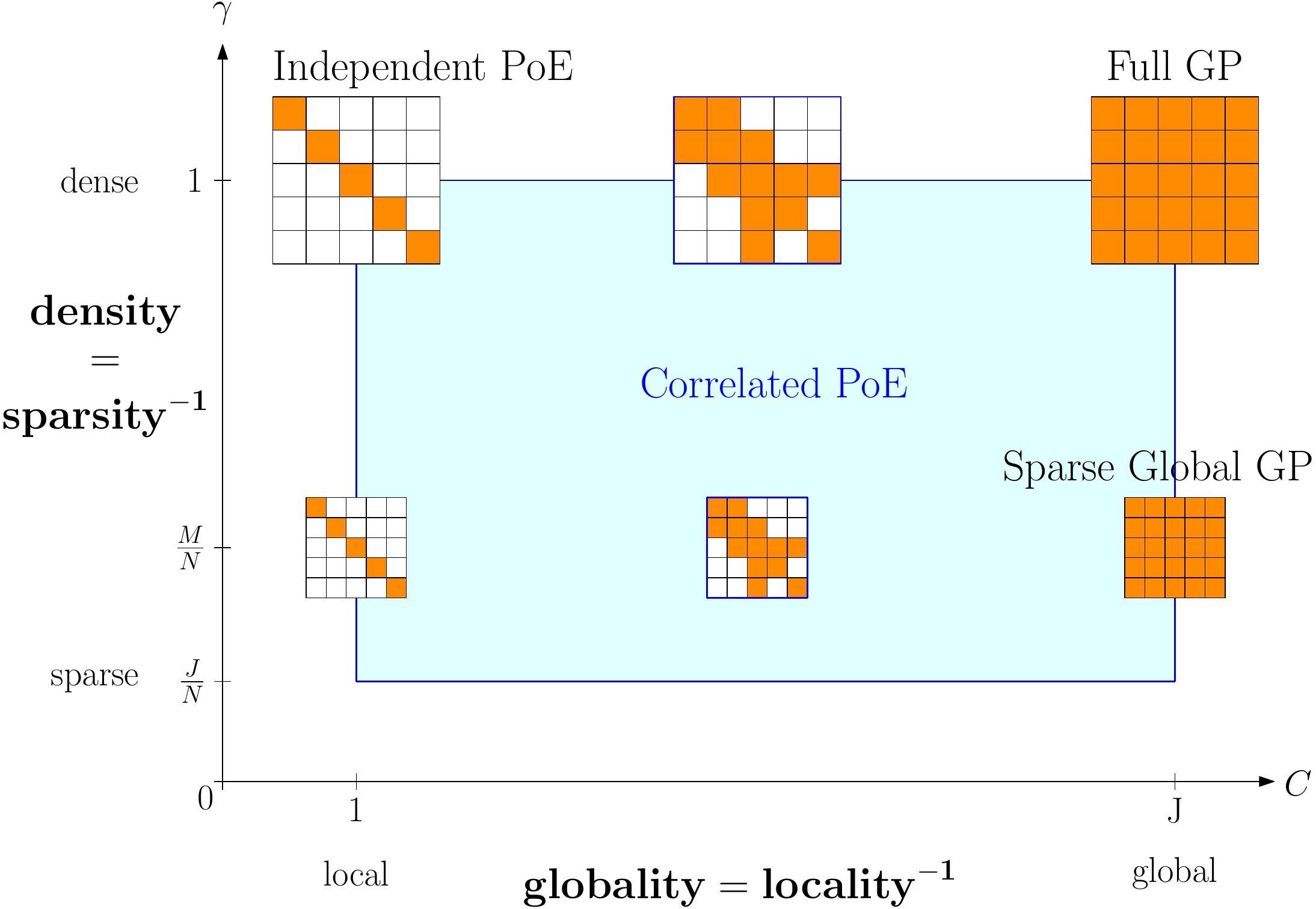}
}
\subfloat[KL to full GP.]{
    \includegraphics[width=0.4\linewidth]{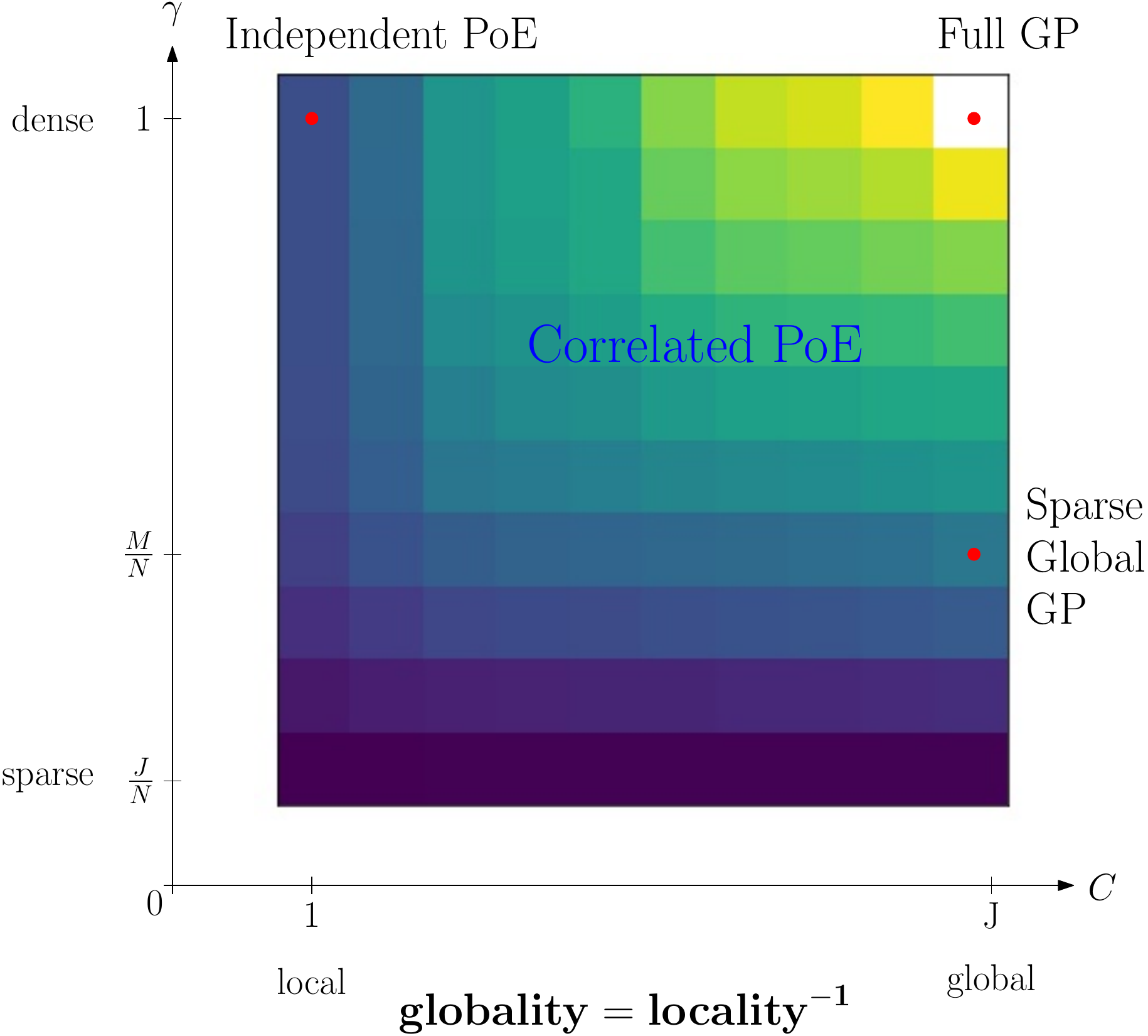}
}
\end{figure}
Our approach 
has the aim to overcome these  limitations by introducing  a framework based on $J$ correlated
experts 
so that it 
approximates full GP in two orthogonal directions: sparsity and locality.
Thereby, our model is
 a generalization of the independent PoEs 
 and sparse global GPs by  introducing local correlations between  
 experts. These experts  correspond to local and sparse 
 GP models represented by a  set of \textit{local inducing points}, which are points on the GP summarizing locally the dependencies of the training data. 
The degree of correlation  $C$ between the experts can vary between independent up to  fully correlated experts in a consistent way,
 so that our model recovers 
 independent PoEs, sparse global GP and
 full GP 
 in the limiting cases. 
%
%
%
%
%
%
Our method exploits the conditional independence between the experts resulting in a sparse and low-rank
prior as well as posterior
precision (inverse of covariance) matrix, which 
can be used to efficiently obtain local and correlated predictions from each expert.
These
correlated predictions are 
aggregated by the  covariance intersection method \cite{julier1997non}, 
which is useful for combining consistently several estimates with unknown correlations.
The resulting predictive distribution is a smooth weighted average of the predictive distributions of the individual experts.
Our algorithm works with a general kernel function 
and perform well with variables in higher dimensions.
The number of hyperparameters to optimize of our method is the same as for full GP, which are just a few parameters (depending on the kernel).
These parameters can be similarly estimated via the log marginal likelihood which is analytically and efficiently computable for our model. In our inference, also log normal priors can be incorporated leading to maximum-a-posteriori estimates for the hyperparameters.
%

Compared to the independent Product of Experts, the performance can already significantly improve by modelling just a few of the pairwise correlations between the experts.
 Compared to the number of \textit{global} inducing
point $M_g$ which is usual much smaller than the number of data points $N$, our approach allows a much
higher of total \textit{local} inducing points 
 in the order of $N$ 
which helps to cover the space and therefore model more complicated functions. 
Our method shares also some similarities 
with other sparse precision matrix GP approximations.
The works
\cite{durrande2019banded, grigorievskiy2017parallelizable}
 exploit a band precision matrix together with univariate kernels
whereas \cite{bui2014tree} propose a precision structure according to  a tree. 
 %
%
The authors
\cite{datta2016hierarchical, 
katzfuss2021general} use a more general precision matrix structure, however they need to know the prediction points in advance and are only well suited for low dimensional data (i.e. 1D and 2D)
which is usually not useful in the context of machine learning 
where the dimension is higher and predictions  are needed after training. 

%

In Section \ref{se:GP}, we briefly review full GP for regression and   \textit{sparse and global } as well as \textit{independent and local } approaches for GP approximation.
 In Section \ref{se:CPoE}, 
 we propose our method \textit{Correlated Product of Experts} (CPoEs) where
we introduce the  graphical model \eqref{se:graphicalModel} 
of our method and explain the local and sparse character  of the prior approximation \eqref{se:priorAprox}. 
Further, we discuss how to make inference \eqref{se:inference} and prediction \eqref{se:prediction} in our model.
In Section \ref{se:properties}, we show 
that the quality of our approximation
consistently improves in terms of Kullback-Leibler-(KL)-divergence \eqref{eq:KLgeneral} w.r.t.\ full GP for increasing degree of correlation.
Moreover, we present 
 deterministic and stochastic hyperparameter optimization techniques \eqref{se:pracDet}  and comparisons  \eqref{se:examples_evaluation}
 against state-of-the-art GP approximation methods in a time versus accuracy sense, for synthetic 
 as well as several real-world datasets.
We demonstrate superior performance of our proposed method   
 for different kernels in multiple dimensions.
Section \ref{se:conclusion} concludes the work and presents future research directions.

\section{GP Regression}
\label{se:GP}

Suppose we are given a training set 
$\mathcal{D} = \left\{ y_i, X_i \right\}_{i=1}^N$ 
of $N$ pairs of inputs $X_i\in \RR^D$ and noisy scalar outputs $y_i$ generated by adding independent Gaussian noise to a latent function $f$, that is $y_i = f(X_i)+\varepsilon_i$, where $\varepsilon_i\sim \NNo{0,\sigma_n^2}$. 
We denote $\by = [y_1,\ldots,y_N]^T$ the vector of observations and with $\bX = [X_1^T,\ldots,X_N^T]^T \in \RR^{N\times D}$.
We model $f$ with a \textit{Gaussian Process}, i.e.\ $f\sim $ GP($m, k_{\bthe}$)
with mean $m(X)$ and a covariance function (or kernel) $k_{\bthe}(X,X')$  for any $X,X'\in \RR^D$
where $\bthe$ is a set of  hyperparemeters.
  For the sake of simplicity, we assume
$m(X)\equiv 0$
  and  a \textit{squared exponential} (SE) kernel with individual lengthscales for each dimension if not otherwise stated, 
however, the mean function can be arbitrary and the covariance any   positive definite kernel function 
   (consider e.g.\ \cite{rasmussen2006gaussian}).
   For any input matrix  
   $\bA=[A_1;\ldots;A_M] \in\RR^{M\times D}$ consisting of rows $A_i\in\RR^D$, we define the GP output value
   $\ba = f\left(\bA\right)=\left[f(A_1),\ldots,f(A_{M})\right]^T
   =\left[a_1,\ldots,a_{M}\right]^T
   \in \RR^M$ so that the joint distribution 
   $\pp{\ba}=\pp{a_1,\ldots,a_{M}}$
   is Gaussian $
   \NN{\ba}{\bO, \bK_{\bA\bA}}$
   with a kernel matrix $\bK_{\bA\bA}\in \RR^{M\times M}$ where 
   the entries
   $\left[ \Kab{\bA}{\bA} \right]_{ij}
   =
   \Kab{A_i}{A_j} $
   correspond to the kernel evaluations
   $
   k_{\bthe}(A_i,A_j)\in\RR
   $.
    
\begin{figure}[htp!]
\centering
\subfloat[Training.]{
    \includegraphics[width=0.55\linewidth]{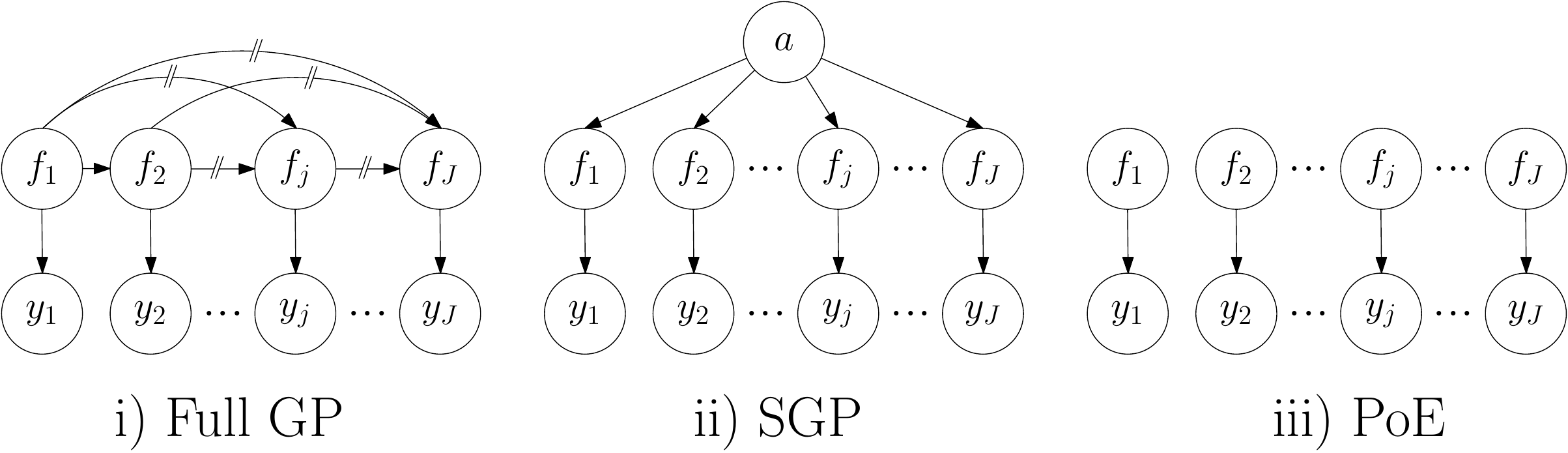}
}
~~~
\subfloat[Prediction.]{
    \includegraphics[width=0.4\linewidth]{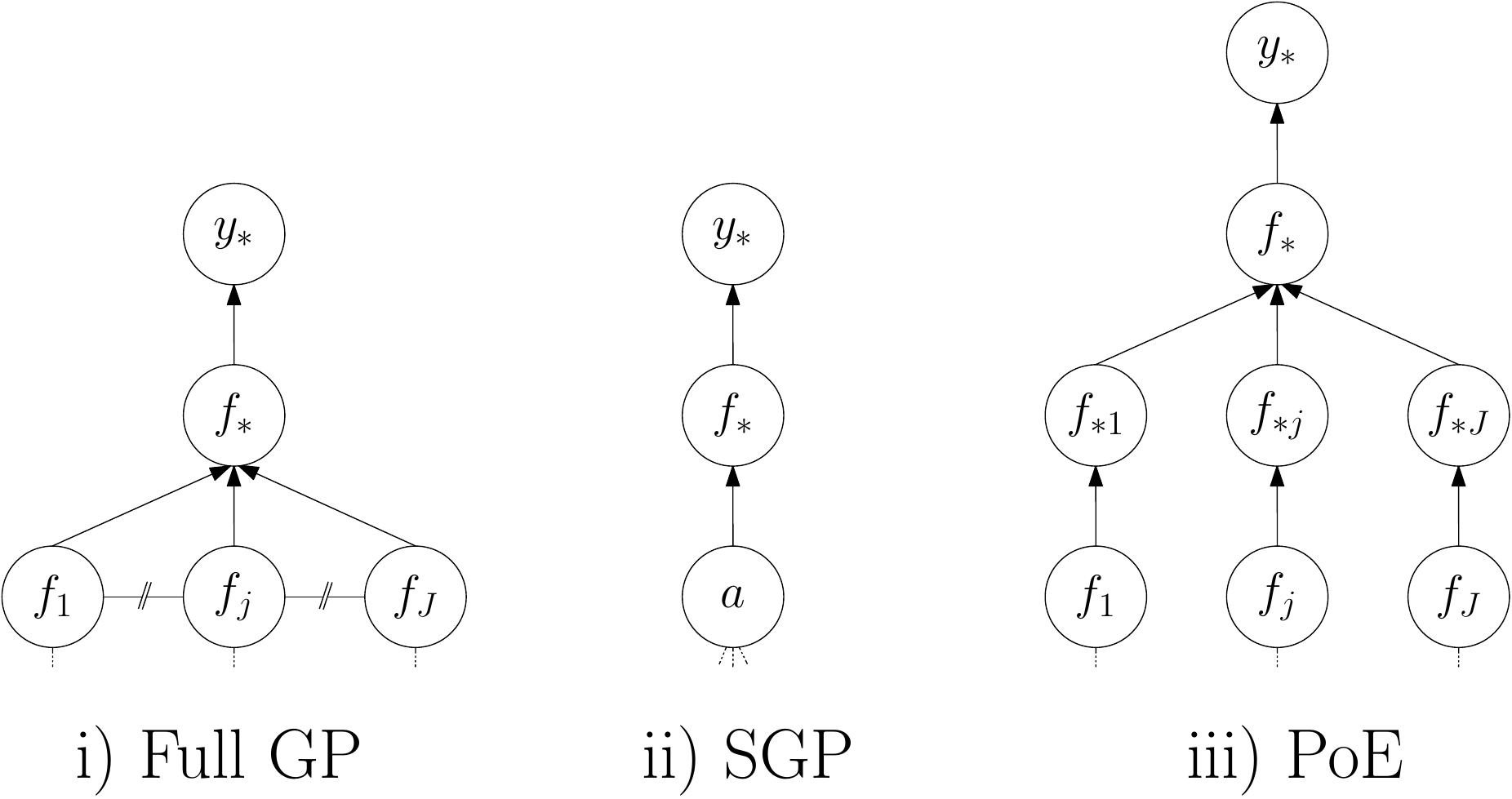}
}
\caption{ Graphical models of different GP approaches.}
\label{fig:GPmods}
\end{figure}

\noindent In particular, the joint distribution $p(\bff, f_*)$ 
of the training 
	values 
	$\bff = f\left(\bX\right)=\left[f(X_1),\ldots,
	f(X_N)\right]^T$
	and a test function value $f_*=f(X_*)$ at test point $X_*\in \RR^D$ is Gaussian
	$\NNo
{\bO,\Kab{
[\bX ;
X_*]
}{
[\bX ;
X_*]
}
}$ where $[\bX ;
X_*]$ is the resulting matrix when stacking the matrices above each other.
For GP regression, the  Gaussian likelihood $\pc{\by}{\bff} = \NN{\by}{\bff, \sign \II}$ can be combined with the joint prior  $p(\bff, f_*)$ so that 
the predictive posterior 
	distribution
		can be analytically derived  \cite{rasmussen2006gaussian}.
%
%
		%
		%
		%
	%
	%
%
Alternatively,
we present a two stage procedure to highlight later connections to our model.
The posterior distribution over the latent variables 
given the data 
can be explicitly formulated as
\begin{align}
 \label{eq:postFull}
\pc{\bff}{\by}
\propto
\pp{\bff,\by}
=
\prod_{j=1}^J
\pc{\by_{j}}{\bff_{j}}
\pc{\bff_{j}}{\bff_{1:j-1}},
\end{align}
where the data is split into $J$ mini-batches of size $B$, i.e.
$\mathcal{D} = \left\{ \by_j, \bX_j \right\}_{j=1}^J$ 
with inputs $\bX_j\in \RR^{B\times D}$, outputs $\by_j\in \RR^B$ and the corresponding latent function values $\bff_j = f(\bX_j)\in\RR^B$. 
In  \eqref{eq:postFull}  we used the notation $\bff_{k:j}$ indicating $[\bff_k,\ldots,\bff_j]$ and the conditionals
$\pc{\bff_{j}}{\bff_{1:j-1}}$ can be derived from the joint Gaussian.
%
Given the posterior $p(\bff\vert\by)$, the predictive posterior distribution 
from above
is equivalently obtained as
$
\pc{f_*}{\by}
=
\int \pc{f_*}{\bff} \pc{\bff}{\by} \diff \bff
$ via Gaussian integration
\eqref{eq:int}.
%
The corresponding graphical model
is depicted in Fig.\ \ref{fig:GPmods}(a)i) and \ref{fig:GPmods}(b)i), respectively.

%
%

	The GP depends via the kernel matrix on the 
	hyperparameters $\bthe$,
	which are typically estimated by maximizing the log marginal likelihood
	%
	$
	\log \pc{\by}{\bthe} 
	= 
	\log \NN{\by}{\bO, \Kab{\bX}{\bX} + \sign \II}.$ 
%
	Although GP inference is an elegant probabilistic approach for regression, the computations for inference and parameter optimization require the inversion of the  matrix 
	$\bK_{\bX\bX} + \sigma_n^2 \II\in\RR^{N\times N}$
	which scales as 
	$\mathcal{O}(N^3)$ in time and $\mathcal{O}(N^2)$ for memory which is infeasible for large $N$.

		\subsection{Global Sparse GPs}
	\label{se:sparse_GP}
		Sparse GP regression approximations 
		based on \textit{global inducing points} reduce the computational complexity by introducing  
	$M_g \ll N$ inducing points $\ba \in \RR^{M_g}$ 
	that optimally summarize the dependency of the whole training data globally as illustrated in  the graphical model  in  Fig.\ \ref{fig:GPmods}(a)ii)
	and
		 is denoted in the following as SGP$(M_g)$.
	Thereby the  inducing \textit{inputs} $\bA \in \RR^{M_g\times D}$ are in the $D$-dimensional input data space and the inducing \textit{outputs} $\ba = f(\bA)\in \RR^{M_g}$ are the corresponding GP-function values. 
	%
	%
Similarly to full GP in Eq.  \eqref{eq:postFull}, the posterior over the inducing points $p(\ba\vert\by) \propto \int \pp{\ba,\bff,\by} \diff \bff$ can be derived from the joint distribution
\begin{align}
 \label{eq:sparse}
\pp{\ba,\bff,\by}
=
\prod_{j=1}^J
\pc{\by_{j}}{\bff_{j}}
\pc{\bff_{j}}{\ba}
p(\ba),
\end{align}
where the usual Gaussian likelihood 
$\pc{\by_{j}}{\bff_{j}}=\NNo{\bff_{j},\sigma_n^2\II}$
%
 and the Gaussian conditional $\pc{\bff_{j}}{\ba}$ 
 are used.
	Based on the  joint distribution 
	in \eqref{eq:sparse}, 
	the posterior $p(\ba\vert \by)$ can be derived from which  
	prediction can be performed using the predictive conditional $\pc{f_*}{\ba}$ 
as  more precisely explained in 
	\ref{se:sparse_GP2} in the Appendix and illustrated in 
	Figure \ref{fig:GPmods}(b)ii).
	Batch inference in these sparse global models can be done in $\mathcal{O}(M_g^2N)$ time and $\mathcal{O}(M_g N)$ space (\cite{quinonero2005unifying}). 
	\\		
	In order to find optimal inducing inputs $\bA$ and hyperparameters $\bthe$, a sparse variation of the log marginal likelihood similar to full GP
	 can be used 
	 \cite{bui2016unifying, snelson2006sparse, titsias2009variational}.
	For larger datasets, stochastic optimization  has been applied e.g.\ \cite{bui2017streaming, hensman2013gaussian, kaniasparse, schurch2020recursive} to obtain faster and more data efficient optimization procedures.
	  For  recent reviews on the subject we refer to 
	 \cite{liu2020gaussian,quinonero2005unifying,rasmussen2006gaussian}.

%
%

\subsection{Local Independent GPs}\label{se:localGPR}
\begin{figure}[htp!]
	\centering
  \includegraphics[width=.9\textwidth]{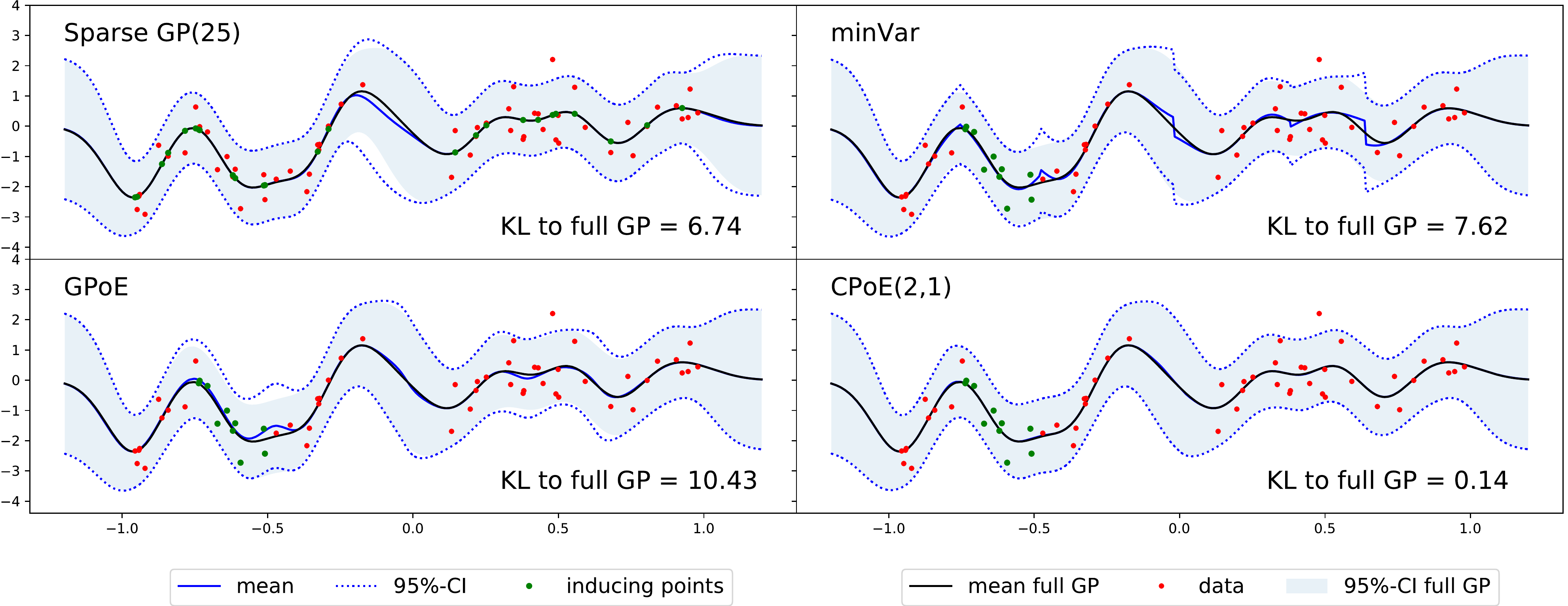}
%
	\caption{
Different GP approximations 
(with comparable time complexity)
indicated with  predictive mean (solid blue) and $95\%$-credible interval (dotted blue) compared to full GP (black and shaded blue area). 
The number in the right bottom corner indicates the KL-divergence \eqref{eq:KLgeneral} to full GP.
 In the bottom plot, our method \textit{Correlated Prodcuct of Expert (CPoE)} is presented for a degree of correlation $C=2$ and sparsity $\gamma=1$.
 }
	\label{fig:toy1D}
\end{figure}

Local approaches constitute an alternative to 
 global sparse inducing point methods,
which exploit multiple local GPs combined with averaging techniques to perform predictions. In this work we focus on 
 \textit{Product of Expert (PoE)} 
 \cite{hinton2002training},
where individual predictions 
from $J$ experts based on the local data $\by_j$
are aggregated to the final predictive distribution
\begin{align}
\label{eq:PoEagg}
\pc{f_*}{\by}
=
\prod_{j=1}^{J}
g_j\left(
\pc{f_{*j}}{\by_{j}}
\right)
\end{align}
where
$g_j$ is a function 
introduced in order to
 increase or decrease the importance of the experts 
 and 
 depends on the particular PoE method 
\cite{hinton2002training, fleet2014generalized, tresp2000bayesian, liu2018generalized, liu2020gaussian}.
Note, in particular, the \textit{generalized PoE (GPoE)} \cite{fleet2014generalized}, where the weights are set to the difference in entropy of the local prior and posterior.
%
The individual predictions $\pc{f_{*j}}{\by_{j}}$ are based on a local GP
%
for which the implicit 
joint 
posterior   can be formulated as
\begin{align}
 \label{eq:PoE}
\pc{\bff}{\by}
\propto
\pp{\bff,\by}
=
\prod_{j=1}^J
\pc{\by_{j}}{\bff_{j}}
\pp{\bff_{j}},
\end{align}
where the corresponding graphical model is depicted in Figure \ref{fig:GPmods}iii)
and more details are provided in Appendix \ref{se:localGPR2}.
Other important contributions in this field are 
distributed local GPs \cite{deisenroth2015distributed} and local experts with consistent aggregations 
 \cite{rulliere2018nested, nakai2021nested}.
Simple baseline methods for local methods are the \textit{minimal variance (minVar)} and the
\textit{nearest expert (NE)} aggregation, where only the prediction from the expert with minimal variance and nearest expert is used, respectively. 
Although both  methods show often surprisingly good performance, they suffer from the important disadvantage that there are serious discontinuities at the boundaries between the experts (see for instance Fig.\ 	\ref{fig:toy1D}) and thus often not useful in practice. This is also the main limitation of all local methods  based only on the prediction of one expert  which was the main  reason for introducing smooth PoEs with combined experts.
We 
refer to \cite{liu2020gaussian} for a recent overview.

%
%
%
%
%

\section{Correlated Product of Experts}
\label{se:CPoE}
In this section we present our GP regression method \textit{Correlated Product of Expert} CPoE$(C,\gamma)$ which is 
 a generalization of the independent PoEs and sparse global GPs.
The first generalization is the introduction of correlations between the experts which can be adjusted by the parameter
$1\leq C\leq J$
 and allows to interpolate between local and global models.
Secondly, similar to the sparse global approximation, our method allows to sparsify the inducing points by sparsity parameter $0<\gamma\leq 1$.
We refer to Table \ref{tab:notation} in the Appendix for an overview of the used notation.
%
%
%

\subsection{Graphical Model}
\label{se:graphicalModel}
Assuming $N=BJ$ data samples 
which are divided into $J$ ordered partitions (or experts) of size $B$, i.e.
$\mathcal{D} = \left\{ \by_j, \bX_j \right\}_{j=1}^J$ 
with inputs $\bX_j\in \RR^{B\times D}$ and outputs $\by_j\in \RR^B$.
We denote  $\bff_j = f(\bX_j)\in\RR^B$ the corresponding latent function values on the GP $f$.
We abbreviate $\by=\by_{1:J}\in\RR^{N},\bX=\bX_{1:J}\in \RR^{N\times D}$ and $\bff=\bff_{1:J}\in \RR^{N}$.

\begin{definition}[Local Inducing Points]
\label{def:localIndPoints}
We refer to local inducing points 
$\left\{ \ba_j, \bA_j\right\}_{j=1}^J$
 with inducing inputs $\bA_j \in\RR^{L\times D}$ and  the corresponding inducing outputs
$\ba_j = f(\bA_j)\in \RR^{L}$
of size $L=\lfloor \gamma B \rfloor$ with $0<\gamma\leq 1$.
\end{definition}

These $L$ local inducing points $\left( \ba_j, \bA_j\right)$ of expert $j$ serve as local summary points for the data 
$\left( \by_j, \bX_j\right)$ 
where the sparsity level can be adjusted by $\gamma$. 
 If $\gamma=1$, the inducing inputs $\bA_j$ correspond exactly to $\bX_j$ and correspondingly $\ba_j = \bff_j$.
We abbreviate $\ba = \ba_{1:J} \in \RR^{M}$ with $M=LJ$ for all local inducing outputs with the corresponding local inducing inputs 
$\bA = \bA_{1:J} \in \RR^{M\times D}$.
%
%

Next, we model connections between the
experts by a set of neighbour experts according to the given ordering.
%
\begin{definition}[Predecessor and Correlation Index Sets]
\label{def:predSets}
Let $\phi_{i}(j)\in \{1,\ldots,j-1\}$ the index of the $i$th predecessor of the $j$th expert.
For a given correlation parameter $1\leq C\leq J$, we introduce 
the \textit{predecessor set}
$\bs{\pi}_C(j)=\bigcup_{i=1}^{I_j} \phi_{i}(j)$ 
satisfying
\begin{align*}
\bs{\pi}_C(j) \subset \{1,\ldots,j-1\}
\quad\quad\text{and} \quad\quad
\bs{\pi}_{C+1}(j) = \bs{\pi}_C(j) ~\cup~ \phi_{C+1}(j)
\end{align*}
such that the size of the set
 $I_j = \vert \bs{\pi}_C(j) \vert = \min(j-1,C-1) $.
 Further,
 we define the region of correlation with the \textit{correlation indices}
\begin{align*}
\bs{\psi}_C(j)
=\begin{cases}
               \bs{\pi}_C(j) ~\cup ~\{j,\ldots,C\},\quad &\text{if} \quad j < C \\
               \bs{\pi}_C(j) ~\cup ~j , \quad &\text{if} \quad j \geq C
            \end{cases}
  \end{align*}  
such that 
$\vert \bs{\psi}_C(j)\vert = C$
and  $\bs{\psi}_C(j) = \bs{\psi}_C(C) = \{1,\ldots,C\}$ for all $j\leq C$.
\end{definition}
%
%
%
 %
 %
 The purpose of these predecessor and  correlation indices are to model the local  correlations  among the experts of degree $C$.
 If for all $j$ the indices $\bs{\pi}_C(j)$ are the $C-1$ previous indices,
we say that the predecessors are \textit{consecutive} and \textit{non-consecutive} otherwise.
%
 If $C$ is clear from the context, $\bs{\pi}_C(j)$ and $\bs{\psi}_C(j)$ are abbreviated by $\bs{\pi}(j)$ and $\bs{\psi}(j)$, respectively.
Details about the specific choices of the ordering,  partition, inducing points and predecessor indices are given in Section \ref{se:graphDetails}.
%
%
%

\begin{figure}[htp!]
	\centering
  \includegraphics[width=.99\textwidth]{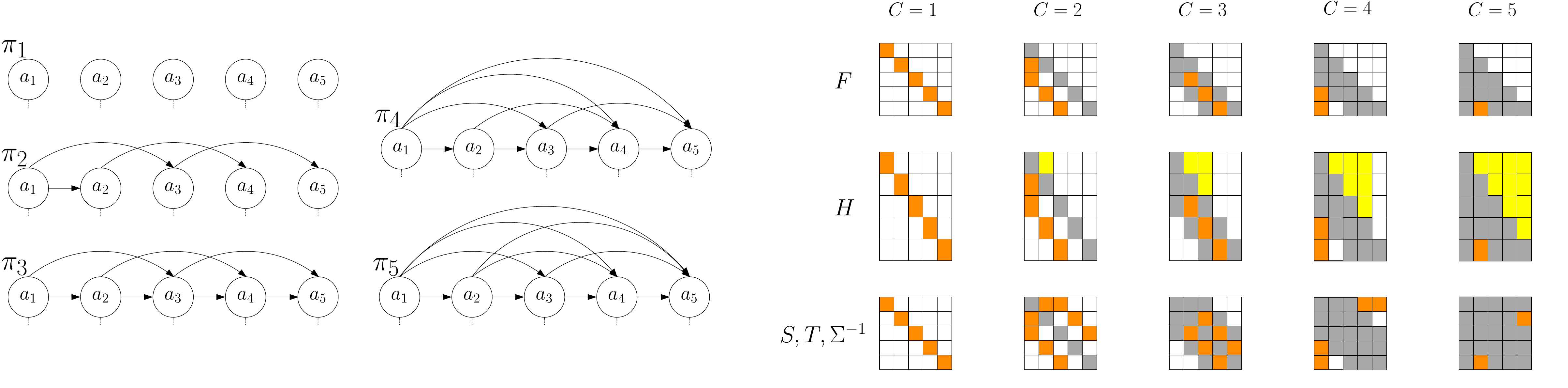}
	\caption{ 
Correlation structure $\bs{\pi}_C$ between the $J=5$ experts for different
degrees of correlation
	 $1\leq C\leq J$. Left: Graphical model among the local inducing points $\ba_j$.
	 Right: 
	  Structure of sparse transition matrix $\bF$, projection matrix $\bH$, prior precision $\bS$, likelihood precision $\bT$ and posterior precision $\bSig^{-1}$.
	  Note that $\bs{\pi}_C$ does not have to be consecutive, e.g $2 \notin \bs{\pi}_2(3)$.
%
%
}
	\label{fig:GPmods4}
\end{figure}

\begin{definition}[Graph]
\label{def:graph}
We define  a directed graph 
$\mathcal{G}(V,E)$ 
with nodes 
$V=
\ba \cup
\bff \cup
\by
%
$
and directed edges 
\begin{align*}
E=
\{~
&\{
(\ba_{\bs{\pi}_C^i(j)} ,\ba_j)
\}_{i=1}^
{I_j}
~\cup~
\{
(\ba_{\bs{\psi}_C^i(j)} ,\bff_j)
\}_{i=1}^
{C}
~\cup ~
(\bff_j,\by_j)
~
\}_{j=1}^J,
\end{align*}
where  
$\bs{\pi}_C^i(j)$
and $\bs{\psi}_C^i(j)$
denote the $i$th element in the corresponding set.
\end{definition}

\begin{figure}[htp!]
\centering
\subfloat[Training.]{
    \includegraphics[width=0.61\linewidth]{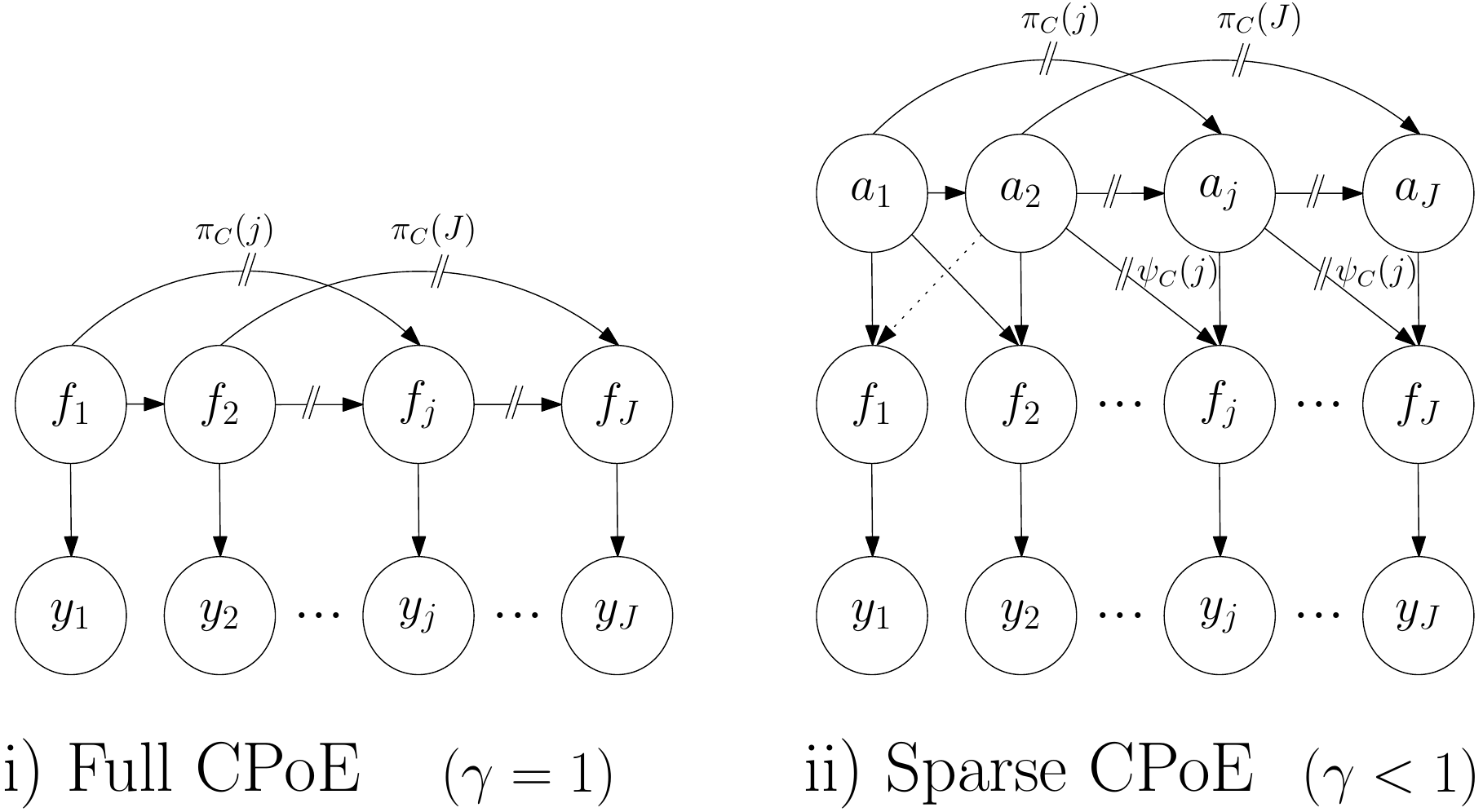}
    \label{fig:GPmods2}
}
~~~
\subfloat[Prediction.]{
    \includegraphics[width=0.31\linewidth]{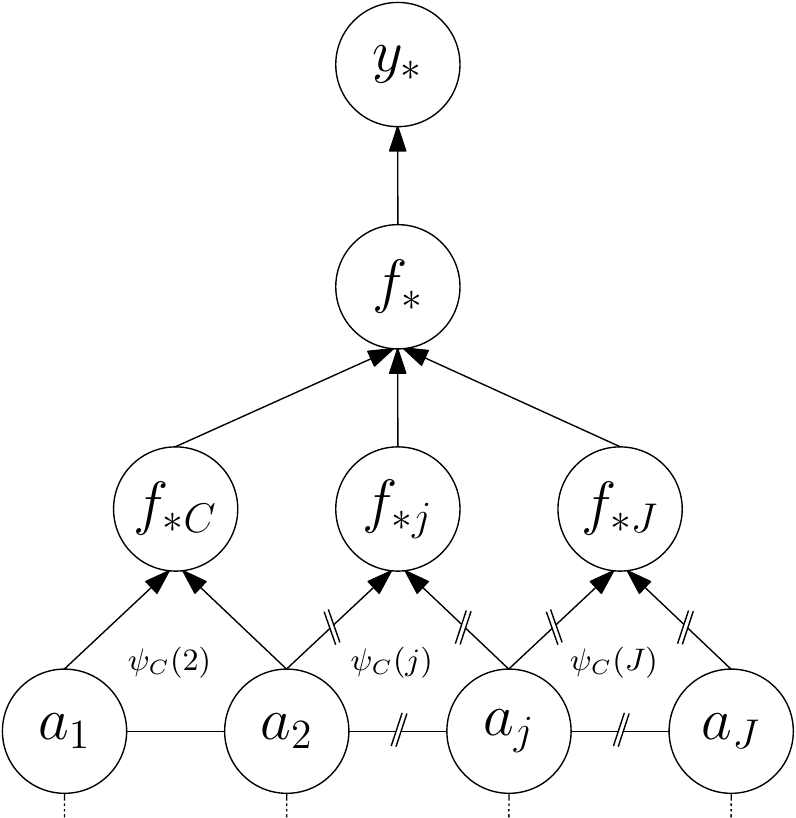}
     \label{fig:pred2}
}
\caption{Graphical model for training and prediction  of CPoE($C,\gamma$). }
\end{figure}

The directed graph  $\mathcal{G}$ is depicted in
Fig.\ \ref{fig:GPmods2}ii)
 where the local inducing points of the $j$th
expert
are connected
 with the 
 inducing points of the $I_j$ experts in $\bs{\pi}_C(j)$. Further, the function values $\bff_j$ are connected in the region of correlation $\bs{\psi}_C(j)$ to the local inducing points. 
%
%
%
%
%
The graph $\mathcal{G}=(V,E)$ can be equipped with a probabilistic interpretation, in particular, each node $\bv\in V$ and each incoming edge $(\bv_i,\bv)\in E$ for all
 predecessors  $i=1,\ldots,I$ 
can be interpreted as a conditional probability density
$\pc{\bv}{\bv_1,\ldots,\bv_{I}}$.
%
%

\newpage

\begin{proposition}[Graphical Model; Proof \ref{proof:graphicalModel}]
\label{prop:graphicalModel}
We define a graphical model corresponding to the graph
$\mathcal{G}(V,E)$ with the conditional probability distributions
\begin{align}
\label{eq:likelihood}
\pc{\by_j}{\bff_{j}} 
&=
\NN{\by_j}{\bff_{j},\sigma^2_n\II}
\\
\label{eq:gaussDef}
\pc{\bff_j}{\ba_{\psB{j}}} 
&=
\NN{\bff_j}{\bH_{j}\ba_{\psB{j}},\overline{\bV}_{j}} 
\\
\label{eq:gaussDef2}
\pc{\ba_j}{\ba_{\prB{j}}} 
&=
\NN{\ba_j}{\bF_{j}\ba_{\prB{j}},\bQ_{j}},
\end{align}
where \eqref{eq:likelihood} is the usual Gaussian likelihood for GP regression with noise variance $\sigma^2_n$, \eqref{eq:gaussDef} the projection conditional and  \eqref{eq:gaussDef2} the prior transition. Thereby, the matrices
are defined as
\begin{align*}
\bH_{j}
&=
\bK_{\bX_{j}\bA_{\psB{j}}}  \bK_{\bA_{\psB{j}}\bA_{\psB{j}}}^{-1}
\in\RR^{B\times LC}
;\\
\overline{\bV}_{j}
&=
Diag[
\bK_{\bX_j\bX_j}  -
\bK_{\bX_j\bA_{\psB{j}}} \bK_{\bA_{\psB{j}}\bA_{\psB{j}}}^{-1} \bK_{\bA_{\psB{j}}\bX_j}
]
\in\RR^{B\times B}
;\\
\bF_{j}
&=
\bK_{\bA_{j}\bA_{\prB{j}}}  \bK_{\bA_{\prB{j}}\bA_{\prB{j}}}^{-1}
\in\RR^{L\times LI_j}
;\\
\bQ_{j}
&=
\bK_{\bA_j\bA_j}  -
\bK_{\bA_j\bA_{\prB{j}}} \bK_{\bA_{\prB{j}}\bA_{\prB{j}}}^{-1} \bK_{\bA_{\prB{j}}\bA_j}
\in\RR^{L\times L}
\end{align*}
with $\bF_1 = \bs{0}$ and $\bQ_1=\bK_{\bA_1\bA_1}$.
\end{proposition}
%
%


The 
two  conditional distributions 
\eqref{eq:gaussDef}  and  \eqref{eq:gaussDef2} 
can  
 be derived from the true joint prior distribution $p(\ba,\bff,\by)$ 
as shown in 
Proof \ref{proof:graphicalModel}.
Alternatively, a generalization of this model can be obtained when using a  modified  projection distribution $\pc{\bff_j}{\ba_{\psB{j}}} $ so that 
for $C\rightarrow J$  and $\gamma <1$ our model recovers a range of well known global sparse GP methods as described in Section \ref{se:variational} and Prop.\ \ref{prop:equality}.
In any case,
these local conditional distributions lead to the following joint distribution.

%

\newpage
\begin{definition}[Joint Distribution]
\label{def:jointDistr}
For the graphical model corresponding to graph $\mathcal{G}$,
the joint distribution 
over all variables 
$\bff, \ba,\by$
can be written as
\begin{align*}
q_{c,\gamma}(\bff, \ba,\by)
=
\prod_{j=1}^J
\pc{\by_{j}}{\bff_{j}}
\pc{\bff_{j}}{ \ba_{\psB{j}} }
\pc{\ba_{j}}{ \ba_{\prB{j}} }.
\end{align*}
%
In the case $\gamma=1$ and thus 
$\ba=\bff$, the joint distribution 
simplifies (Proof\ \ref{proof:jointDistr_full}) to
\begin{align*}
q_{c,1}(\bff,\by)
=
\prod_{j=1}^J
\pc{\by_{j}}{\bff_{j}}
\pc{\bff_{j}}{ \bff_{\prB{j}} }.
\end{align*}
\end{definition}
We use $q=q_{c,\gamma}$ instead of $p$ in order to indicate that it is an approximate distribution.
%
%
%
%
%
%
The joint distributions in Def.\ \ref{def:jointDistr} 
 and the corresponding graphical model in Fig.\ \ref{fig:GPmods2}
%
%
 allow interesting comparisons to other GP models in 
 Fig.\ \ref{fig:GPmods} and the corresponding formulas \eqref{eq:postFull}, \eqref{eq:sparse}, \eqref{eq:PoE}. Whereas the conditioning set 
for full GP
 are all the previous latent values $\bff_{1:j-1}$, for  sparse GPs 
 some global  inducing points $\ba$ 
and for local independent experts 
 the empty set, we propose to condition on the $C-1$ predecessors $\bff_{\prB{j}}$ (or a sparsified version in the general case).
%
%
From this point of view, we can notice that our probabilistic model is equal to full GP, sparse GP and PoEs under certain circumstances which are more precisely formulated in Prop.\ \ref{prop:equality}.

\subsection{Sparse and Local Prior Approximation}
\label{se:priorAprox}

\begin{figure}[htp!]
	\centering
  \includegraphics[width=.85\textwidth]{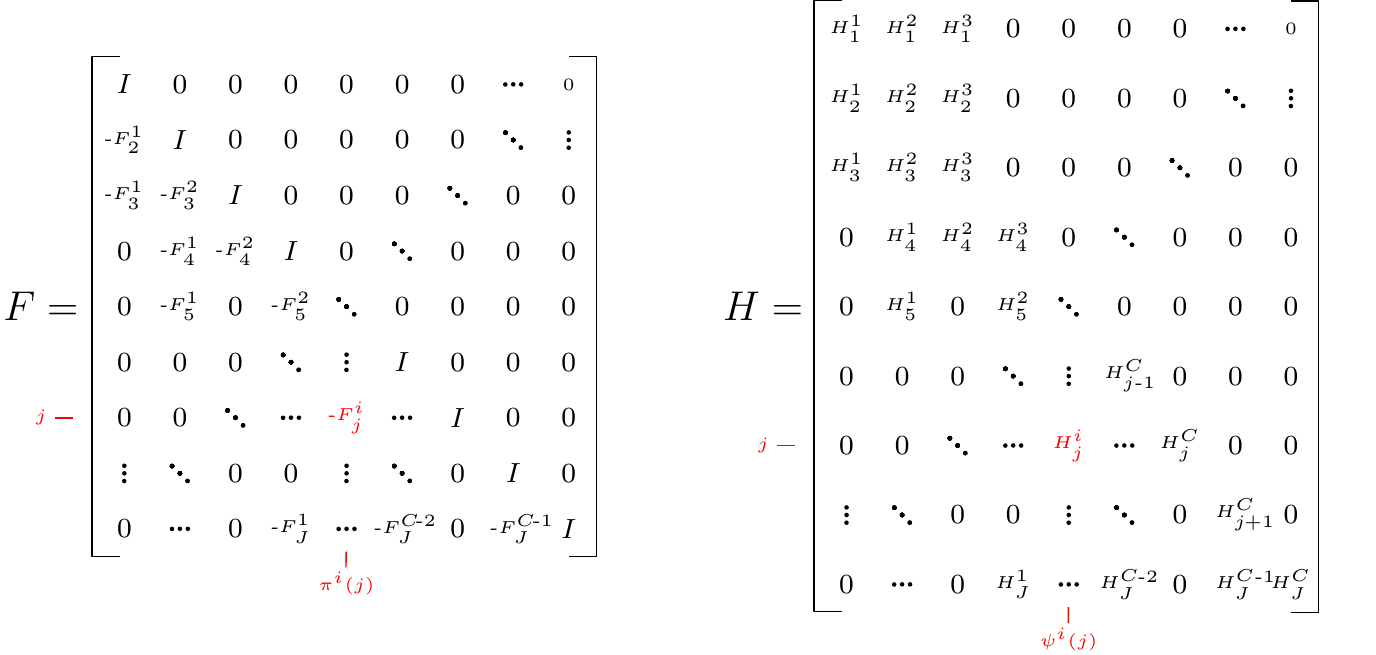}
  \caption{Sparse transition matrix $\bF \in \RR^{M\times M}$ and sparse projection matrix
  $\bH\in \RR^{N\times M}$, where $\bF_j^i \in \RR^{L\times L}$ and 
  $\bH_j^i\in \RR^{B\times L}$ are the $i$th part of $\bF_j \in \RR^{L\times L(C-1)}$ and  $\bH_j \in \RR^{B\times LC}$, respectively, corresponding to the contribution of the $i$th predecessor 
    $\bs{\pi}^i(j)$
    and
    $\bs{\psi}^i(j)$.
  }
 \label{fig:transF}
\end{figure}

The conditional independence  assumptions between the experts induced by the predecessor structure 
$\bs{\pi}_{C}$ 
lead to an approximate prior $q_{c,\gamma}(\ba)$ and approximate projection $q_{c,\gamma}(\bff\vert\ba)$ yielding a sparse and local joint prior $q_{c,\gamma}(\ba,\bff,\by)$.
\begin{proposition}[Joint Prior Approximation, Proof \ref{proof:priorApprox}]
\label{prop:priorApprox}
The  prior over all local inducing points $\ba$
in our CPoE model 
is
\begin{align*}
q_{c,\gamma}(\ba)
=
\prod_{j=1}^J
\pc{\ba_{j}}{ \ba_{\prB{j}} }
=
\NN{\ba}{\bs{0},\bS_C^{-1}},
\end{align*}
with the prior precision 
$\bS_C = \bS=\bF^T \bQ^{-1} \bF \in \RR^{M\times M}$ where
$\bQ =\DIAG{\bQ_1,\ldots,\bQ_J} \in \RR^{M\times M}$ 
 and 
$\bF \in \RR^{M\times M}$ is given as the  
sparse lower triangular matrix
in Fig.\ \ref{fig:transF}. 
%
%
%
%
Morover, the projection
 is
\begin{align*}
q_{c,\gamma}(\bff\vert\ba)
=
\prod_{j=1}^J
\pc{\bff_{j}}{ \ba_{\psB{j}} }
=
\NN{\bff}{\bH\ba,\overline{\bV}},
\end{align*}
where
$\bH \in \RR^{N\times M}$ defined 
in Figure \ref{fig:transF}
and
$\overline{\bV} =\DIAG{\overline{\bV}_1,\ldots,\overline{\bV}_J} \in \RR^{N\times N} $.
Together with the exact likelihood
$\pc{\by}{\bff}=\prod_{j=1}^J p(\by_j\vert \bff_j )=
\NN{\by}{\bff,\sigma_n^2\II}$ determines the joint approximate prior
\begin{align*}
q_{c,\gamma}(\ba,\bff,\by)
&=
p(\by\vert \bff)~
q_{c,\gamma}(\bff\vert\ba)~
q_{c,\gamma}(\ba).
\end{align*}
\end{proposition}
%

%
%
Note that the joint prior $q_{c,\gamma}(\ba,\bff,\by)$ is Gaussian 
$\NNo{\bO,\bW}$
with dense covariance $\bW$ and sparse precision $\bZ = \bW^{-1}$ as shown in Fig.\  \ref{fig:joint_prior_wo_star} in the Appendix.
If the predecessor set is consecutive, the matrix $\bF$ is a lower band (block)matrix with bandwidth $C$ and in the non-consecutive case each row has exactly $C$ non-zero blocks. The sparsity pattern of $\bF$ is inherited to  
the prior precision $\bS = \bF^T \bQ^{-1} \bF$ which is also a sparse matrix (see Fig.\ \ref{fig:GPmods4}). 
For the consecutive case, $\bS$ is a block-band matrix with bandwidth $2C-1$.
Note that the inverse $\bS^{-1}$ is dense.
%
%
%
%
%
%
%
The likelihood matrix $H$ is exact in the corner up to indices $C$ which ensures that we recover  sparse global GP in the limiting case $C=J$.
%
%
%
%
%
%
%
%
The quality of the approximation of our CPoE$(C, \gamma)$ model 
is discussed in Section \ref{se:properties} where we show that $q_{c,\gamma}(\ba,\bff,\by)$ converges to the true prior $p(\ba,\bff,\by)$ for $C \rightarrow J$.

\subsection{Inference}
\label{se:inference}

\begin{figure}[htp!]
	\centering
  \includegraphics[width=.95\textwidth]{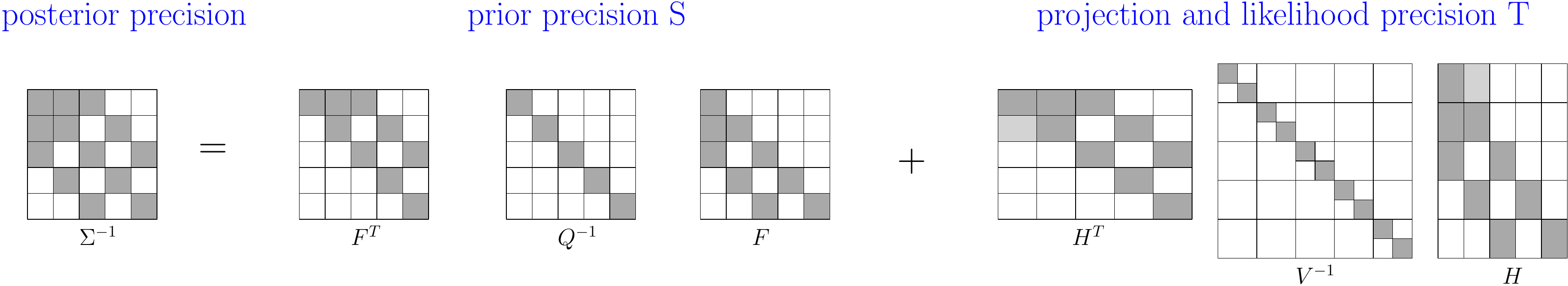}
  \caption{Sparse  posterior precision approximation.}
    \label{fig:precision}
\end{figure}

For our model it is possible to infer analytically the posterior $q_{c,\gamma}(\ba\vert \by)$ and the marginal likelihood $q_{c,\gamma}(\by)$ used later for prediction  and for hyperparameter estimation, respectively.
%
%
%
%
\begin{proposition}[Posterior Approximation; Proof \ref{proof:posterior}]
\label{prop:posterior}
From the joint distribution,
 the latent function values 
$\bff$ 
can be integrated out 
yielding
\begin{align*}
q_{c,\gamma}( \ba,\by)
&=
\int q_{c,\gamma}(\bff, \ba,\by)\diff \bff
=q_{c,\gamma}(\by\vert \ba)
q_{c,\gamma}(\ba)
=
\NN{\by}{\bH\ba,\bV}
\NN{\ba}{\bO,\bS^{-1}}
\end{align*}
with
$\bV=\overline{\bV} + \sigma_n^2\II \in \RR^{N\times N}.$
The posterior 
can be analytically computed by
\begin{align*}
q_{c,\gamma}( \ba\vert \by)
&=
\frac{q_{c,\gamma}( \ba,\by)}{q_{c,\gamma}( \by)}
\propto 
q_{c,\gamma}(  \ba,\by)
=
\NN{\ba}{\bmu,\bSig},
\end{align*}
with
$\bSig
=\left(
\bT
+
\bS
\right)^{-1} \in \RR^{M\times M}$,
$\bmu
= \bSig\bb \in \RR^{M}$, 
$\bb = \bH^T\bV^{-1}\by \in \RR^{M}$
and
$\bT = \bH^T\bV^{-1}\bH \in \RR^{M\times M}$.
\end{proposition}
%
%
%
%
%
%
%
%
The posterior precision matrix  $\bSig^{-1}=
\bT
+
\bS$ inherits the sparsity pattern of the prior since the addition of 
the projection precision $\bT=\bH^T\bV^{-1}\bH$ has the same sparsity structure  as depicted in Figs.\ \ref{fig:GPmods4} and
 \ref{fig:precision}.
%
On the other hand, the posterior covariance  $\bSig$ is dense, therefore it will be never explicitly fully computed.
Instead,  the sparse linear system of equations
$
\bSig^{-1}\bmu = \bb
$
can be efficiently solved for $\bmu = \bSig\bb$. 
%

Further, in our CPoE model, 
the marginal likelihood $q_{c,\gamma}(\by\vert \bthe)$
can be analytically computed by
$\int q_{c,\gamma}(\by,\ba) \diff \ba
=
  \NNo{\bO,\bP}$
  (see Proof \ref{prop:margLik}) 
 with the (dense) matrix
$\bP=\bH\bS^{-1}\bH^T +\bV \in 
\RR^{N\times N}$
 which is used in Section \ref{se:hyperEst} for hyperparameter optimization.
 
	 The  posterior approximation $q_{c,\gamma}(\ba\vert\by)$ as well as
	 the
	  approximate marginal likelihood $q_{c,\gamma}(\by)$ 
	  converge to the true distributions $\pc{\ba}{\by}$ and
	  $\pp{\by}$, respectively,
	  for $C\rightarrow J$. In particular, they correspond exactly 
	 to the posterior and marginal likelihood of full GP 
	 and sparse global GP
with $\lfloor \gamma N\rfloor$ inducing points
	 for 
	 $C=J, \gamma=1$ and
   $C=J, \gamma<1$, respectively.

\quad\\
\subsection{Prediction}
\label{se:prediction}
The final predictive posterior distribution 
is obtained by an adaptation of the PoE aggregation
in \eqref{eq:PoEagg}.
The main idea is to consistently aggregate weighted local predictions form the experts 
such that the correlations between them are taken into account
resulting in a smooth and continuous predictive distribution.
\begin{proposition}[Prediction Aggregation; Proof \ref{proof:predAgg}]
\label{prop:predAgg}
Similarly to the PoE aggregation in
\eqref{eq:PoEagg}, 
we define the final predictive posterior distribution 
$q_{c,\gamma}(f_{*}\vert \by)$ 
for a query point $\bx_*\in\RR^D$ as
\begin{align*}
q_{c,\gamma}(f_*\vert\by)
=
\prod_{j=C}^{J}
q_{c,\gamma}(f_{*j}\vert\by)
^{\beta_{*j}},
\end{align*}
involving 
 the local predictions
$q_{c,\gamma}( f_{*j}\vert \by)=\NNo{m_{*j},v_{*j}}$  
and 
 weights $\beta_{*j}\in \RR$  
 defined in Prop.\ \ref{prop:locPred} and Def.\ \ref{def:aggWeights}, respectively.
 %
 Moreover, the distribution $q_{c,\gamma}(f_*\vert\by)=\NNo{m_*,v_*}$ with
$
m_*
=
v_{*}
\sum_{j=C}^{J}
\beta_{*j}
\frac{m_{*j}}{v_{*j}}
$
and
$
\frac{1}{v_*}
=
\sum_{j=C}^{J}
\frac{\beta_{*j}}{v_{*j}}$
 is  analytically available.
 The  final noisy prediction is
 $\pc{y_*}{\by}=\NNo{m_*,v_* + \sigma_n^2}$.
\end{proposition}
The graphical model corresponding to this prediction procedure 
is depicted in Fig.\ 	\ref{fig:pred2}.
Note that the first $C-1$ experts are  only implicitly considered since 
 $\psB{j} = \psB{C}$
 for $j\leq C$, resulting in $J_2=J-C+1$ predictive experts
 so that
the proposed prediction aggregation
interpolates 
between predictions from $J_2=J$ completely independent experts 
and predictions from $J_2=1$ fully correlated expert
 which is depicted in Fig.\ 	\ref{fig:predAgg} in the Appendix.


 %
%
\begin{proposition}[Local Predictions, Proof \ref{proof:locPred}]
\label{prop:locPred}
The local prediction 
$q_{c,\gamma}(f_{*j}\vert\by)=\NNo{m_{*j},v_{*j}}$ 
of the $j$th expert are based on the region $\psB{j}$ where the correlations
 are modelled and can be computed as
\begin{align*}
q_{c,\gamma}(f_{*j}\vert\by)
=
\int
\pc{f_{*j}}{\ba_{\psB{j}}}
q_{c,\gamma}(\ba_{\psB{j}}\vert\by)
\diff
\ba_{\psB{j}}
=
\NNo{\bh_*\bmu_{\psB{j}},
\bh_*^T\bSig_{\psB{j}} \bh_*+v_*}
\end{align*}
involving 
the local posteriors
$
q_{c,\gamma}(\ba_{\psB{j}}\vert\by)
=
\NNo{\bmu_{\psB{j}}, \bSig_{\psB{j}}}
$
and the
predictive conditional 
$\pc{f_{*j}}{\ba_{\psB{j}}}
=
\NNo{\bh_*\ba_{\psB{j}},v_*}$ 
 (which is exactly defined in  Proof \ref{proof:locPred}).
\end{proposition}
The local posteriors
with  mean $\bmu_{\psB{j}}$  and covariance entries $\bSig_{\psB{j}}$
could be obtained from the
corresponding 
entries $\psB{j}$
of
$\bmu$ and $\bSig$. However,  computing explicitly some entries in the dense covariance   $\bSig$ based on the sparse precision $\bSig^{-1}$ 
is not straightforward since in the inverse the blocks are no longer independent. However, we can exploit the particular sparsity  and block-structure of our precision matrix
and obtain an efficient implementation of this part which is key to achieve a competitive performance of our algorithm.
More 
details are given in the Appendix 
in 
 Section \ref{se:systemAndInv}.

\begin{definition}[Aggregation Weights]
\label{def:aggWeights}
The input depending weights $\beta_{*j}=\beta_j(X_*)$ 
  at query point $X_*$ models the influence of expert $j$. In particular, the unnormalized weights 
\begin{align*}
\begin{split}
\bar{\beta}_{*j}
&= 
H[\pp{f_{*}}]
-
H[\pc{f_{*j}}{\by}]
=
\frac{1}{2}
\log
 \left(
\frac{v_{*0}}{v_{*j}} \right),
\end{split}
\end{align*}
are set to the difference in 
entropy $H$ \eqref{eq:entropy} before and after seeing the data similarly proposed by \cite{fleet2014generalized}. Thereby, 
the predictive prior is
$\pp{f_{*}}=\NNo{0,v_{*0}}$ 
with
$v_{*0} = \bs{k}_{X_*X_*}$ and the predictive posterior defined in Prop.\ \ref{prop:locPred}. The normalized weights are then obtained by
$\beta_{*j} = b^{-1}\bar{\beta}_{*j}^Z$
where
$b = \sum_{j=C}^J \bar\beta_{*j}^Z$
and $Z=\log(N) C$. 
\end{definition}

These weights 
bring the flexibility of increasing or reducing the importance of the experts
based on the predictive uncertainty.
%
%
%
However, independent of the particular weights, our aggregation of the predictions 
is consistent since it is based on the \textit{covariance intersection} method \cite{julier1997non}, 
which is useful for combining several estimates of random variables with known mean and variance but unknown correlation between them.
%
%
%
%
%
The $Z$ in the exponent of the normalization of the weights 
 has a sharpening effect, so that 
  the informative experts have even more weight compared to the non-informative experts
for more data $N$ and more correlations $C$.
This is a heuristic but showed quite robust performance in experiments. Moreover, the consistency properties are more relevant than the particular weights. 
\subsection{Properties}
\label{se:properties}
\begin{proposition}[Equality; Proof \ref{proof:equality}]
\label{prop:equality}
Our  model correlated Product of Experts CPoE$(C,\gamma)$ is equal to full GP for $C=J$ and $\gamma=1$.
For $\gamma<1$, our model correspond to sparse global GP
with $M_g= \lfloor \gamma N \rfloor$ inducing points. 
Further, with $C=1$ and $\gamma=1$, our model is equivalent to independent PoEs. 
That is, we have
$$
\text{CPoE}(J,1 ) = \text{GP};
\quad
\text{CPoE}(J,\gamma ) = \text{SGP}(\lfloor\gamma N\rfloor);
\quad
\text{CPoE}(1,1) = \text{GPoE}^*,
$$
where SGP refers to the FTIC model \cite{snelson2006sparse}
 and 
GPoE$^*$ correspond to GPoE  \cite{fleet2014generalized} with slightly different weights ($Z=1$) in the prediction.
\end{proposition}

In Section  \ref{se:variational} in the Appendix
we present a generalization of  our model
 so that CPoE($J,\gamma$) correspond to a range of other well known versions of sparse global GP 
by
changing 
the projection distribution 
and 
adding 
a correction term 
in the log marginal likelihood
similarly discussed in
\cite{schurch2020recursive} for the global case. 
For instance, we can extend our model analogously to the variational version of \cite{titsias2009variational}.

\begin{figure}[hp!]
	\centering
  \includegraphics[width=.9\textwidth]{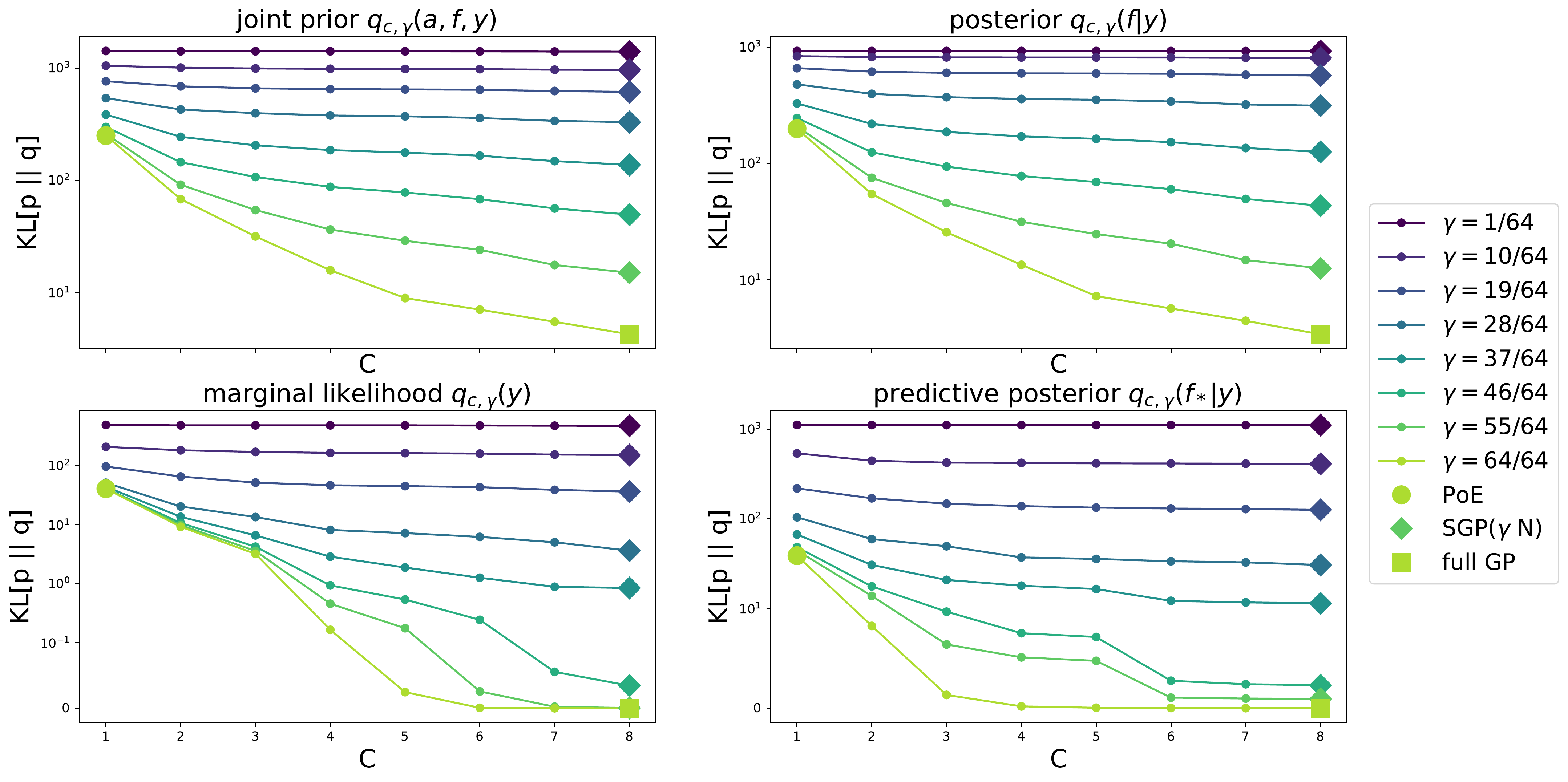}
	\caption{ 
Decreasing KL$[p \vert\vert q]$ between true distribution $p$ of full GP and approximate distribution $q=q_{c,\gamma} $ of
CPoE for increasing values of $C$ and $\gamma$
for the joint prior, posterior, marginal likelihood and predictive posterior for synthetic GP data ($N=1024, D=2$, SE kernel).
}
	\label{fig:KLpriorpost}
\end{figure}

For correlations between the limiting cases $C=1$ and $C=J$, 
we investigate the difference in KL of the true GP model with
CPoE$(C,\gamma )$ and CPoE$(C_2,\gamma )$ for $1\leq C\leq C_2\leq J$.
%
%
%
%
For that reason, we define the difference in KL
 between the true distribution of $\bx$ and two different approximate distributions, i.e.
$$
\mathbb{D}_{(C,C_2)}[\bx]
=
KL[\pp{\bx} \mid\mid q_{c,\gamma}(\bx)]
-
KL[\pp{\bx} \mid\mid q_{{c_2,\gamma}}(\bx)]
$$
Similarly, the difference in KL for a conditional distribution is defined in Eq.\ \eqref{eq:KL_diff2}.
%
%
%
Using these definitions, 
 we show that the approximation quality of the prior  $q_{c,\gamma}(\ba)$ 
 and projection approximation $q_{c,\gamma}(\bff \vert \ba)$ 
 monotonically improves for $C\rightarrow J$ so that 
the KL between the true joint distribution $p(\ba, \bff, \by)$ and our approximate joint distribution $q_{c,\gamma}(\ba, \bff, \by)$ 
 is decreasing for $C\rightarrow J$.
\begin{proposition}[Decreasing KL; Proof \ref{proof:convJointPrior}]
\label{prop:convJointPrior}
For any predecessor structure $\bs{\pi}_C$ and any
 $0<\gamma\leq 1$ 
 and $1\leq C\leq C_2 \leq J$,
the difference in KL 
of the
marginal
prior, projection and data likelihood are non negative, i.e.
\begin{align*}
\mathbb{D}_{(C,C_2)}[\ba] \geq~ 0;
\quad\quad
\mathbb{D}_{(C,C_2)}[\bff\vert\ba] \geq~ 0;
\quad\quad
\mathbb{D}_{(C,C_2)}[\by \vert \bff] = ~ 0,
\end{align*}
so that the joint difference in KL is also non-negative
\begin{align*}
&~\mathbb{D}_{(C,C_2)}[\ba,  \bff, \by]
=
\mathbb{D}_{(C,C_2)}[\ba]
+
\mathbb{D}_{(C,C_2)}[\bff\vert \ba]
+
\mathbb{D}_{(C,C_2)}[\by\vert \bff]
~\geq~ 0.
\end{align*}
Moreover, we can quantify the  approximation quality, in particular
\begin{align*}
\mathbb{D}_{(C,C_2)}[\ba]
=\frac{1}{2}
\log
\frac{\vert\bQ_{C}\vert}{\vert\bQ_{C_2}\vert}
\quad\quad\text{and}\quad\quad
\mathbb{D}_{(C,C_2)}[\bff\vert \ba]
=
\frac{1}{2}
\log
\frac{\vert\bar{\bV}_{C}\vert}{\vert\bar{\bV}_{C_2}\vert}.
\end{align*}
\end{proposition}
The last statement demonstrates that our CPoE model is a sound GP prior precision approximation which converges monotonically to the true prior for $C\rightarrow J$.
Moreover, we can quantify the relative approximation quality of our model which
 constitute an approach of estimating the needed $C$
since
it is independent of the true (and non-calculable) full GP distribution.
The decreasing KL of the joint prior is depicted in Fig.\ \ref{fig:KLpriorpost} together with the decreasing KL of the posterior, marginal likelihood and predictive posterior.
More details and proofs are given in Appendix \ref{se:proofAdd}.

\subsection{Computational Details}
\label{se:pracDet}

\subsubsection{Graph}
\label{se:graphDetails}
\begin{figure}[htp!]
	\centering
  \includegraphics[width=.8\textwidth]{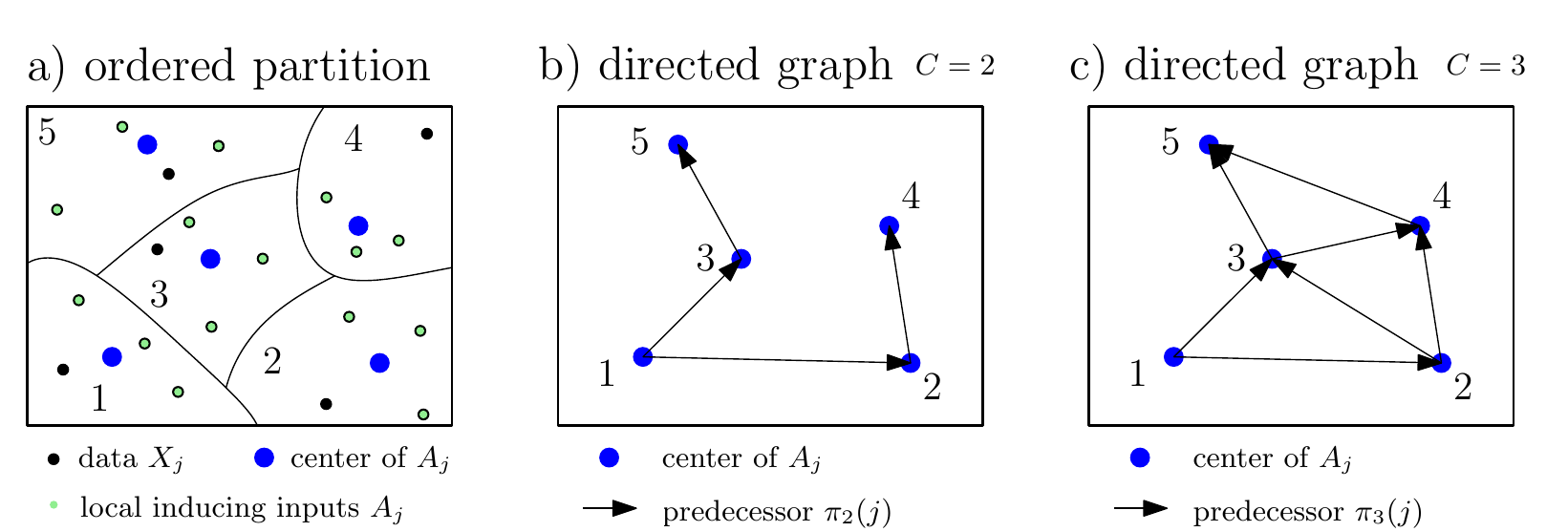}
	\caption{ Toy example for partition, local inducing points, predecessors and directed graph illustrated for
	  $D=2$ with $J=5$ experts/partitions each with $B=4$ samples, $\gamma=0.75$ and thus $L=3$ local inducing points. In a) the ordered partition with the data (black), local inducing points (green) and their mean (blue) are depicted. In b) and c) the directed graph for $C=2$  and $C=3$ are shown with corresponding predecessors 
	$\bs{\pi}_2(1)=\{\}$, $\bs{\pi}_2(2)=\{1\}$, $\bs{\pi}_2(3)=\{1\}$, $\bs{\pi}_2(4)=\{2\}$, $\bs{\pi}_2(5)=\{3\}$
and
		$\bs{\pi}_3(1)=\{\}$, $\bs{\pi}_3(2)=\{1\}$, $\bs{\pi}_3(3)=\{1,2\}$, $\bs{\pi}_3(4)=\{2,3\}$, $\bs{\pi}_3(5)=\{3,4\}$, respectively.
In the previous example, $\bs{\pi}_3$ is consecutive and $\bs{\pi}_2$ is non-consecutive.	
}
	\label{fig:GPmods3}
\end{figure}

The graphical model in Section \ref{se:graphicalModel} is generically defined and several choices are left for  
 completely specifying 
 the graph $\mathcal{G}(V,E)$ for a particular dataset:
the partition method, the ordering of the partition, the selection of the predecessors and the local inducing points. We tried to make these choices as simple and straightforward as possible with focus on computational efficiency, however, there might be more sophisticated heuristics.
Concretely, we use KD-trees \cite{maneewongvatana2001efficiency} for partitioning the data $\mathcal{D}$ into $J$ regions and 
the ordering starts with a random partition which is then greedily extended by the closest  partition in euclidean distance (represented by the mean of the inducing points).
The $L\leq B$ inducing inputs $\bA_j\in \RR^{L\times D}$ of the $j$th partition (or expert) can be in principle arbitrary, however, in this work they are chosen as a random subset of the data inputs $\bX_j\in \RR^{B\times D}$ of the $j$th expert for the sake of simplicity.
For the predecessors (block-)indices $\bs{\pi}_C$, the $C-1$ closest partitions among the previous (according to the ordering) predecessors in euclidean distance are greedily selected.
These  explained concepts are illustrated for a toy example
 in Fig.\ \ref{fig:GPmods3}.
%

%

\subsubsection{Hyperparameter Estimation}
\label{se:hyperEst}

In Section \ref{se:CPoE},
we introduced CPoE for fixed hyperparameters $\bthe$ where implicitly all distributions are conditioned on  $\bthe$, however, we omitted the dependencies on $\bthe$ in the most cases for the sake of brevity. Similar to full GP, 
sparse GP or PoEs, the 
\textit{log  marginal likelihood (LML)}
can be used as an objective function for optimizing the few hyperparameters $\bthe$.
The 
log of the marginal likelihood 
of our model 
formulated in Section \ref{se:inference}
is
$
\mathcal{L}(\bthe)
=
\log \qc{\by}{\bthe}   = \log  \NNo{\bO,\bP}
$
  with
$\bP=\bH\bS^{-1}\bH^T +\bV$
which can be efficiently computed as detailed in Section \ref{se:hyperEst2} and can be used for \textit{deterministic optimization} with full batch $\by$  
for moderate sample size $N$.
However,
in order to scale this parameter optimization part to larger number of samples $N$ in a competitive time, \textit{stochastic optimization} techniques exploiting subsets of data have to be developed  similarly done  for the global sparse GP model (SVI \cite{hensman2013gaussian}; REC \cite{schurch2020recursive}; IF \cite{kaniasparse}). 
We adapt the hybrid approach IF of
\cite{kaniasparse} where we can also exploit an independent factorization of the log marginal  likelihood which decomposes into a sum of $J$  terms, so that it can be used for stochastic optimization.
This constitutes a very fast and accurate alternative for our method as shown in the Appendix \ref{se:hyperEst2}
and will also be exploited in Section \ref{se:examples_evaluation} for large data sets.
\\
\quad
\\
\textbf{Prior on Hyperparameters}\\
Alternatively to the 
log marginal likelihood 
(LML) maximization as presented above,  the \textit{maximum a posteriori }(MAP) estimator for $\bthe$ can be used.
It is  
the log of the posterior distribution  
$\pc{\bthe}{\by} \propto \pc{\by}{\bthe} \pp{\bthe}$ where $\pp{\bthe}$ is a  suitable prior on the hyperparameters yielding
%
$
	\log \pc{\bthe} {\by}
	= 
	\log \pc{\by}{\bthe}  + 	\log \pp{\bthe}.$
In the following, we assume $\pp{\bthe}=\prod_{j}\pp{\bthe_j}$ and a log-normal prior for each hyperparameter
$
	\pp{\bthe_j}=\log\NN{\bthe_j}{\nu_j,\lambda_j^2}
	$
for means $\nu_j$ and variances $\lambda_j^2$.
\\
For the deterministic case,
the MAP estimator 
can be straightforwardly computed by just adding the log prior on $\bthe$ to the batch log marginal likelihood, 
 i.e.\
$\log \pc{\bthe} {\by} = \mathcal{L}(\bthe) + \log \pp{\bthe}$.
Similarly for the stochastic case,
the stochastic MAP can be decomposed as
$
	\log \pc{\bthe} {\by}
	\approx
	\sum_{j=1}^J \left(  l_j(\bthe) + 	\frac{1}{J}\log \pp{\bthe} 
	\right),
	$
	where 
	$l_j(\bthe)$ is the $j$th term in the stochastic marginal likelihood (defined properly in the Appendix \ref{se:hyperEst2}),
%
so that it can be  used again for stochastic mini-batch optimization.
An example using priors for the hyperparameters is presented in Section \ref{se:timeseries}.

 \subsubsection{Complexity}
 \label{se:complexity}
The time complexity for computing the posterior and the marginal likelihood in our algorithm is dominated by $J$ operations which are cubic in $LC$ (inversion, matrix-matrix multiplication, determinants).
This leads to $\mathcal{O}(NB^2\alpha^3)$ and $\mathcal{O}(NB\alpha^2)$ for time and space complexity, respectively, where we define the approximation quality parameter $\alpha=C\gamma$. Similarly, for $N_t$ testing points the time and space complexities are 
 $\mathcal{O}(N B\alpha^2 N_t)$
 and
  $\mathcal{O}(N \alpha N_t)$ (an approach to remove the dependency of $N$ is outlined in  \ref{se:complexity2}).
In Table \ref{tab:complexity}, the asymptotic complexities of our model together with other GP algorithms are indicated.
It is interesting that for $\alpha= 1$, our algorithm has the same asymptotic complexity for training as sparse global GP with $M_g=B$ global inducing points but we can have $M = LJ=\gamma BJ=\gamma N$ total local inducing points! Thus,
	our approach allows much
more total local inducing points $M$ in the order of $N$ (e.g. $M=0.5N$ with $C=2$) whereas for sparse global GP usually $M_{g} \ll N$. This  has the consequence that the local inducing points can cover the input space much better and therefore represent much more complicated functions. As a consequence,  there is also no need to optimize the local inducing points resulting in much fewer parameters to optimize. Consider the following example with $N=10'000$ in $D=10$ dimensions. Suppose a sparse global GP model with $M_g=500$ global inducing points. A CPoE model with the same asymptotic complexity has a batch size $B=M_g=500$ and $\alpha=1$. Therefore, we have $J=\frac{N}{B}=20$ experts and we choose $C=2$ and $\gamma=\frac{1}{2}$ such that we obtain $L=\gamma B=250$ local inducing points per experts and $M = \gamma N = 5'000$ total local inducing points! Further, the number of hyperparameters to optimize with a SE kernel is for global sparse GP $M_g D+\vert\bthe\vert = 5012$, whereas for CPoE there are only $\vert\bthe\vert = 12$.
For an extended version of this section  consider  \ref{se:complexity2} in the Appendix.

\begin{table}
    \begin{small}
\centering
  \begin{tabular}{ l | c  c c c }
    \toprule
     & full GP & sparse GP & PoE & CPoE \\ 
     \midrule
    time &$ \mathcal{O}(N^3)$ & $ \mathcal{O}(NM_g^2)$ & $ \mathcal{O}(NB^2)$ & $ \mathcal{O}(NB^2\alpha^3)$ \\ 
    space & $\mathcal{O}(N^2)$  & $ \mathcal{O}(NM_g)$ & $ \mathcal{O}(NB)$& $ \mathcal{O}(NB\alpha^2)$ \\
     \midrule
    time$_{t}$ & $ \mathcal{O}(N^2 N_t)$
    & $ \mathcal{O}(M_g^2 N_t)$ 
    & $ \mathcal{O}(NBN_t)$ 
    &  $ \mathcal{O}(NBN_t \alpha^2)$ 
    \\
      space$_{t}$ & $ \mathcal{O}(N N_t)$
    & $ \mathcal{O}(M_g N_t)$ 
    & $ \mathcal{O}(NN_t)$ 
    &  $ \mathcal{O}(NN_t \alpha)$ 
    \\
      \midrule
    $\#$pars & $\vert\bthe\vert$
    & $MD+ \vert\bthe\vert$ & $\vert\bthe\vert$ & $\vert\bthe\vert$
    \\
    \bottomrule
  \end{tabular}
  \caption{Complexity for training, pointwise predictions for $N_t$ points and number of optimization parameters for different GP algorithms. }
  \label{tab:complexity}
    \end{small}
  \end{table}

\section{Comparison}
\label{se:examples_evaluation}

%

In this section we compare the performance with competitor methods for GP approximations using several \textit{synthetic} and \textit{real world }datasets 
as summarized in Table \ref{tab:datasets}.
%
%
More details about the experiments  
are provided in
\ref{se:imp_details},
 \ref{se:experiments}
 and  
  \ref{se:tables} in the Appendix.

\begin{figure*}[htp!]
	\centering
  \includegraphics[width=.95\textwidth]{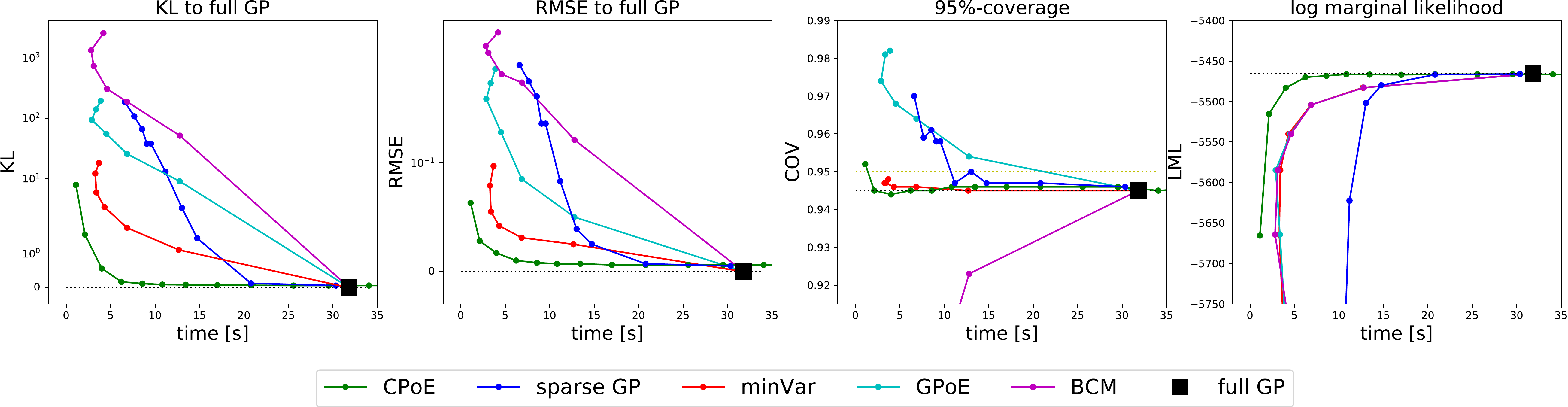}
	\caption{  Average accuracy vs. time performance of different GP algorithms.
	 }
	\label{fig:synthetic2D}
\end{figure*}

%
First, we examine the accuracy vs. time performance of different GP algorithms for fixed hyperparameters in a simulation study with \textit{synthetic GP data}.
We generated $N=8192$ data samples in $D=2$ with 5 repetitions from the sum of two SE kernels 
with a shorter and longer lengthscale  such that both global and local patterns are present in the data (compare Fig.\ \ref{fig:data2D}).
In 	Fig.\ \ref{fig:synthetic2D} the mean results are shown for the KL and RMSE to full GP, the 95\%-coverage 
and the log marginal likelihood against time in seconds. 
The results 
for
 sparse GP with increasing number of 
global inducing points $M$ are shown in blue,  the results for minVar, GPoE and BCM
 for increasing number of experts $J$ are depicted in red, cyan and magenta, respectively. 
For CPoE, the results for increasing correlations  $C$ are  shown in green.
We observe  superior performance of our method compared to competitors  in terms of accuracy compared to full GP vs. time.
Moreover, one can observe that the confidence information of our model are reliable already for small approximation orders since it is based on the consistent covariance intersection method.
A precise description of the experiment is provided in Section \ref{se:synthData2} in the Appendix.


 %
  %
\begin{table}
\begin{small}
\centering

\begin{tabular}{l | rrrrr | rrrrr}
\toprule
&&& \textbf{KL} &&&&& \textbf{time}\\
\midrule
&  concrete &    mg &  space &  abalone &    kin &  concrete &    mg &  space &  abalone &    kin \\
\midrule
fullGP   &       0.0 &   0.0 &    0.0 &      0.0 &    0.0 
	 &       7.3 &  25.5 &  114.8 &    237.9 &  161.5 \\ \hline
SGP(100) &     352.9 &   9.9 &  108.1 &     15.6 &  603.7 
	 & 36.4 &  14.4 &   46.6 &     58.9 &   42.2\\
minVar   &     122.2 &  19.4 &   63.6 &     25.1 &  211.0 
	&       1.5 &   2.0 &    7.2 &      6.4 &    9.3\\
GPoE     &     174.4 &  54.2 &   98.0 &     50.3 &  342.3 
	&       1.4 &   1.9 &    7.2 &      6.3 &    9.4 \\
GRBCM    &     224.6 &  69.1 &  105.6 &     36.4 &  129.8 
	&       1.7 &   2.3 &    6.5 &      7.6 &   11.9\\
\textbf{CPoE(1) } &     111.1 &  12.2 &   63.0 &     16.8 &  152.4 
	&       1.5 &   2.1 &    7.8 &      6.4 &    9.2\\
\textbf{CPoE(2) } &      89.6 &   8.4 &   36.5 &      8.1 &   79.9 
	&       2.1 &   2.8 &   10.6 &      7.5 &   12.9\\
\textbf{CPoE(3) } &      82.2 &   7.8 &   36.3 &      6.2 &   46.9 
	 &       2.5 &   3.1 &   12.9 &      9.3 &   19.8\\
\textbf{CPoE(4) } &      \textbf{79.5} &   \textbf{7.6} &   \textbf{36.0} &      \textbf{4.7} &   \textbf{32.8} 
	 &       2.8 &   3.3 &   14.9 &     10.4 &   27.8\\
\bottomrule
\end{tabular}
\caption{Average KL to full GP (left) and time (right) for different GP methods and 5 datasets with 10 repetitions. 
More 
results are provided in
Appendix  \ref{se:tables}.
}
\label{tab:KL_time_real}
\end{small}

\end{table}

\newpage
Second,  we benchmark our method 
with 
 10  real world datasets as summarized in Table \ref{tab:datasets} 
  For the 5 smaller datasets in the first block we  use  deterministic  parameter  optimization for which
  the average results over 10 training/testing splits 
  are depicted
  in Table \ref{tab:KL_time_real}.
  In particular, the KL to full GP (left) and time (right)  for different GP methods and  are shown. 
Similarly, the average accuracy and times 
for the 4  larger datasets in the second block where stochastic parameter optimization is exploited
can be found 
in Table  \ref{tab:CRPS_time_real} in the Appendix.

In general, the local methods perform better than the global sparse method.
Further, the performance of our correlated PoEs is superior to the one of independent PoEs for all datasets. In particular, the KL to full GP can be continuously improved  for increasing degree of correlation, i.e. larger $C$ values.
The time for CPoE(1) is comparable with the independent PoEs and for increasing $C$, our approximation has a moderate increase in time with a significant decrease in KL.
 For more details about the experiments consider Section 
  \ref{se:experiments}
 in the Appendix and 
 more results 
  including standard deviations are provided in Appendix \ref{se:tables}.

\begin{table}[htp!]
   \begin{small}
   \begin{minipage}{.5\textwidth}
\centering
\subfloat[Description of  datasets.
   ]{
 \begin{tabular}{ l | c  c  c | c}
     \toprule
 \quad    & $N$ & $D$ & $N_{test}$ & $J$  \\ 
   \midrule
     concrete & 927 & 8 & 103  & 4 \\
     mg & 1247 & 6 & 138 & 8  \\
     space & 2797 & 6 & 310 & 8  \\
     abalone & 3760 & 8 & 417  & 16 \\
     kin & 5192 & 8 & 3000  & 16 \\
       \midrule
        kin2 & 7373 & 8 & 819  & 16  \\
     cadata & 19640 & 8 & 1000 & 64  \\
       sarcos & 43484 & 21 & 1000 & 128 \\
     casp & 44730 & 9 & 1000 & 128  \\
      \midrule
         elecdemand & 2184 & 3 & 15288  & 13 \\
    \bottomrule
  \end{tabular}
      \label{tab:datasets}
  }
   \end{minipage}
  ~
   \begin{minipage}{.5\textwidth}
   \subfloat[
 KL to full GP and time of different  methods.
 ]{
\begin{tabular}{l | rrr | r}
\toprule
{} &     KL &  KL IN &  KL OUT &   time \\
\midrule
full GP  &    0.0 &    0.0 &     0.0 &  404.3 \\
\midrule
SGP(100) &  120.9 &  110.5 &   146.7 &   56.3 \\
SGP(200) &  114.9 &   65.6 &   238.3 &   75.2 \\
minVar   &  503.0 &  406.5 &   744.5 &   20.7 \\
GPoE     &  328.0 &  336.0 &   307.9 &   20.4 \\
GRBCM    &  393.4 &  382.1 &   421.8 &   28.2 \\
\midrule
\textbf{CPoE(1) } &  289.5 &  255.1 &   375.5 &   20.5 \\
\textbf{CPoE(2)}  &  113.1 &  108.5 &   124.3 &   36.8 \\
\textbf{CPoE(3)}  &   86.4 &   61.9 &   147.6 &   39.7 \\
\textbf{CPoE(4)}  &   \textbf{58.3} &   \textbf{59.4} &    \textbf{55.5} &   52.9 \\
\bottomrule
\end{tabular}

  \label{tab:KL_ELEC}
  }
   \end{minipage}
    \caption{Summary of used datasets 
   and results for  the \textit{elecdemand} time series.}

      \end{small}
  \end{table}

\subsection{Application}
\label{se:timeseries}

\begin{figure}[htp!]
	\centering
  \includegraphics[width=.99\textwidth]{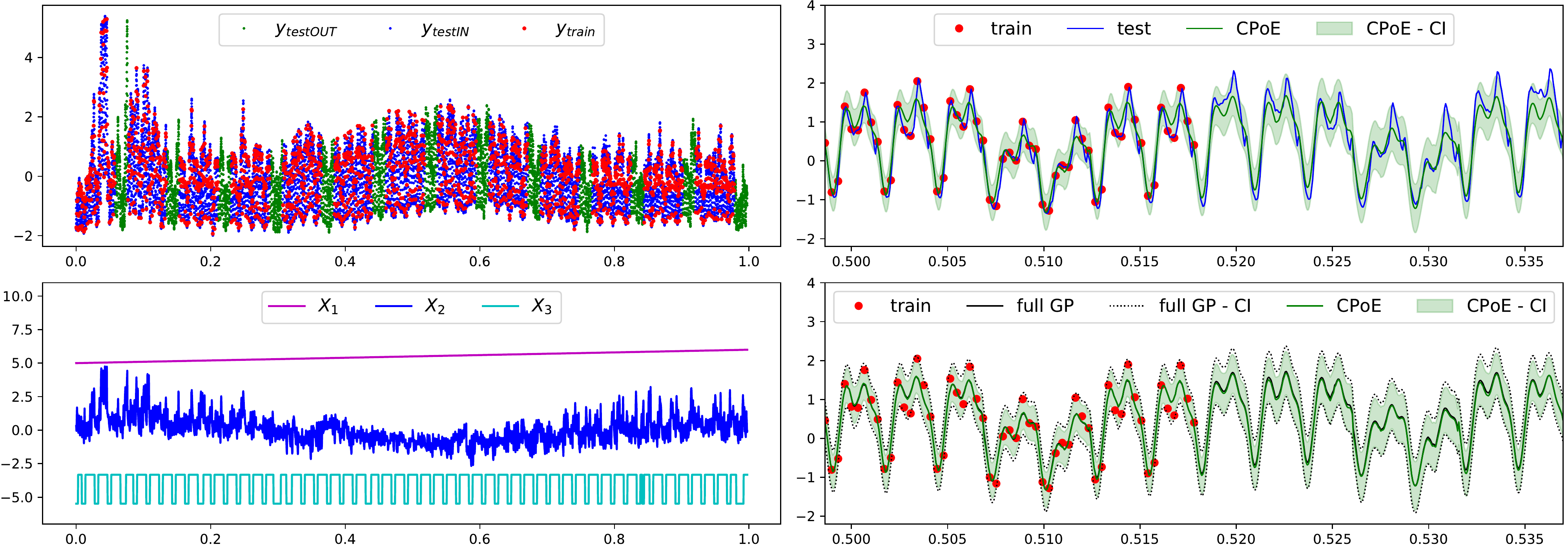}
	\caption{Time series data with covariates and prior on hyperparameters.
}
	\label{fig:timeSeries}
\end{figure}

In this Section, our method is applied on time series data with covariates  using a rather complicated and non-stationary kernel together with priors on the hyperparameters as discussed in Section \ref{se:hyperEst}. In recent work 
\cite{corani2021time},
the authors have shown that GPs constitute a competitive method for modelling time series using  a sum of several kernels including priors on the hyperparameters which are  previously learnt from a large set of different time series. We adapt their idea by using a slightly modified kernel and the same priors.
In particular, 
for two data points $\bx_1=[t_1, x_{1,2}, \ldots,x_{1,D}]$ and $\bx_2=[t_2, x_{2,2}, \ldots,x_{2,D}]$ we model the kernel as the sum of 4 components
\begin{align*}
k_{\bthe}(\bx_1,\bx_2) &=
k_{P_1}(t_1,t_2)
+
k_{P_2}(t_1,t_2)
+
k_{SM}(t_1,t_2)
+
k_{SE}(\bx_1,\bx_2),
\end{align*}
where $k_{P_1}$ and $k_{P_1}$ are standard periodic kernels with period $p_1$ and $p_2$, respectively, $k_{SM}$ a spectral-mixture kernel and $k_{SE}$ a squared-exponential kernel. Note that the former 3 kernels only depend on the first variable which correspond to time, whereas the SE-kernel depends on all variables, thus models the influence of the additional variables. With our CPoE model it is straightforward to handle time series with covariates, as opposed to other time series methods \cite{benavolistate, corani2021time, sarkka2013spatiotemporal, 
hyndman2018forecasting}.
%
%
The 
kernel  $k_{\bthe}$ depends on several hyperparameters $\bthe$ for which we use the parametrization in \cite{corani2021time}. 
We assume a log-normal prior 
on  $\bthe$
as described in 
Section \ref{se:hyperEst} 
	in which the corresponding means and variances are taken from Table 1 in \cite{corani2021time}.
We demonstrate the MAP estimation for $\bthe$
on
 the \textit{elecdemand} time series (\cite{fpp2}, Table \ref{tab:datasets}) which contains the electricity demand as response $y$ together with the time as the first variable $X_1$, the the corresponding temperature as $X_2$ and the variable whether it is a working day as $X_3$
which 
is depicted in the  plots in  Fig.\ \ref{fig:timeSeries} on the left,
where we shifted the first and third variable in the second plot for the sake of clarity.
Similarly as in the previous section, we run full GP, SGP, PoEs and CPoE and optimized the hyperparameter deterministically using the MAP as objective function taking into account the priors. 
The results are provided
in Table 
\ref{tab:KL_ELEC} and  in Fig.\ \ref{fig:timeSeries} on the right, which again show very competitive performance also for a general kernel with priors on the hyperparameters. 
A complete description of the experiment is given in
Section 
\ref{se:appplication2} in the Appendix.
%


\section{Conclusion}
\label{se:conclusion}

In this paper we introduce a novel GP approximation algorithm CPoE where the degree of approximation can be adjusted by a locality and a sparsity parameter so that the proposed method recovers independent PoEs, sparse global GP and full GP.  
%
We show that our method consistently approximates full GP, in particular,
we proved that increasing the correlations between the experts 
decreases monotonically the 
KL of the joint prior of full GP to them of our model.
%
The presented algorithm has only a few hyperparameters which allows an efficient deterministic and stochastic optimization.  Further, our presented algorithm
 works with a general kernel, with several variables and also priors on the hyperparameters can be included. 
Moreover, the time and space complexity 
is linear in the number of experts and number of data samples, which makes it highly scalable. This is demonstrated with efficient implementations so that a dataset with several ten thousands of samples can be processed in around a minute on a standard Laptop.
%
%
In several experiments with synthetic and real world data, superior performance  in a accuracy vs. time sense compared to state-of-the-art GP approximations methods is demonstrated for the deterministic and stochastic case which makes our algorithm a competitive method for GPs approximations.
%
%
%

Our approach could be enhanced in several directions.  The first improvement would be more practical. While the current implementation of our algorithm works very competitively for moderate large datasets (on a standard Laptop), further work has been done to scale it up to very large datasets. The current limitations are particularly  factorizing and solving the sparse block Cholesky matrices.
We are convinced, that the theoretical properties of our algorithm 
 - in particular the linearity in the number of experts and data samples - 
 enables large scale implementations  when exploiting more low level linear algebra tools.
%
Another interesting direction would be to investigate the connection of our sparse precision matrix 
to
state space systems such that sequential learning algorithm could be exploited which might be interesting for an online version of this algorithm which is briefly outlined in \ref{se:stateSpace}. Together with the competitive results in the application to time series with covariates 
makes this idea very promising.
Further, it would be interesting to apply variational methods to our model so that a connection to full GP in a posterior sense might be established where
some ideas are outlined in 
\ref{se:variational}.
%
%
%
%
%
%

	\paragraph{Acknowledgements} This work is supported by the Swiss National Research Programme 75 "Big Data" (NRP 75) with grant number 167199.
	
\bibliographystyle{plain} 
	\bibliography{ref}
	
	\appendix

%
%

  \begin{table}
  \begin{footnotesize}
\centering
  \begin{tabular}{ l | c | c }
    \toprule
     variable & domain & explanation \\
    \midrule
     $N$ & $\mathbb{N}^+$ & number of data samples \\
     $D$ & $\mathbb{N}^+$  & number of variables/dimension of data \\
     $J$ & $\{1,\ldots,N\}$ & number of experts/partitions \\
     $B$ & $\{1,\ldots,N\}$ & size of expert/partition \\
     $L$ & $\{1,\ldots,B\}$ & number of local inducing points\\
     $M$ & $\{1,\ldots,N\}$ & number of total local inducing points \\
     $C$ & $\{1,\ldots,J\}$ & degree of correlation \\
     $\gamma$ &  $(0,1]$ & sparsity parameter \\
     $\alpha$ &  $(0,J]$ & approximation quality parameter \\
     $y_i$ &  $\RR$ & individual data output \\
     $\by_j$ &  $\RR^{B}$ & data output of expert $j$ \\
     $\by_{k:j}$ &  $\RR^{B(j-k1+)}$ & data output of experts $k$ up to $j$ \\
     $y_*$ &  $\RR$ & pointwise noisy prediction output \\
     $\by$ &  $\RR^{N}$ & all data output \\
     $\bx_i$ &  $\RR^{D}$ & individual data indput \\
     $\bX_j$ &  $\RR^{B \times D}$ & data input of expert $j$ \\
     $\bX_{k:j}$ &  $\RR^{B(j-k+1) \times D}$ & data input of experts $k$ up to $j$ \\
     $\bX$ &  $\RR^{N \times D}$ & all data input \\
     $\bx_*$ &  $\RR^D$ & query input for prediction \\
     $\bff_j$ &  $\RR^{B}$ & latent function outputs of expert $j$ \\
     $\bff_{k:j}$ &  $\RR^{B(j-k+1)}$ & latent function outputs of experts $k$ up to $j$ \\
     $\bff$ &  $\RR^{N}$ & all latent function outputs \\
     $f_*$ &  $\RR$ & pointwise (latent) prediction output \\
     $f(\bX_j)$ &  $\RR^{B}$ & GP evaluation for input matrix  \\
     $\ba_j$ &  $\RR^{L}$ & local inducing outputs of expert $j$ \\
     $\ba_{k:j}$ &  $\RR^{L(j-k+1)}$ & local inducing outputs of experts $k$ up to $j$ \\
     $\ba$ &  $\RR^{M}$ & all local inducing outputs \\
     $\bA_j$ &  $\RR^{L\times D}$ & local inducing inputs of expert $j$ \\
     $\bA_{k:j}$ &  $\RR^{L(j-k+1)\times D}$ & local inducing inputs of experts $k$ up to $j$ \\
     $\bA$ &  $\RR^{M\times D}$ & all local inducing inputs \\
     $I_j$ & $\{1,\ldots,C-1\}$ & number of predecessors of expert $j$ \\
     $\bs{\phi}_i(j)$ &  $\{1,\ldots,j-1\}$ & $i$th predecessor of expert $j$ \\
     $\bs{\pi}(j)$ &  $\{1,\ldots,j-1\}^{I_j}$ & predecessor index set \\
     $\bs{\pi}^+(j)$ &  $\{1,\ldots,j\}^{I_j+1}$ & predecessor index set including $j$ \\
     $\bs{\psi}(j)$ & $\{1,\ldots,max(j,C)\}^{C}$ & correlation index set \\
     $\sigma_n^2$ & $\RR^+$ & observation noise variance  \\
     $\bthe$ & $\RR^{\vert \bthe \vert}$ & kernel hyperparameters including $\sigma_n^2$ \\
          $k_{\bthe}(\bx,\bx')$ & $\RR$ & kernel evaluation for 2 query points\\
     $\bK_{\bA\bB}$ & $\RR^{M_a \times M_b}$ & kernel matrix of two query matrices 
     \\
     $p(\bz)$ & $\RR$ & evaluation of (true) probability density   \\
     $q(\bz)$ & $\RR$ &  evaluation of approximated probability density  \\
     $\bS$ & $\RR^{M\times M}$ & prior precision matrix\\
     $\bT$ & $\RR^{M\times M}$ & projection precision matrix\\
     $\bSig^{-1}$ & $\RR^{M\times M}$ & posterior precision matrix\\
     $\bSig$ & $\RR^{M\times M}$ & posterior covariance matrix\\
     $\bmu$ & $\RR^{M}$ & posterior mean vector\\
     $\bmu_{\bs{\psi}(j)}$ & $\RR^{CL}$ & local posterior mean\\
     $\bSig_{\bs{\psi}(j)}$ & $\RR^{CL\times CL}$ & local posterior covariance\\
     $\bF$ & $\RR^{M\times M}$ & prior transition matrix\\
     $\bQ$ & $\RR^{M\times M}$ & prior noise matrix\\
     $\bH$ & $\RR^{N\times M}$ & projection matrix\\
     $\overline{\bV}$ & $\RR^{N\times N}$ &  projection noise matrix\\
     $\bV$ & $\RR^{N\times N}$ &  projection noise matrix including observation noise\\
     $\bP$ & $\RR^{N\times N}$ &  marginal likelihood covariance matrix\\
     $J_2$ & $\mathbb{N}^+$ & number of prediction experts\\
     $\bar{\beta}_{*j}$ & $\RR^+$ & unnormalized predictive weight  of expert $j$ at $\bx_*$\\
     $\beta_{*j}$ & $\RR^+$ & normalized predictive weight  of expert $j$ at $\bx_*$\\
     $m_{*j}$ & $\RR$ & predictive mean of expert $j$ at $\bx_*$\\
     $v_{*j}$ & $\RR^+$ & predictive variance of expert $j$ at $\bx_*$\\
    $\mathbb{D}_{[C,C_2]}$ & $\RR^+$ & difference in KL between two approximate models\\
    \bottomrule
  \end{tabular}
     \caption{Overview of notation.}
   \label{tab:notation}
    \end{footnotesize}
  \end{table}

\section{Extensions and Details}

\subsection{Generalized CPoE }
\label{se:variational}

Alternatively to the graphical model defined in Def.\ \ref{def:jointDistr} 
(and more precisely in 
Prop.\ \ref{prop:graphicalModel} with Proof \ref{proof:graphicalModel})
which recovers sparse global GP model
FITC \cite{snelson2006sparse} in the limiting case $C\rightarrow J$ (as shown in Prop.\ \ref{prop:equality}),
we present in this section 
 a generalization of our CPoE model such that it recovers other sparse global GP models such as VFE \cite{titsias2009variational} or PEP \cite{bui2016unifying}.
As shown by the authors in \cite{schurch2020recursive} for the global case, these model differ in the training only by the choice of the projection matrix 
$\overline{\bV}_{j}$  in Def.\ \ref{def:jointDistr} and in the hyperparameter optimization by a modification of the log marginal likelihood $\mathcal{L}(\bthe)
=
\log \qc{\by}{\bthe}$
in Section \ref{se:hyperEst}. These two changes can also be made for our local sparse CPoE model.
In particular,  using  
$\overline{\bV}_{j}$ and $ \lambda_j$
according to the values in Table \ref{tab:generalizations} in
the projection conditional 
 $$
 \pc{\bff_j}{\ba_{\psB{j}}} 
=
\NN{\bff_j}{\bH_{j}\ba_{\psB{j}},\overline{\bV}_{j}} $$
and 
in  a lower bound to the log marginal likelihood
$$
\tilde{\mathcal{L}}(\bthe)  = \mathcal{L}(\bthe) - \sum_{j=1}^J \lambda_j(\bthe)
$$
and 
$
\tilde{l}_j(\bthe)  = l_j(\bthe) -  \lambda_j(\bthe)
$
in the deterministic and 
stochastic case, respectively,
generalizes the CPoE method such that for $C\rightarrow J$ we recover the mentioned method global methods in Table \ref{tab:generalizations}. Thereby,  we used 
$$\bD_{j}
=
\bK_{\bX_j\bX_j}  -
\bK_{\bX_j\bA_{\psB{j}}} \bK_{\bA_{\psB{j}}\bA_{\psB{j}}}^{-1} \bK_{\bA_{\psB{j}}\bX_j}$$
which is the difference of the true and local approximated covariance.

\begin{table}
\begin{center}

\begin{tabular}{l | c c}
\toprule
 & $\overline{\bV}_{j}$ & $\lambda_j$ \\
\midrule
DTC &  0 & 0 \\
FITC &  $Diag[\bD_{j}]$ & 0 \\
PITC &  $\bD_{j}$ & 0 \\
VFE &  0 & $\frac{1}{2\sigma_n^2} tr[\bD_{j}]$ \\
PEP &  $\alpha Diag[\bD_{j}]$ & 
$\frac{1-\alpha}{2\alpha}\sum_i \log\left(1 + \frac{\alpha}{\sigma_n^2}\bD_j^{(i)} \right)$
 \\
 PEP$_B$ &  $\alpha\bD_{j}$ & 
$\frac{1-\alpha}{2\alpha}\sum_i \log\vert \II + \frac{\alpha}{\sigma_n^2}\bD_j \vert$
 \\
\end{tabular}
\caption{Generalizations of CPoE model. }
\label{tab:generalizations}
\end{center}
\end{table}

The setting in VFE \cite{titsias2009variational} is particularly interesting, since it constitutes in the global case a direct posterior approximation derived via a variational maximization of the lower bound of the log marginal likelihood. 
Moving a bit away from the true marginal likelihood 
of full GP has the effect that  overfitting (w.r.t.\ full GP) can not happen when optimizing the hyperparameters with the lower bound.
This is particularly important when all inducing inputs are optimized as it is usually recommended in sparse global methods which is not the case for our model since it allows to have a number of inducing points in the order of the number of data samples. In the adapted 'local VFE' CPoE model when using $\overline{\bV}_{j}=0$ and minimize also $ \lambda_j=tr\{\bD_{j}\}$ has the effect that the model is locally variationally optimal, 
however, it would be interesting to  directly derive  a lower bound analogously to \cite{titsias2009variational}  so that the posterior of our CPoE model is rigorously connected to full GP.
Since this is not a straight-forward extension, we 
we postpone this task to future work. Below, we present the connection to full GP for this adapted model in the joint prior sense analogously to Prop.\ \ref{prop:convJointPrior} for the local FITC model.

%
%
\begin{proposition}[Local VFE]
\label{prop:convPriorVFE}
Using a deterministic projection $q(\bff_j \vert \ba_{\psi(j)})=\NN{\bff_j }{\bH_j \ba_{\psi(j)}, 
\overline{\bV}_j}$ in the graphical model  in Def.\ \ref{def:jointDistr} 
and
Prop.\ \ref{prop:graphicalModel}, that is,
setting the covariance $\overline{\bV}_j=0$ in the projection step recovers global VFE for $C\rightarrow J$.
Moreover, the difference in KL to full GP of the joint prior is also decreasing.
In particular,
the difference in KL 
of the prior of the local VFE model
for
 $1\leq C\leq C_2\leq J$ is
\begin{align*}
\mathbb{D}_{(C,C_2)}[\ba]
%
%
%
%
=\frac{1}{2}
\log
\frac{\vert\bQ_{C}\vert}{\vert\bQ_{C_2}\vert}
\geq 0.
\end{align*}
Further, 
the difference in KL of the projection is 
\begin{align*}
\mathbb{D}_{(C,C_2)}[\by\vert \ba]
=
\frac{1}{2 \sigma_n^2}
tr\{\bar{\bV}_{C} - \bar{\bV}_{C_2}\}
\geq 0.
\end{align*}
The overall prior approximation quality is
\begin{align*}
\mathbb{D}_{(C,C_2)}[\ba, \by] = \frac{1}{2}
\log
\frac{  \vert\bQ_{C}\vert
}{
 \vert\bQ_{C_2}\vert 
}
+
\frac{1}{2 \sigma_n^2}
tr\{\bar{\bV}_{C} - \bar{\bV}_{C_2}\}
\end{align*}
where
$$tr\{\bar{\bV}_{C}\}
=\sum_{i=1}^N 
K_{\bX_i\bX_i} -
K_{\bX_i\bA_{\bs{\psi}(j_i)}}
K_{\bA_{\bs{\psi}(j_i)} \bA_{\bs{\psi}(j_i)}}^{-1}
K_{\bA_{\bs{\psi}(j_i)}\bX_i}.
$$
\end{proposition}

Compared to the FITC model is the difference in the trace instead of the fraction of the log-determinants.
%
%
%
%
%


\subsection{Solving Linear System \& Partial Inversion}
\label{se:systemAndInv}


For solving the sparse linear system 
$
\bSig^{-1}\bmu = \bb
$
in 
Prop.\ \ref{prop:posterior},
 sparse Cholesky decomposition is exploited, that is,
$\bM \bSig^{-1} \bM^T  = \bL \bL^T =:\ \bs{Y}$ is computed 
so that  $\bs{\nu}$ and 
$\overline{\bmu} $ can be efficiently obtained via solving
$\bL \bs{\nu} = \bb$ and $\bL^T \overline{\bmu} = \bs{\nu}$, respectively,
where $\bM$ is a so-called fill-reduction permutation matrix such that the Cholesky matrix $\bL$ is as sparse as possible
and thus $\bmu = \bM^{-1}\overline{\bmu}$. Note that
$\bM$ is computed only via the structure on the block level which is only $J$ dimensional instead of $JL$.
\\
Additionally to the mean $\bmu$,  
also some entries  $\bSig_{\psB{j}}$ in the covariance matrix $\bSig $ has to be explicitly computed which are needed for computing local predictions 
(Section \ref{se:prediction})
and (derivatives) of the marginal likelihood
(Section \ref{se:derivatives}), respectively. The needed entries correspond to the non-zeros in the precision 
matrix 
$\bSig^{-1}$. 
Computing efficiently these entries is not straightforward since in the inverse the blocks are no longer independent. However, we can exploit the particular sparsity  and block-structure of our precision matrix
and obtain an efficient implementation of this part which is key to achieve a competitive performance of our algorithm.
\\
Computing some entries in $\bs{Z}=\bs{Y}^{-1}$ is also known as \textit{partial inversion}. We adapted the approach in
\cite{takahashi1973formation} 
where 
 the recursive equations with $J$ blocks for computing the full inverse $\bs{Z}$ are provided
\begin{align*}
\bs{Z}_{B_j}  =
-\bs{Z}_{C_j} \bL_{B_j} \bs{L}_{A_j}^{-1}
\quad\quad\text{and}\quad\quad
%
\bs{Z}_{A_j} =
\bL_{A_j}^{-T}\bL_{A_j}^{-1}
-\bs{Z}_{B_j}^T \bL_{B_j}\bL_{A_j}^{-1}
\end{align*}
where the recursion starts from $j=J$ with 
$\bs{Z}_{A_J} =
\bL_{A_J}^{-T}\bL_{A_J}^{-1}$.
 \begin{figure}[htp!]
	\centering
  \includegraphics[width=.75\textwidth]{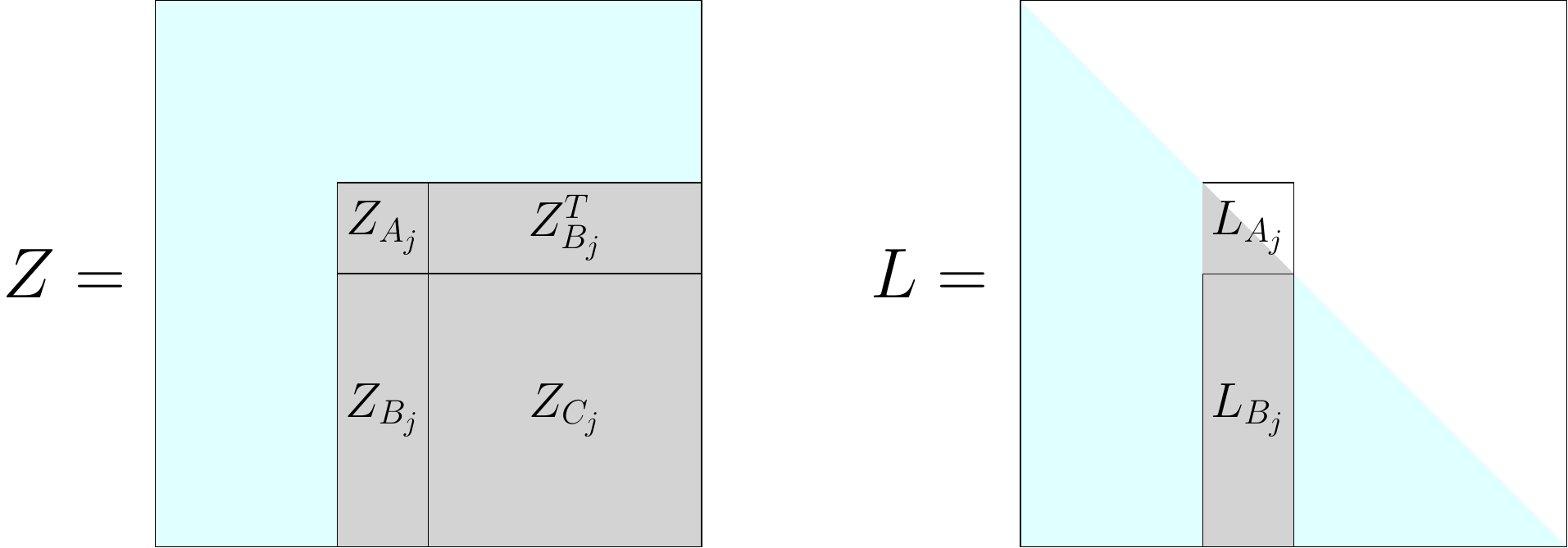}
\end{figure}
\\
Instead of computing the full inverse using this recursion, we exploited the block-sparsity structure of our posterior precision matrix in order to gain significant speed-up.
 We only computed the entries in the inverse $\bs{Z}$ which are symbolically non-zero in $\bL$. In Algorithm \ref{alg:alg1} in the Appendix we provide efficient pseudo-code using sparse-block-matrices in the block-sparse-row format.

\begin{algorithm}
\caption{Partial Sparse Block Inversion}\label{alg:cap}
\label{alg:alg1}
\begin{algorithmic}
\Require Cholesky matrix $\bL$ of size $JB\times JB$ in block-sparse-row (bsr) format with $J\times J$ total blocks, block-size $B$ and $N$ non-zero blocks. Data array $d$ of size $N\times B\times B$, the column-block-indices $r$ of size $N$, row-block-pointer $p$ of length $J+1$, and lookup table $M$ of dimension $J\times J$.
\Ensure The lower part of the symmetric partial inversion is computed in bsr-format with the same row-block-indices $r$  and  row-block-pointer $p$ and data array $z$ of size $N\times B\times B$.
\For{$i \in \{J,\ldots,1 \}$} 
\State $L_{A}^{-1} \gets d[M[i,i], :, :]^{-1}$
\State $z[M[i,i]] \gets (L_{A}^{-1})^T\cdot  L_{A}^{-1}$
\For{$j \in \{r[p[i+1]], r[p[i+1]-1], \ldots,r[p[i]] \}$} 
\State $Q \gets 0$
\For{$l \in \{r[p[i]], r[p[i]+1], \ldots,r[p[i+1]] \}$} 
\State $R \gets z[M[j,l]]$
\If{$l > j$}
 \State $R \gets R^T$
\EndIf
\State $Q \gets Q + R \cdot d[M[l,i]] \cdot L_{A}^{-1}$
\EndFor
\State $z[M[i,j]] \gets z[M[i,j]] - Q$
\EndFor
\EndFor
\end{algorithmic}
\end{algorithm}

Alternatively for computing the Cholesky factor of $\bSig^{-1} = \bS + \bH \bV^{-1} \bH$, we could directly exploit that the prior precision $\bS = \bF^T \bQ^{-1} \bF$ is already decomposed into a upper/lower-triangular form  since $\bF$ lower triangular. However, when updating the Cholesky factor with $\bH \bV^{-1} \bH$ needs quadratic time in the number of nonzeros for each expert.

\subsection{Hyperparameter Estimation}
\label{se:hyperEst2}

In Section \ref{se:CPoE},
we introduced CPoE for fixed hyperparameters $\bthe$ where implicitly all distributions are conditioned on  $\bthe$, however, we omitted the dependencies on $\bthe$ in the most cases for the sake of brevity. Similar to full GP, 
sparse GP or PoEs, the 
\textit{log  marginal likelihood (LML)}
can be used as an objective function for optimizing the few hyperparameters $\bthe$.
\subsubsection{Deterministic Optimization}
The 
log of the marginal likelihood 
of our model 
formulated in Section \ref{se:inference}
can be written as
%
%
%
$$
\mathcal{L}(\bthe)
=
\log \qc{\by}{\bthe}   = \log  \NNo{\bO,\bP}
=
-\frac{1}{2}\left(
\by^T\bP^{-1}\by
+\log\vert\bP\vert
+N \log 2\pi \right)$$
  with
$\bP=\bH\bS^{-1}\bH^T +\bV$.
Since $\bP$ is dense, we can apply
the inversion \eqref{eq:matrix_inv_lemma} and determinant lemma \eqref{eq:det_lemma} to $\bP$ and exploit 
$\vert \bF \vert = 1$ yielding
\begin{align}
\begin{split}
\label{eq:margLikDet}
\mathcal{L}(\bthe)
&=
-\frac{1}{2}\left(
\by^T\bV^{-1}\by -\bmu^T\bSig^{-1}\bmu
+\log
\frac{ 
\vert\bSig^{-1}\vert~  \vert\bV\vert 
}{\vert\bS\vert}
+N \log 2\pi
\right)
\\
&=
-\frac{1}{2}\left(
\by^T\bV^{-1}\by -\bmu^T\bSig^{-1}\bmu
+\log
\vert\bSig^{-1}\vert~  \vert\bV\vert ~  \vert\bQ\vert
+N \log 2\pi
\right)
\end{split}
\end{align}
so that all involved quantities  $\bSig^{-1}$, $\bV$ and $\bQ$ are sparse. 
%
%
For efficient parameter minimization, the derivative of the log marginal likelihood with respect to each parameter in $\bthe$ is needed for which the  derivations are provided in Appendix \ref{se:derivatives}. Thereby also some parts of the covariance matrix $\bSig$ are needed which is explained in Section \ref{se:systemAndInv}. 
Alternatively to the marginal likelihood, we can maximize a lower bound of it which is a generalization of  our model so that we recover a range of well known sparse global GP models for $C\rightarrow J$ as discussed is Section \ref{se:variational}.
\cite{schurch2020recursive, bui2016unifying}.

\begin{figure*}[htp!]
	\centering
  \includegraphics[width=.99\textwidth]{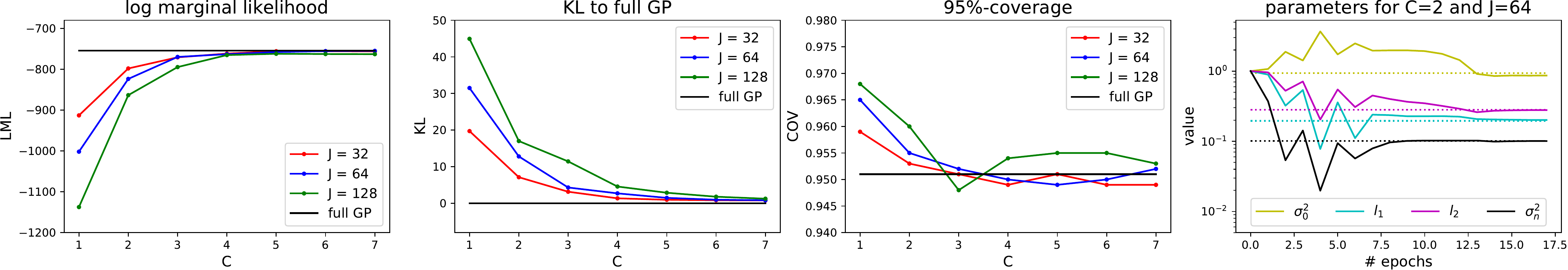}
	\caption{  Convergence of deterministic batch hyperparameter optimization 
	for increasing $C$
	and 
	trace of the parameters (solid) compared to the optimal values of full GP (dotted). }
	\label{fig:batchConv}
\end{figure*}

For moderate sample size $N$,
\textit{deterministic optimization} with full batch $\by$ can be performed. That means,  the log marginal likelihood for the whole data is computed for which the sparse system of equations with the sparse  posterior precision as well as  the partial inversion of the posterior covariance has to be solved.
In particular,  the  functions for computing 
$\mathcal{L}(\bthe)$
 and 
  $\frac{\partial\mathcal{L}(\bthe) }{\partial \bthe}$ 
 for each $\bthe$ and full data $\by$ are repetitively called by a numerical minimizer.
Fig.\ \ref{fig:batchConv} illustrates the performance of this  deterministic batch hyperparameter optimization where the convergence  for the log marginal likelihood, average KL divergence, $95\%$-coverage
(both quantities exactly defined in Appendix \ref{se:tables}) for different number of experts $J$  compared to full GP are depicted.
The $N=2048$ data samples are generated with a $D=2$-dimensional SE-kernel and the test KL and coverage mean values are reported for $N_{test}=1000$ samples with $5$ repetitions. We used $\gamma=1$ and $C\in\{1,\ldots,7\}$.
We observe that the log marginal likelihood and KL are getting better for increasing $C$, and the deterministic parameter estimates converge to the ones of full GP for increasing function calls. It is interesting to observe that also for smaller $C$ values, the coverage of our methods are consistent. In particular, they are slightly too big, meaning our confidence information are conservative. This is due to the aggregation based on the covariance intersection method with normalized weights, which guarantees consistent second order information.\\
\subsubsection{Stochastic Optimization}
\begin{figure}[htp!]
	\centering
  \includegraphics[width=.8\textwidth]{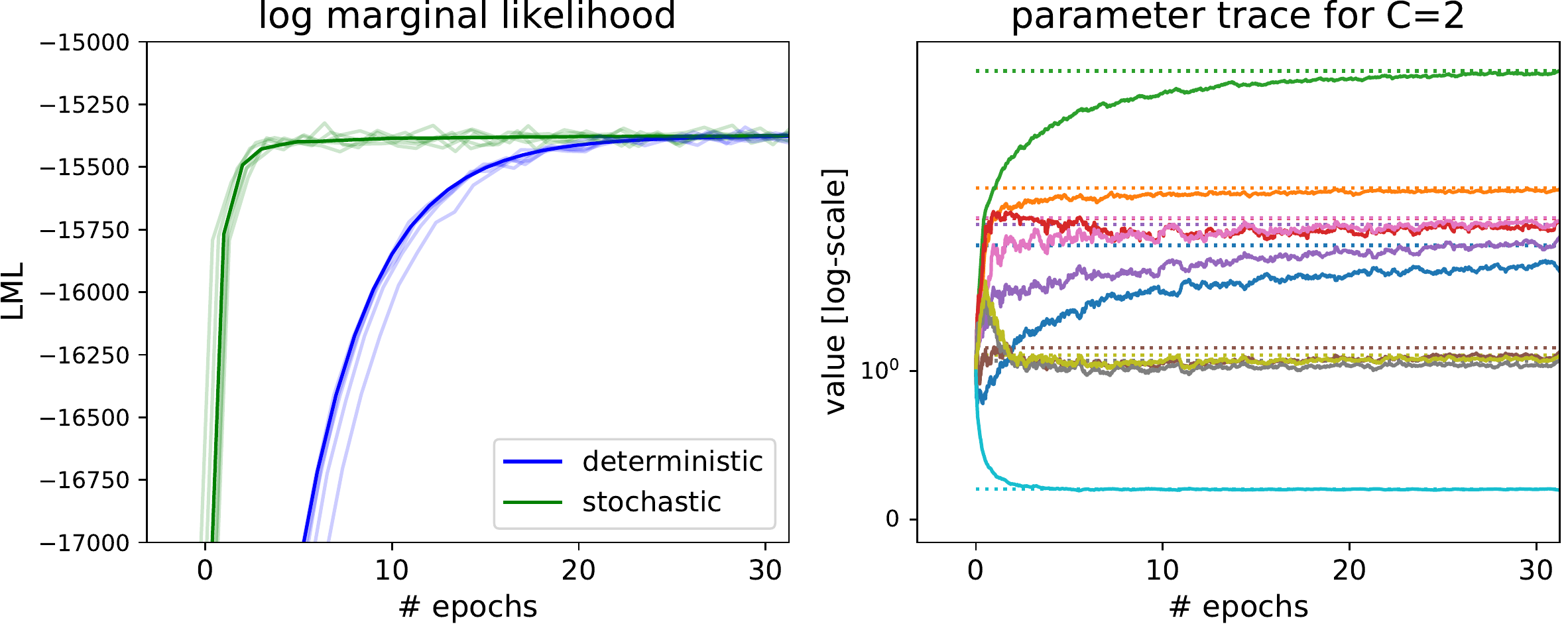}
	\caption{  Convergence of stochastic vs. deterministic  hyperparameter optimization of our model CPoE.
%
This
experiment 
compares the convergence 
 of stochastic vs. deterministic  hyperparameter optimization for the log marginal likelihood and the trace of the $10$ parameters $\bthe_j$ for the dataset \textit{cadata} with $N_{tot}=20640$ and $D=8$. We used $5$ different splits with  $N=0.9N_{tot}$ training data and the rest for testing. The  values for our algorithm are $C=2$, $J=64$, $\gamma=1$ and learning rate $\delta=0.01$. In the right plot, the dotted horizontal lines  and the solid traces correspond to the final deterministic value and the current stochastic values, respectively. We note that the stochastic LML and trace of hyperparameters converge faster to a very similar value as in the deterministic case. 
%
}
\label{fig:stochConv}
\end{figure}

The presented method in the previous section works fine for small datasets, 
however,
in order to scale this parameter optimization part to larger number of samples $N$ in a competitive time, \textit{stochastic optimization} techniques has to be exploited  similarly done  for the global sparse GP model (SVI \cite{hensman2013gaussian}; REC \cite{schurch2020recursive}; IF \cite{kaniasparse}). In the approximation method REC \cite{schurch2020recursive}, the recursive derivatives are exactly propagated which would also be possible for our model, however, it turned out that in practice the differences in accuracy are very small when using instead the hybrid approach IF of
\cite{kaniasparse}. Thereby, the independent factorization of the log marginal  likelihood is used for the  computations of the optimization part, whereas the exact posterior is used for inference and prediction.
Adapted to our setting, the independent factorized log marginal likelihood 
$
\log \qc{\by}{\bthe}$
can be approximated by
\begin{align*}
\log \qc{\by}{\bthe}
&\approx
\log \prod_{j=1}^J \int \qc{\by_j}{\ba_{j}} \q{\ba_{j}} \diff \ba_{j}\\
&=
\log \prod_{j=1}^J \int \NNo{\bH_j\ba_j,\bV_j}\NNo{\bO,\bS_j^{-1}}  \diff \ba_{j}
\\
&=
\sum_{j=1}^J \log  \NNo{\bO,\bP_j}
=:
\tilde{\mathcal{L}}(\bthe)
\end{align*}
%
%
%
where
$\bP_j=\bH_j\bS_j^{-1}\bH_j^T +\bV_j$ with
  $\bS_j=\bK_{\bA_{j}\bA_{j}}^{-1}$.
 The difference compared to 
 the deterministic case 
 in \eqref{eq:margLikDet}
 and to \cite{kaniasparse} for the global sparse model is the 
independent prior
$ \q{\ba_{j}}$
instead of 
$ \q{\ba}$ and 
$ \pp{\ba}$,
  respectively.
 In the approximate case, we can write
%
%
%
$$
\tilde{\mathcal{L}}(\bthe)
=
-\frac{1}{2}N \log (2\pi)  
 +
\sum_{j=1}^J l_j(\bthe)
$$ with
$
l_j(\bthe)
=
-\frac{1}{2}
\left(
\by_j^T\bP_j^{-1}\by_j
+\log\vert\bP_j\vert
\right)
$
which has
the advantage 
that it decomposes
into the $J$ terms $l_j$ in the sum, so that it can be used for stochastic optimization.
This constitutes a very fast and accurate alternative for our method as shown in 
Figure \ref{fig:stochConv}
and 
is
exploited in Section \ref{se:examples_evaluation} for large data sets.
\\ \quad \\
%
%

%

 \subsection{Complexity}
 \label{se:complexity2}
%

\begin{wrapfigure}{r}{0.35\linewidth}
 \vspace{-10pt}
\centering
    \includegraphics[width=1\linewidth]{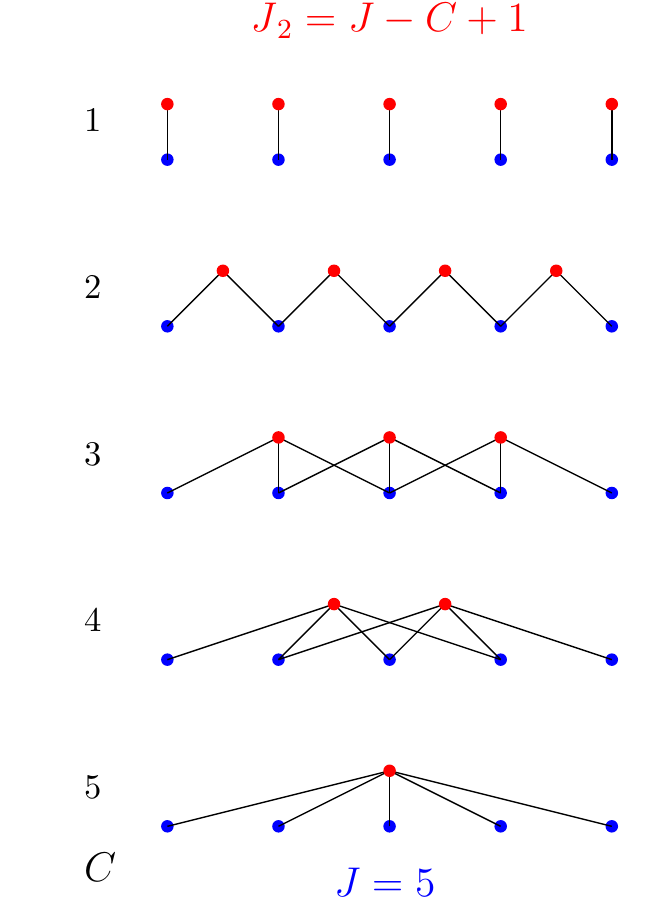}
\caption{Prediction Aggregation in CPoE($C,\gamma$) model with  $J$ base experts and $J_2 = J-C+1$ predictive experts. }
 \label{fig:predAgg}
	 \vspace{-10pt}
\end{wrapfigure}

The time complexity for computing the posterior and the marginal likelihood in our algorithm is dominated by $J$ operations which are cubic in $LC$ (inversion, matrix-matrix multiplication, determinants).
This leads to $\mathcal{O}(J(LC)^3) = \mathcal{O}(J(BC\gamma)^3)
= \mathcal{O}(NB^2\alpha^3)$ where we define the approximation quality parameter $\alpha=C\gamma$. Similarly for the needed space  
$\mathcal{O}(J(LC)^2) = \mathcal{O}(J(BC\gamma)^2)
= \mathcal{O}(NB\alpha^2)$.
For $N_t$ testing points, the time 
for (pointwise) predictions is dominated by $J$ inversions of matrices with dimension $LC$ and matrix multiplications with dimensions $LC\times LC\times N_t$ leading to
$\mathcal{O}(J(LC)^3 +  J(LC)^2 N_t  ) =
\mathcal{O}(J(B\gamma C)^3 + J (B\gamma C)^2 N_t  ) =
\mathcal{O}(N B^2\alpha^3 + N B \alpha^2 N_t  )
 $ where the operations independent of the test points can be precomputed in the inference part leading to $\mathcal{O}(N B\alpha^2 N_t)$ for testing. Similarly for the space. A further reduction in complexity would be achieved if the product over all experts in Prop.\ \ref{prop:predAgg} is approximated only with the $W < J$  nearest experts, leading to 
 $\mathcal{O}((LC)^2 N_t W)
 = 
 \mathcal{O}((B\gamma C)^2 N_t W)
 =\mathcal{O}(N\frac{W}{J} B\alpha^2 N_t)$ 
  time complexity for testing. This might be interesting if we want to make fast predictions for many points $N_t$. For reasonable values  of $W$, for instance  $W=1$, $W=C$ or $W=Z=\log(N)C$ (used in prediction aggregation), preliminary experiments show very comparable  performance. Note that the consistency properties for covariance intersection method are preserved as long as the weights are normalized over the used $W$ experts.
Table \ref{tab:complexity} compares the asymptotic complexities with other GP algorithms.

It is interesting that for $\alpha= 1$, our algorithm has the same asymptotic complexity for training as sparse global GP with $M_g=B$ global inducing points but we can have $M_{l}=LJ=\gamma BJ=\gamma N$ total local inducing points! Thus,
	our approach allows much
more total local inducing points $M$ in the order of $N$ (e.g. $M=0.5N$ with $C=2$) whereas for sparse global GP usually $M_{g} \ll N$. This  has the consequence that the local inducing points can cover the input space much better and therefore represent much more complicated functions. As a consequence,  there is also no need to optimize the local inducing points resulting in much fewer parameters to optimize. Consider the following example with $N=10'000$ in $D=10$ dimensions. Suppose a sparse global GP model with $M_g=500$ global inducing points. A CPoE model with the same asymptotic complexity has a batch size $B=M_g=500$ and $\alpha=1$. Therefore, we have $J=\frac{N}{B}=20$ experts and we choose $C=2$ and $\gamma=\frac{1}{2}$ such that we obtain $L=\gamma B=250$ local inducing points per experts and $M = \gamma N = 5'000$ total inducing points! Further, the number of hyperparameters to optimize for a SE kernel is for global sparse GP $M_g D+\vert\bthe\vert = 5012$, whereas for CPoE there are only $\vert\bthe\vert = 12$.
\\
For our method, the time and space complexity is linear in the number of samples $N$ and the number of experts  $J$  which makes our approach highly scalable. 
The approximation quality parameter $\alpha=C\gamma$ appears cubic/quadratic in the time/space complexity. The optimal approximation quality (and thus equivalent to full GP) is achieved for $\alpha=J$ which implies $C=J$ and $\gamma=1$. However, it is clear that this is not feasible for big datasets and thus some moderate values of $C$ and $\gamma$ have to be selected to trade off time and accuracy which is illustrated in the Appendix in Table \ref{tab:timesPp} and Fig.\ \ref{fig:covsPp}.

 \begin{figure}[htp!]
	\centering
  \includegraphics[width=.75\textwidth]{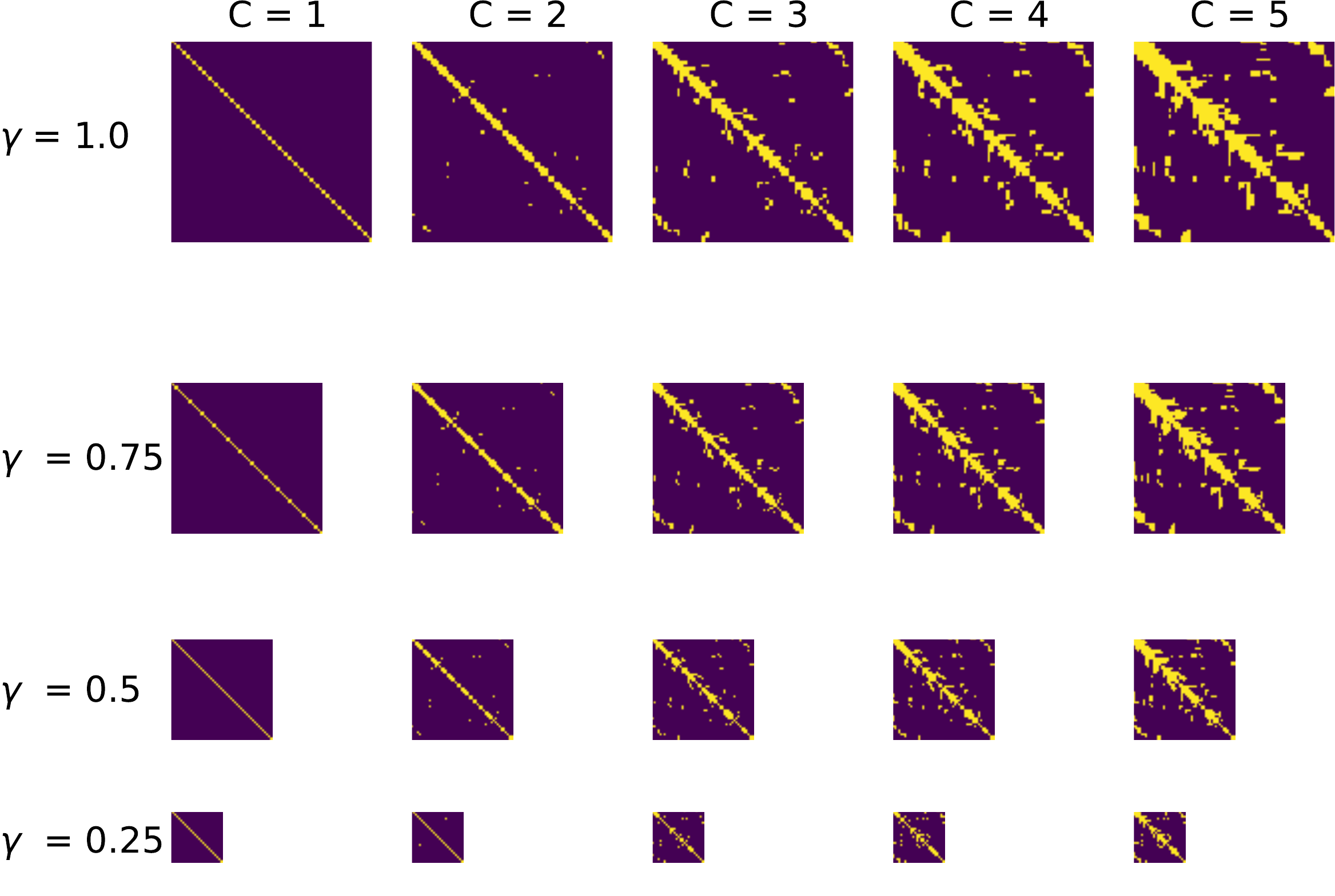}
	\caption{ Influence of the approximation order $C$ and sparseness parameter $\gamma$ to the non-zeros and size of the posterior precision matrix $\bSig^{-1}$.
	 for an example with synthetic GP data with $D=2$, $N=8192$, $J=64$ and $B=128$ and a SE-kernel.
	 Compare also Table \ref{tab:timesPp}. 
	}
	\label{fig:covsPp}
\end{figure}

 \begin{table}
      \begin{small}
  \begin{center}
  \begin{tabular}{ l | c  c c c r }
     \toprule
     \textbf{KL} & C=1 & C=2 & C=3 & C=4 & C=5  \\ \midrule
    $\gamma$ = 1/4 & 12.3 & 5.0 & 1.3& 0.9 & 0.7 \\ 
    $\gamma$ = 1/2 & 12.2 & 4.9 & 1.0 & 0.8 & 0.6 \\
    $\gamma$ = 3/4 & 12.1 & 4.9 & 0.9 & 0.7 & 0.5 \\
    $\gamma$ = 1 & 12.1 & 4.8 & 0.9 & 0.6 & 0.4 \\
    \bottomrule
    \toprule
    \textbf{time} & C=1 & C=2 & C=3 & C=4 & C=5  \\ \midrule
    $\gamma$ = 1/4  & 0.2 & 0.4 & 0.9 & 1.2 & 1.4 \\ 
    $\gamma$ = 1/2 & 0.4 & 0.7 & 1.9 & 2.7 & 3.8 \\
    $\gamma$ = 3/4 & 0.9 & 2.4 & 4.1 & 5.7 & 9.1 \\
    $\gamma$ = 1  & 1.5 & 3.0 & 6.4 & 12.4 & 15.7\\
    \bottomrule
  \end{tabular}
  \caption{KLs to full GP (above) and times (below) of our method CPoE for varying $C$ and $\gamma$ for experiment in Section \ref{se:complexity}. Compare also Fig.\ \ref{fig:covsPp} .
 }
  \label{tab:timesPp}
  \end{center}
  \end{small}    
  \end{table}

\begin{figure}[bp!]
	\centering
  \includegraphics[width=.75\textwidth]{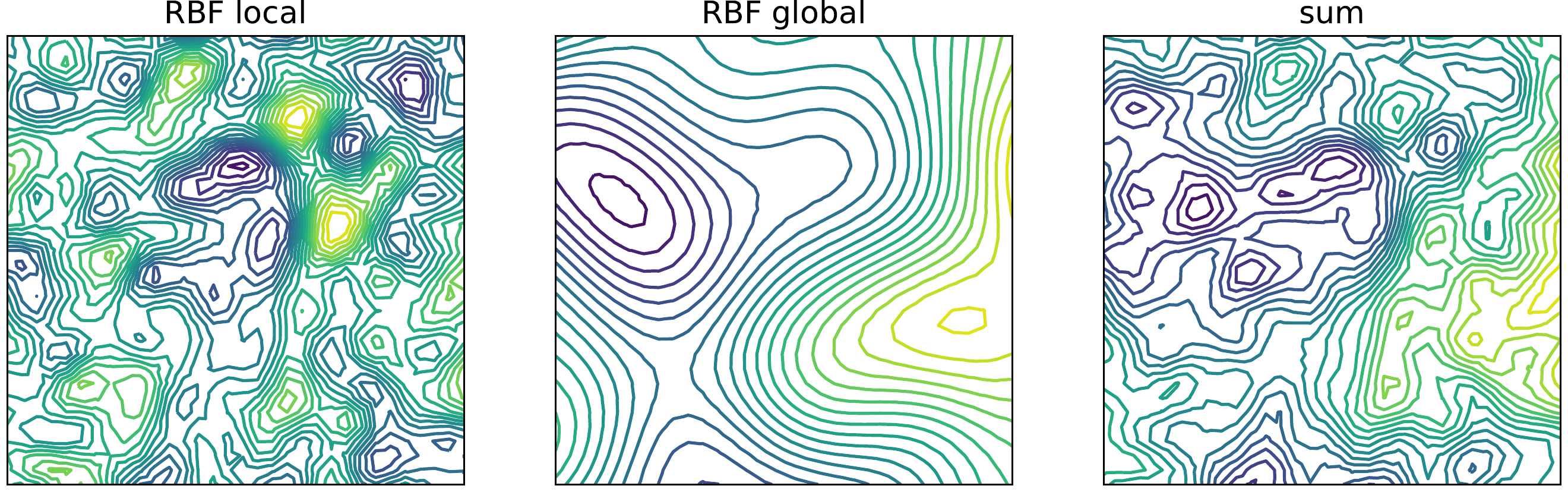}
	\caption{  Generated data with a sum of two SE-kernels with local and global lengthscales for experiment in Section \ref{se:examples_evaluation}.
	 }
	\label{fig:data2D}
\end{figure}

\subsection{Implementation Details}
\label{se:imp_details}

All experiments were run on a standard Laptop (IntelCore i7, 8 CPU 1.9GHz). Our code is implemented in Python and will be available on Github. 
\\
For solving the sparse linear system of equations, we used Cholmod \cite{chen2008algorithm} in the Python package scikit-sparse which relies on sparse Cholesky decomposition.
It would  be  advantageous to  use/implement a sparse block Cholesky decomposition and solver which exploits directly our structure. 
This was indeed needed for computing some entries in the posterior covariance, since with  available implementation of \textit{partial matrix inversion } we could not exploit the block sparsity and thus did not obtain competitive performance as discussed in Section \ref{se:systemAndInv}. An efficient implementation of this part is presented in Algorithm \ref{alg:alg1}.
\\
In our current implementation the size of each partition has to be equal; which is in theory not necessary, but it allows more efficient implementations since then the block character can be easily exploited in the computation of the sparse posterior precision.  Using the KD-tree construction with $J=2^K$, the sizes of the partitions differ at most by $1$. 
Thus, if the partitions are not equal, the number of local inducing points are set to $L=min(B_j)_{j=1}^J$.
\\
Our implementation exploits the kernel and likelihood functions of GPy \cite{gpy2014}.
For the optimization of the hyperparameters we used the L-BFGS-B algorithm in the Python package scipy in the deterministic full batch case. For stochastic optimization we used
 the stochastic optimizer ADAM \cite{kingma2014adam} (implemented from scratch) 
 with appropriate learning rates which are learnt in preliminary experiments.
\\
\\ 
For the competitor methods we used the implementation in GPy \cite{gpy2014} for full and sparse global GP (the approach of \cite{titsias2009variational}). For PoE, GPoE, BCM, RBCM and GRBCM
we implemented the corresponding aggregation algorithms based on the GPy implementations for the independent experts in Python for the sake of comparisons.
For the stochastic version of SGP, the hybrid information filter approach in \cite{kaniasparse} and their implementations are used. We also run the approaches REC \cite{schurch2020recursive} and SVI \cite{hensman2013gaussian}, however the former approach shows superior accuracy vs. time performance in preliminary experiments.
\\
For the sparse global GP model there is the choice of optimized or fixed inducing points. For the same number of inducing points the accuracy is obviously better with optimized inducing points, however taking into account the time for optimizing them, we found in the experiments with batch optimization (i.e. also smaller datasets) that the fixed random subset approach was superior. Therefore we report here the results for fixed (random subset of data) inducing points in the deterministic case and optimized in the stochastic case.
The reason for that is that the sparse global approximation with unknown inducing inputs
has $MD+\vert\bthe\vert$ (variational) parameters to optimize in the batch version. In the stochastic version REC \& IF there are as well $MD+\vert\bthe\vert$ parameters, whereas  SVI has even $M+0.5M^2+MD+\vert\bthe\vert$ number of parameters since the posterior mean and covariance has to be optimized.
On the other hand, full GP has only a few kernel hyperparameters $\vert\bthe\vert$ to optimize. Similarly, our method CPoE (and also independent PoEs) inherit this property because there is no necessity to optimize the local inducing points  since the total amount of them can be in the order of $N$. This is also true for the  stochastic version of our algorithm.
Assume for instance $D=8$ and $M=100$, the number of parameters with a SE kernel for full GP and CPoE are only $\vert\bthe\vert=10$ parameters to optimize, whereas
for batch SGP, REC \& IF $810$ and even $5910$ for SVI. 
For fixed inducing points, SGP and IF also only have $\vert\bthe\vert=10$ hyperparameters which allows to have more inducing points but speed-up the optimization a lot and makes the accuracy vs. time comparison more competitive.
\\ \quad \\
We used the KD-partition for our method as discussed in \ref{se:graphicalModel} while
in the PoE-literature \cite{fleet2014generalized, deisenroth2015distributed,  hinton2002training, rulliere2018nested, tresp2000bayesian}, often K-Means is used for partitioning. However, for large $J$ and $N$ this is quite inefficient and often the partition sizes for each expert differs significantly which introduces an imbalance among the experts in the prediction aggregation as well as in the stochastic optimization. Therefore we also used the KD-tree partition for these algorithms for the sake of comparisons. 
\\
\quad
\\
For assessing the quality of the different algorithms in the next sections, we report the two quantities 
 \textit{the Kullback-Leibler-(KL)-divergence}  to full GP and the \textit{Continuous Ranked Probability Score (CRPS)} both depending on the pointwise predictive distributions $\pc{f_*}{\by}$.
 The reported values correspond always to an average of $N_{test}$ prediction points which are not contained in the training data.

  \subsection{Experiments}
  \label{se:experiments}

\subsubsection{Synthetic Data}
\label{se:synthData2}

In this section we provide more details about the experiment in Section \ref{se:examples_evaluation}.
In this simulation study with synthetic GP data we examine the accuracy vs. time performance of different GP algorithms for fixed hyperparameters.
We generated $N=8192$ data samples in $D=2$ with 5 repetitions from the sum of two SE kernels 
with a shorter and longer lengthscale ($l_s=0.125, v_s=0.2$ and $l_l=0.5, v_l=1.1$; see Fig.\ \ref{fig:data2D}) such that both global and local patterns are present in the data.
In 	Fig.\ \ref{fig:synthetic2D} the mean results are shown for the KL and RMSE to full GP, the 95\%-coverage 
and the log marginal likelihood against time in seconds. 
\\
For the sparse GP, we use different number of fixed global inducing points $M=\{50,\ldots,1000\}$  for which the results are shown in blue.\footnote{We also run sparse GP with optimized inducing points, however the performance compared to time was worse.} From the PoE-family,  the results for minVar, GPoE and BCM
 are depicted for different number of experts $J=\{1,2,4,\ldots,64\}$ in red, cyan and magenta, respectively. 
For our correlated PoEs, the results for the correlations $C=\{1,\ldots,13\}$ are  shown in green for $J=32$ and $\gamma=0.5$.
\\In the first two plots, the superior performance of our method compared to competitors  in  accuracy to full GP vs. time can be observed.  
Our method constitutes a fast and accurate method for  a range of different approximation qualities. Moreover, in the third plot, one can observe that the confidence informations are reliable already for small approximation orders since it is based on the consistent covariance intersection method.


%
%

\subsubsection{Real World Data}
\label{se:realData2}

   \begin{table}
   \begin{small}
\centering
%
%
%
%
\begin{tabular}{l | rrrr | rrrr}
\toprule
&& \textbf{CRPS} &&&& \textbf{time}\\
\midrule
&   kin2 &  cadata &  sarcos &   casp &   kin2 &  cadata &  sarcos &   casp \\
\midrule
SGP(500)  &  0.183 &   0.253 &   0.069 &  0.329 
	 &  112.1 &   346.9 &   730.1 &   632.9 \\
SGP(1000) &  0.166 &   0.252 &   0.063 &  0.325
	 &  244.1 &   727.6 &  1718.5 &  1362.5 \\
minVar    &  0.173 &   0.257 &   0.052 &  0.294 
	 &   14.4 &    28.2 &    71.3 &    45.8 \\
GPoE      &  0.193 &   0.289 &   0.086 &  0.302
	&   14.4 &    28.3 &    71.4 &    45.6 \\
GRBCM     &  0.164 &   0.262 &   0.060 &  0.310 
	 &   16.5 &    33.5 &    84.6 &    59.4 \\
\textbf{CPoE(1)}   &  0.163 &   0.259 &   0.052 &  0.289 
	 &   13.8 &    24.5 &    45.4 &    45.1\\
\textbf{CPoE(2) }  &  0.155 &   0.251 &   0.051 &  0.287 
	&   18.9 &    33.4 &    67.3 &    70.3\\
\textbf{CPoE(3)}   &  \textbf{0.151} &   \textbf{0.249} &   \textbf{0.051} &  \textbf{0.282} 
	&   31.7 &    52.0 &   134.3 &   123.8\\
\bottomrule
\end{tabular}

\caption{Average CRPS  (left) and time (right) for different GP methods and 4 datasets with 5 repetitions. 
More details and results are provided in
Sections \ref{se:realData2} and \ref{se:tables}  in the Appendix.
}
\label{tab:CRPS_time_real}
\end{small}
\end{table}

Here we provide more details about the experiments with real world data as summarized in Section \ref{se:examples_evaluation}.
We downloaded all datasets form UCI repository \cite{asuncion2007uci} except the \textit{elecedemand} dataset is taken from \cite{fpp2}. We standardized all variables to mean zero and standard deviation of one (for \textit{elecdemand} see details below).
We
 use $N=min( 0.9N_{tot}, 1000)$ data sample for training, the rest for testing; except for \textit{kin} and \textit{elecdemand} we run experiments with $N_{test}=3000$  and $N_{test}=15288$ such that we could also run full GP a standard Laptop. 
For each dataset we fixed the number $J$ of experts (given in Table \ref{tab:datasets}) such that the partitions/mini-batches have a reasonable size ($\approx 500$).
   \\ 
For the deterministic SGP we used $M=100$ and for the stochastic SGP $M\in\{500, 1000\}$ inducing points (more results are provided in Appendix \ref{se:tables}). 
For our method CPoE we run the algorithm for $C\in\{1,2,3,4\}$ for the small and $C\in\{1,2,3\}$ for the large datasets with always $\gamma=1$.
For the stochastic versions we used learning rates $\delta=0.03$ for the dataset \textit{kin2} and $\delta=0.01$ for the remaining for all methods. The maximum number of epochs is set to $15$ together with a relative stopping criteria of $1e^{-2}$.
  We use a SE-kernel with a different lengthscale per dimension and initialized all hyperparameters to $1$, and the global inducing point to a random subset of the data.

\subsubsection{Application}
\label{se:appplication2}

%

This section contains additional details to the application described in Section \ref{se:examples_evaluation}
where  our method is applied to  the \textit{elecdemand} time series \cite{fpp2} which contains the half-hourly measured electricity demand together with the corresponding temperature and the variable whether it is a working day for 1 year.
In particular, the 
preprocessed 
dataset contains the standardized electricity demand (mean=0, sd=1) as  the response variable $y$, the normalized time as the first variable $X_1 \in [0,1]$, the standardized temperature and indicators as $X_2$ and $X_3$,
 respectively. The data is depicted in the first two plots in  Fig.\ \ref{fig:timeSeries},
where we shifted the first and third variable in the second plot for the sake of clarity.
%
 We removed the last day resulting in 364 days  = 52 weeks = 13 "months" consisting of 4 weeks. In each of the 13 "months",
we used the first 3 weeks for training and the last week for testing the out-of-sample accuracy.
In order  that it is possible to run full GP as comparison, we only used every 6th sample (corresponding to a measurement every 3h) of the training weeks for the actual training and the remaining for testing the in-sample accuracy. 
This gives $N=2184$, $N_{IN}=10920$ and $N_{OUT}=4368$ samples as depicted in the first plot in Fig.\ \ref{fig:timeSeries}.
\
Similarly as in the previous section, we run full GP, SGP and PoEs and CPoE and optimized the hyperparameter deterministically using the MAP as objective function taking into account the priors. 
For SGP we used $M\in\{100,200\}$ fixed inducing points, for PoEs and CPoE we used $J=13$ partitions which are obtained by splitting the first variable into $J$ blocks.
For CPoE we used $C\in\{1,2,3,4\}$ and $\gamma=1$.
The results are provided
in Table 
\ref{tab:KL_ELEC} which again shows very competitive performance also for a general kernel with priors on the hyperparameters.

\section{More Details about GPR}
\label{se:GP2}

In this section we provide more details for Section \ref{se:GP}.

Suppose we are given a training set 
$\mathcal{D} = \left\{ y_i, \bx_i \right\}_{i=1}^N$ 
of $N$ pairs of inputs $\bx_i\in \RR^D$ and noisy scalar outputs $y_i$ generated by adding independent Gaussian noise to a latent function $f(\bx)$, that is $y_i = f(\bx_i)+\varepsilon_i$, where $\varepsilon_i\sim \NNo{0,\sigma_n^2}$. 
We denote $\by = [y_1,\ldots,y_N]^T$ the vector of observations and with $\bX = [\bx_1^T,\ldots,\bx_N^T]^T \in \RR^{N\times D}$.
\\
We can model $f$ with a \textit{Gaussian Process} (GP), which defines a prior over functions  and can be converted into a posterior over functions once we have observed some data (consider e.g. \cite{rasmussen2006gaussian}).
To describe a GP, we only need to specify a mean $m(\bx)$ and a covariance function $k_{\bthe}(\bx,\bx')$ where $\bthe$ is a set of a few hyperparemeters.
Thereby, $k_{\bthe}$ is a positive definite kernel function (see \cite{rasmussen2006gaussian}),
for instance 
 the \textit{squared exponential} (SE) kernel with individual lengthscales for each dimension, that is
$k_{\bthe}(\bx,\bx') 
=\sigma_0^2 \exp\left(-\frac{1}{2}
\left(\bx-\bx'\right)^T \bs{L}^{-1} \left(\bx-\bx'\right)
\right)$ with $L=\DIAG{l_1^2,\ldots,l_D^2}$
and 
$\{\sigma_0, l_1,\ldots,l_D\}\in \bthe$.
  For the sake of simplicity, we assume
$m(x)\equiv 0$,
  however it could be any function.
Given the training 
	values $\bff = f\left(\bX\right)=\left[f(\bx_1),\ldots,f(\bx_N)\right]^T$
	and a test latent function value $f_*=f(\bx_*)$ at a test point $\bx_*\in \RR^D$, then the joint distribution $p(\bff, f_*)$ is Gaussian
	$\NNo
{\bO,\Kab{
[\bX ;
\bx_*]
}{
[\bX ;
\bx_*]
}
}$.
Thereby, 
	 we use the notation $[\bA_1;\bA_2]$ for the resulting matrix after stacking $\bA_1\in\RR^{{N_1}\times D}$ and $\bA_2 \in\RR^{{N_1}\times D}$  above each other 
and 
	 $\bK\in\RR^{M_1\times M_2}$ denotes the kernel covariance matrix with
	 entries
	$\left[ \Kab{\bA}{\bB} \right]_{ij}$ corresponding to the kernel evaluation $k_{\bthe}(\ba_i,\bb_j)$ with the corresponding rows $\ba_i,\bb_j$ for any $\bA\in \RR^{{M_1}\times D}$ and $\bB\in \RR^{{M_2}\times D}$.
\\
Typically, in GP regression, the  likelihood is Gaussian, that is, $\pc{\by}{\bff} = \NN{\by}{\bff, \sign \II}$, and with Bayes theorem \eqref{eq:GaB} we obtain analytically the predictive posterior 
	distribution
	$\pc{f_*}{\by} = \NN{f_*}{\bmu_*, \bSig_*}$ with
	%
	%
		%
		%
		%
		$
		\bmu_*=
		\Kab{\bx_*}{\bX} \left( \Kab{\bX}{\bX} + \sign \II \right)^{-1} \by$
		and
		$\bSig_*=
		\Kab{\bx_*}{\bx_*} - \Kab{\bx_*}{\bX} \left( \Kab{\bX}{\bX} + \sign \II \right)^{-1} \Kab{\bX}{\bx_*}.$
%
		%
		%
		%
	%
	%
Alternatively to the standard derivation shown above, 
the posterior distribution over the latent variables $\bff$ given the data $\by$
can be explicitly formulated as
\begin{align}
 \label{eq:postFull2}
\pc{\bff}{\by}
\propto
\pp{\bff,\by}
=
\pc{\by}{\bff}\pp{\bff}
=
\prod_{j=1}^J
\pc{\by_{j}}{\bff_{j}}
\pc{\bff_{j}}{\bff_{1:j-1}},
\end{align}
where the data is split into $J$ mini-batches of size $B$, i.e.
$\mathcal{D} = \left\{ \by_j, \bX_j \right\}_{j=1}^J$ 
with inputs $\bX_j\in \RR^{B\times D}$, outputs $\by_j\in \RR^B$ and the corresponding latent function values $\bff_j = f(\bX_j)\in\RR^B$. 
In  \eqref{eq:postFull}  we used the notation $\bff_{k:j}$ indicating $[\bff_k,\ldots,\bff_j]$ and the conditionals
$\pc{\bff_{j}}{\bff_{1:j-1}}$ can be derived from the joint Gaussian 
$\pp{\bff_{j},\bff_{1:j-1}}
=
\NNo
{\bO,\Kab{
[\bX_j ;
\bX_{1:j-1}]
}{
[\bX_j ;
\bX_{1:j-1}]
}
}$
via  Gaussian conditioning \eqref{eq:condGauss}. The corresponding graphical model of
 \eqref{eq:postFull}
is depicted in Figure \ref{fig:GPmods}(a)i).
%
Given the posterior over $\bff\vert\by$, the predictive posterior distribution 
from above
is equivalently obtained as
$
\pc{f_*}{\by}
=
\int \pc{f_*}{\bff} \pc{\bff}{\by} \diff \bff
$ via Gaussian integration
\eqref{eq:int}
where 
$\pc{f_*}{\bff} $ is derivable from the joint via  \eqref{eq:condGauss}. The graphical model of the prediction procedure is depicted in 	Figure \ref{fig:GPmods}(b)i).
We present this  alternative two stage procedure to highlight later connections to our model with full GP.

		\subsection{Global Sparse GPs}
	\label{se:sparse_GP2}
		Sparse GP regression approximations based on \textit{global inducing points} reduce the computational complexity by introducing  
	$M \ll N$ inducing points $\ba \in \RR^M$ 
	that optimally summarize the dependency of the whole training data globally, compare the graphical model  in  Figure \ref{fig:GPmods}b).
	Thereby the  inducing \textit{inputs} $\bA \in \RR^{M\times D}$ are in the $D$-dimensional input data space and the inducing \textit{outputs} $\ba = f(\bA)\in \RR^M$ are the corresponding GP-function values. 
	In the following, this model is denoted by SGP$(M)$.
	%
	%
Similarly to full GP in Eq.  \eqref{eq:postFull}, the posterior over the inducing points $p(\ba\vert\by) \propto \int \pp{\ba,\bff,\by} \diff \bff$ can be derived from the joint distribution
\begin{align}
 \label{eq:sparse2}
\pp{\ba,\bff,\by}
=
\pc{\by}{\bff}
\pc{\bff}{\ba}
p(\ba)
=
\prod_{j=1}^J
\pc{\by_{j}}{\bff_{j}}
\pc{\bff_{j}}{\ba}
p(\ba_j \vert\ba_{1:j-1} )
,
\end{align}
where the usual Gaussian likelihood 
$\pc{\by_{j}}{\bff_{j}}=\NNo{\bff_{j},\sigma_n^2\II}$
is used and $\pc{\bff_{j}}{\ba}$ can be derived from the  joint Gaussian 
$\pp{\bff_{j},\ba}
=
\NNo
{\bO,\Kab{
[\bX_j ;
\bA]
}{
[\bX_j ;
\bA]
}
}$
with  \eqref{eq:condGauss}.
%
%
%
%
	%
	Using the posterior computed via  \eqref{eq:sparse} together with the predictive conditional $\pc{f_*}{\ba}$ derived by \eqref{eq:condGauss} from the assumed joint $
\pp{f_*,\ba}
=\NNo{\bO,\Kab{[\bx_*,\bA]}{[\bx_*,\bA]}}
$
and integrating 
	$\int\pc{f_*}{\ba} \pp{\bff_{j},\ba} \diff\ba$ via \eqref{eq:int}
	provides an approximation to the predictive posterior
	of full GP.
	Batch inference in these sparse global models can be done in $\mathcal{O}(M^2N)$ time and $\mathcal{O}(MN)$ space (e.g. \cite{quinonero2005unifying}). 
	\\		
	In order to find optimal inducing inputs $\bA$ and hyperparameters $\bthe$, a sparse variation of the log marginal likelihood similar 
	 can be used e.g.\ \cite{bui2016unifying, snelson2006sparse, titsias2009variational}.
In particular,
the authors in \cite{titsias2009variational}  proposed to maximize a variational lower bound to the true GP marginal likelihood which
	has the effect that the sparse GP predictive distribution  
	converges  to the full GP predictive distribution
	as the number of inducing points increases. For larger datasets, stochastic optimization  has been applied e.g.\ \cite{bui2017streaming, hensman2013gaussian, kaniasparse, schurch2020recursive} to obtain faster and more data efficient optimization procedures.
	  For  recent reviews on the subject consider
	  e.g.\ \cite{liu2020gaussian,quinonero2005unifying,rasmussen2006gaussian}.

%
%

\subsection{Local Independent GPs}\label{se:localGPR2}
An alternative to the global sparse inducing point methods as presented in the previous section
constitute local approaches which exploit multiple local GPs combined with averaging techniques to boost predictions. 
Beside other averaging techniques (e.g. mixture of experts)
the \textit{Product of Expert (PoE)} scheme was proposed by \cite{hinton2002training}
where individual predictions 
$\pc{f_{*j}}{\by_{j}}$ 
from $J$ experts based on the local data $\by_j$
are aggregated to the final predictive distribution
\begin{align}
\label{eq:PoEagg2}
\pc{f_*}{\by}
=
\prod_{j=1}^{J}
g_j\left(
\pc{f_{*j}}{\by_{j}}
\right)
\end{align}
where
$g_j$ is a function depending on the particular PoE method discussed below and is in the original work of \cite{hinton2002training} just
the identity.
%
Note that we present here the version of PoEs where the noiseless predictions $f_{*j}$ are aggregated instead of noisy aggregation with $y_{*j}$ as described in some work of PoEs.
The individual predictions $\pc{f_{*j}}{\by_{j}}$ are local GP fits
$
\int 
\pc{f_{*j}}{\bff_{j}}
\pc{\bff_j}{\by_j}
\diff
\bff_j
$ 
involving 
the predictive conditionals
$\pc{f_{*j}}{\bff_{j}}$
derived by \eqref{eq:condGauss} from the assumed joint $
\pp{f_{*j},\bff_{j}}
=\NNo{\bO,\Kab{[\bx_*,\bX_j]}{[\bx_*,\bX_j]}}
$
and 
the local posteriors
%
$\pc{\bff_{j}}{\by_{j}}
\propto
\pc{\by_{j}}{\bff_{j}}
\pp{\bff_{j}}
$, 
where the individual prior $\pp{\bff_{j}}=\NNo{\bO,\Kab{\bX_j}{\bX_j}}$. Together with the usual Gaussian likelihood 
$\pc{\by_{j}}{\bff_{j}}=\NNo{\bff_{j},\sigma_n^2\II}$, the final noisy predictive distribution $\pc{y_{*}}{\by}$ can be obtained via 
$
\int 
\pc{y_{*}}{f_*}
\pc{f_*}{\by}
\diff
f_*.
$
Similarly to Eqs. \eqref{eq:postFull} and \eqref{eq:sparse}, the implicit 
posterior in all PoE method is 
\begin{align}
 \label{eq:PoE2}
\pc{\bff}{\by}
\propto
\pp{\bff,\by}
=
\pc{\by}{\bff}
\prod_{j=1}^J
\pp{\bff_{j}}
=
\prod_{j=1}^J
\pc{\by_{j}}{\bff_{j}}
\pp{\bff_{j}},
\end{align}
where the corresponding graphical model is depicted in Figure \ref{fig:GPmods}c) and \ref{fig:GPmods5}c).

The function $g_j$ in \eqref{eq:PoEagg2}  takes as argument the predictive distribution $p_{*j}: = \pc{f_{*j}}{\by_j}$ which depends implicitly also on $\bx_*$. 
In the original work \cite{hinton2002training} the authors used the
 identity $g_j(p_{*j})=p_{*j}$ which produce underconfident prediction variances \cite{liu2020gaussian}.
In order to mitigate 
this issue, 
the aggregation weights $g_j(p_{*j})=p_{*j}^{1/J}$  were proposed \cite{fleet2014generalized} but still resulting in too large predictive uncertainty estimates  \cite{liu2020gaussian}.
%
The reason is that the experts are all equally weighted, however, the predictions at a particular point $\bx_*$ are not equally reliable,
therefore in the generalized PoE (GPoE) \cite{fleet2014generalized} some varying weights  $\beta_j(x_*)$
 were introduced to quantify the contribution of the expert $j$ at $\bx_*$. Thus,  
 $g_j(p_{*j}) = p_{*j}^ {\beta_j(x_*)}$ with
  weights set to the difference in entropy  between the expert's prior  and posterior, that is,
$\bar{\beta}_{*j}
=
\frac{1}{2}
\log
 \left(
\frac{v_{*0}}{v_{*j}} \right).$
This has the effect of
 increase or decreasing the importance of the experts 
 based on the corresponding prediction uncertainty $v_{*0}$ and $v_{*j}$.
 However, these general weights can produce overconfident uncertainty estimates, therefore the authors in  \cite{fleet2014generalized}  proposed also an version with normalized weights such that $\sum_j^J\bar{\beta}_j(x_*)=1$. 
 In the following, PoE and GPoE refer to the version with normalized weights.
Other important contributions in this field are 
BCM \cite{tresp2000bayesian} and its robustified version RBCM \cite{fleet2014generalized},
GRBCM \cite{liu2018generalized}, distributed local GPs \cite{deisenroth2015distributed} and local experts with consistent aggregations 
 \cite{rulliere2018nested, nakai2021nested}. We refer to \cite{liu2020gaussian} for a recent overview.

%

Simple baseline methods are the \textit{minimal variance (minVar)} and the
\textit{nearest expert (NE)} aggregation, where only the prediction from the expert with minimal variance or nearest expert is used, respectively. 
%
Although both these method show often surprisingly good performance, they suffer from an huge disadvantage, namely that there are serious discontinuities at the boundaries between the experts (see for instance Fig.\ 	\ref{fig:toy1D}) and thus often not useful in practice. This is also the main limitation of all local methods  based only on the prediction of one expert (e.g.\  \cite{bui2014tree, datta2016hierarchical, katzfuss2020vecchia, 
katzfuss2021general}) and it was one of the reason for introducing smooth PoEs with combined experts.
Since in basically all cases minVar is better than NE (which is also consistent with the findings in \cite{rulliere2018nested}), we only compare our method to minVar and not NE for the sake of simplicity.

\section{Useful properties} 

\subsubsection{Inversion Lemma}
Given invertible matrices $\bs{A}\in \RR^{B\times B}$ , $\bs{C}\in \RR^{M\times M}$ and matrices $\bs{U}\in \RR^{B\times M}$, $\bs{V}\in \RR^{M\times B}$, it holds
\begin{align}
\begin{split}
\label{eq:matrix_inv_lemma}
\left(
\bs{A} + \bs{U} \bs{C} \bs{V}
\right)^{-1}
=
\bs{A}^{-1} - \bs{A}^{-1}\bs{U}
\left(
\bs{C}^{-1} + \bs{V}  \bs{A}^{-1} \bs{U}
\right)^{-1}
\bs{V}
\bs{A}^{-1}
.
\end{split}
\end{align}

\subsubsection{Determinant Lemma}
Given invertible matrices $\bs{A}\in \RR^{B\times B}$ , $\bs{C}\in \RR^{M\times M}$ and matrices $\bs{U}\in \RR^{B\times M}$, $\bs{V}\in \RR^{M\times B}$, it holds
\begin{align}
\label{eq:det_lemma}
\deT{ \bs{A} + \bs{U} \bs{C} \bs{V} }
= \deT{
\bs{C}^{-1} + \bs{V} \bs{A}^{-1} \bs{U}
}  
\deT{\bs{C}}
\deT{\bs{A}}
.
\end{align}

\subsubsection{Block Inversion}
Given an invertible, symmetric block  matrix 
$$
\bM = 
\begin{bmatrix}
\bA &\bB\\
\bB^T &\bD
\end{bmatrix},$$ the inverse can be computed as
\begin{align}
\label{eq:blockInverse}
\bM^{-1} = 
\begin{bmatrix}
\bA^{-1} + \bA^{-1}\bB   \bZ^{-1} \bB^T\bA^{-1}
&
-\bA^{-1}\bB\bZ^{-1}
\\
- \bZ^{-1} \bB^T \bA^{-1}\
&
\bZ^{-1}
\end{bmatrix}
\end{align}
with
$\bZ = \bD-\bB^T\bA^{-1}\bB$.

\subsubsection{Block Determinant}
Given an invertible, symmetric block  matrix 
$$
\bM = 
\begin{bmatrix}
\bA &\bB\\
\bB^T &\bD
\end{bmatrix},$$ the determinant can be computed as
\begin{align}
\label{eq:blockDet}
\vert \bM \vert 
= \vert \bA\vert ~\vert \bD-\bB^T\bA^{-1}\bB\vert
 = \vert \bD\vert ~\vert  \bA-\bB\bD^{-1}\bB^T\vert.
\end{align}

\subsubsection{Conditional Gaussians}
From the joint Gaussian 
$
[\ba, \bb]^T
\sim 
\mathcal{N}\left(\bO,\Kab{[\bA,\bB]}{[\bA,\bB]}\right)$,
the conditional can be computed as follows
\begin{align}
\begin{split}
\label{eq:condGauss}
\ba \vert \bb
\sim \NNo{
\bK_{\bA\bB}  \bK_{\bB\bB}^{-1}\bb
, 
\bK_{\bA\bA} - \bK_{\bA\bB} \bK_{\bB\bB}^{-1} \bK_{\bB\bA}
}
=
\NNo{\bH_{AB}\bb,\bV_{\bA\bA}^{\bB}}
\end{split}
\end{align}

\subsubsection{Marginalization/Integration}
Given the densities $\pp{\ba}=\NNo{\bmu,\bSig}$ and 
$\pc{\bb}{\ba}=\NNo{\bF\ba +\bv,\bQ}$, then 
\begin{align}
\begin{split}
\label{eq:int}
\pp{\bb}&=
\int  \pp{\ba, \bb} \diff \ba
=
\int \pc{\bb}{\ba} \pp{\ba} \diff \ba
=
\NNo{\bF\bmu+\bv,\bF\bSig\bF^T+\bQ}
\end{split}
\end{align}

\subsubsection{ Gaussian \& Bayes}
Given  the densities $\pp{\ba}=\NNo{\bmu,\bSig}$ and 
$\pc{\bb}{\ba}=\NNo{\bF\ba +\bv,\bQ}$, applying Bayes' formula yields
\begin{align}
\label{eq:GaB}
\pc{\ba}{\bb}=
\NNo{\bP\left( 
\bF^T\bQ^{-1}(\bb-\bv)+\bSig^{-1}\bmu\right)
,\bP},
\end{align}
with
$\bP=\left(
\bSig^{-1}+\bF^T\bQ^{-1}\bF\right)^{-1}$.

\subsubsection{Product of Gaussians}
Assume $J$ Gaussians $p_J(\bx)=\NN{\bx}{\bmu_j,\bSig_j}$ and 
$a_j\in \RR$.
Then the product can be written as
\begin{align}
\begin{split}
\label{eq:prodGauss}
\prod_{j=1}^J p_j(\bx)^{a_j}
=
\NN{\bx}{\bmu,\bSig}
\end{split}
\end{align}
with 
\begin{align*}
\bSig = \left(
\sum_{j=1}^J a_j \bSig_j^{-1}
\right)^{-1}
\text{and}\quad
\bmu = \bSig 
\left(
\sum_{j=1}^J a_j \bSig_j^{-1} \bmu_j
\right)
\end{align*}
as long as $\bSig$ positive-semi-definite (if $a_j \equiv 1$ then always the case).

\subsubsection{Entropy of Gaussian}
The \textit{Entropy} $H$ of $p(\bx)=\NN{\bx}{\bmu,\bSig}$ with $\vert \bx \vert = B$ is defined as
\begin{align}
\begin{split}
\label{eq:entropy}
H[\bx]
=
H[p(\bx)]
=
\frac{1}{2}\left(
\log \vert \bSig\vert +
B(1+\log 2\pi)
\right),
\end{split}
\end{align}
where we use $log$ as the natural logarithm and thus the entropy is measured in nats (natural units).

\subsubsection{Kullback-Leibler-Divergence (KL)}
The KL between $p_0(\bx)$  and $p_1(\bx)$ is defined as
\begin{align}
\label{eq:KLgeneral}
KL[p_0(\bx)~\vert~ p_1(\bx)] = \int 
p_0(\bx)
\log \frac{p_0(\bx)}{p_1(\bx)} \diff \bx.
\end{align}

\subsubsection{KL between 2 Gaussians}
The \textit{Kullback-Leibler-Divergence (KL)} between $p_0(\bx)=\NNo{\bmu_0,\bSig_0}$  and $p_1(\bx)=\NNo{\bmu_1,\bSig_1}$ with $\vert \bx \vert = B$ can be computed by
\begin{align}
\begin{split}
\label{eq:KLmulti}
KL[p_0(\bx)~\vert~ p_1(\bx)]
=
\frac{1}{2}\left(
tr(\bSig_1^{-1}\bSig_0) 
- B
\right.
\\
\left.
+(\bmu_1-\bmu_0)^T\bSig_1^{-1} (\bmu_1-\bmu_0)
+ \log\frac{\vert\bSig_1\vert}{\vert\bSig_0\vert}
\right).
\end{split}
\end{align}

\subsubsection{Difference in KL of Gaussian with Zero Mean}
The difference in KL  between $p_1(\bx)=\NNo{\bO,\bSig_1}$  and $p_2(\bx)=\NNo{\bO,\bSig_2}$ with same base distribution $p_0(\bx)=\NNo{\bO,\bSig_0}$ can be computed by
\begin{align}
\begin{split}
\label{eq:KLdiff}
KL[p_0(\bx)~\vert~ p_1(\bx)]-
KL[p_0(\bx)~\vert~ p_2(\bx)]
=
\frac{1}{2}\left(
tr((\bSig_1^{-1}-\bSig_2^{-1})\bSig_0) 
+ \log\frac{\vert\bSig_1\vert}{\vert\bSig_2\vert}
\right).
\end{split}
\end{align}

\subsubsection{General Difference in KL }
%
%
Let $1\leq C<J$ and $0<\gamma \leq 1$ be fixed. For any $C_2 \in \{C,\ldots,J\}$
we define the difference in KL, denoted as $\mathbb{D}_{(C,C_2)}[\bx]$, between the true distribution of $\bx$ and two different approximate distributions, i.e.
\begin{align}
\label{def:KL_diff}
\mathbb{D}_{(C,C_2)}[\bx]
&=
KL[\pp{\bx} \mid\mid q_{c,\gamma}(\bx)]
-
KL[\pp{\bx} \mid\mid q_{{c_2,\gamma}}(\bx)]
=
\mathbb{E}_{p(\bx)}\left[\log \frac{q_{c_2,\gamma}(\bx)}{q_{c,\gamma}(\bx)}\right]
\end{align}
using the definition of KL \eqref{eq:KLgeneral}. 
%
%
%
%
Similarly,
  we define the 
 the difference in KL, denotes as $\mathbb{D}_{(C,C_2)}[\bx\vert \by]$, of a conditional distribution $\bx\vert\by$ to be
\begin{align}
\begin{split}
\label{eq:KL_diff2}
&KL[\pp{\bx\vert \by} \mid\mid q_{c,\gamma}(\bx\vert \by)]
-
KL[\pp{\bx\vert \by} \mid\mid q_{c_2,\gamma}(\bx\vert \by)]
\\
=
&
\mathbb{E}_{p( \by)}\left[
\mathbb{E}_{p(\bx\vert \by)}\left[\log \frac{q_{c_2,\gamma}(\bx\vert \by)}{q_{c,\gamma}(\bx\vert \by)}\right]
\right].
\end{split}
\end{align}
which follows from the 
 the definition of KL \eqref{eq:KLgeneral}.

\section{Proofs and Additional Results}
\label{se:proofAdd}

\subsection{Additional Results}
\label{se:addRes}

\begin{figure}[htp!]
\centering
    \includegraphics[width=0.8\linewidth]{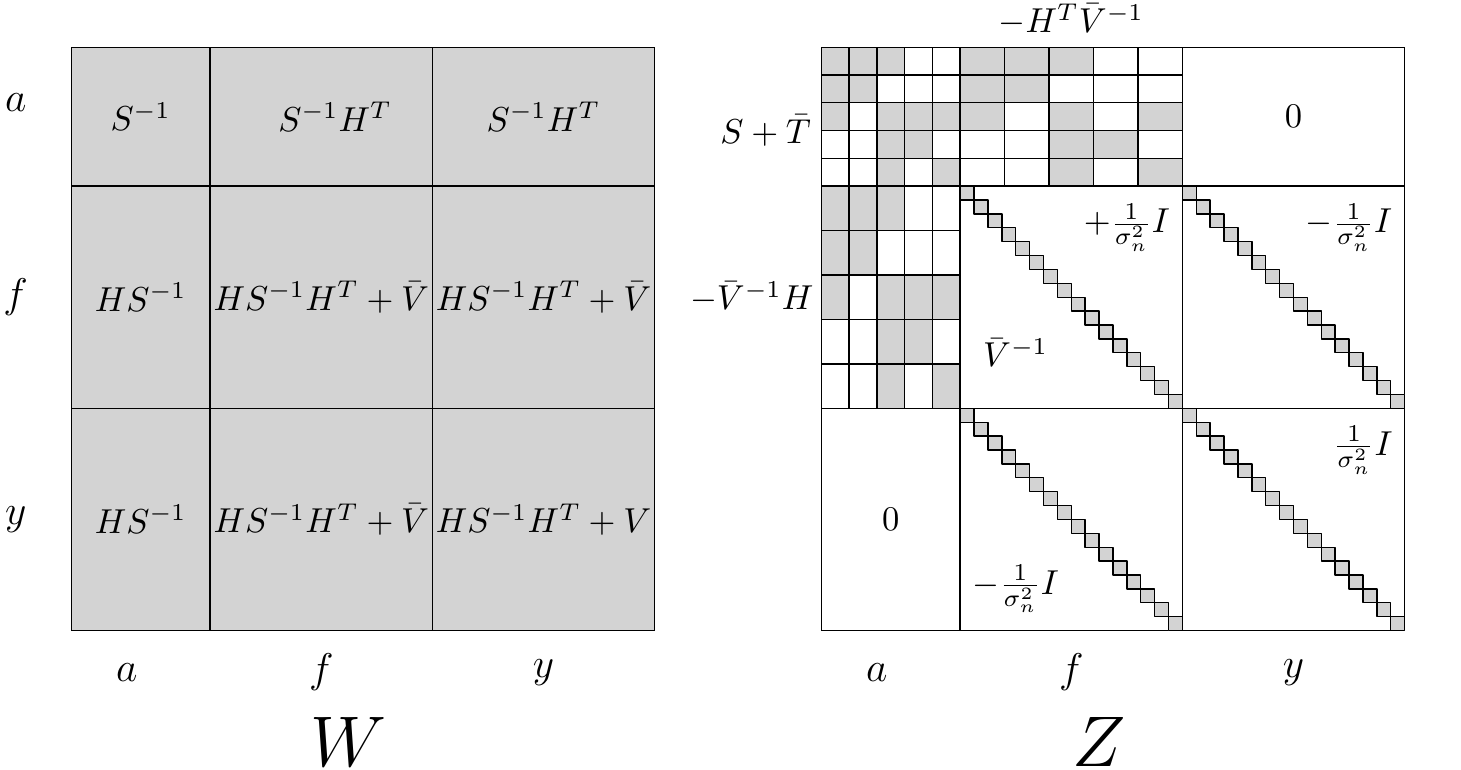}
\caption{Covariance $\bW$ and precision $\bZ=\bW^{-1}$ 
of joint prior approximation $q_{c,\gamma}(\ba,\bff,\by=\NNo{\bO, \bW}$ of CPoE model.
Compare Proof \ref{proof:priorCovariance}.
}
 \label{fig:joint_prior_wo_star}
\end{figure}

\begin{figure}[htp!]
\centering
    \includegraphics[width=0.9\linewidth]{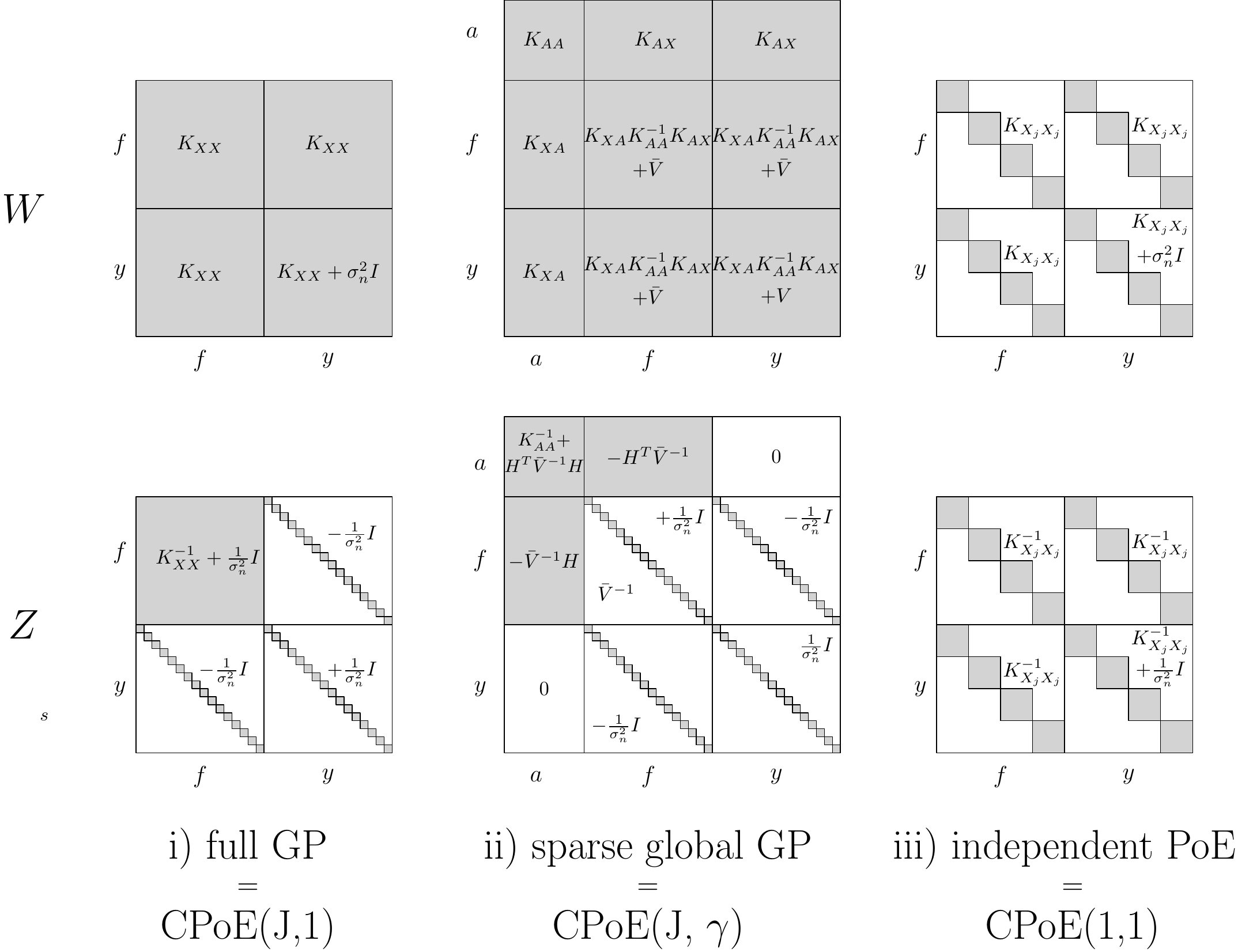}
\caption{
Covariance $\bW$ and precision $\bZ=\bW^{-1}$ 
of joint prior  of different GP models. Compare Figure  \ref{fig:joint_prior_wo_star} for the corresponding covariance and precision matrices for CPoE model.
Note that we used $\bH=\bK_{\bX\bA} \bK_{\bA\bA}^{-1}$ and $\bar{\bV}$ the same  as in the local CPoE.
}
\end{figure}

\begin{figure}[htp!]
\centering
    \includegraphics[width=0.95\linewidth]{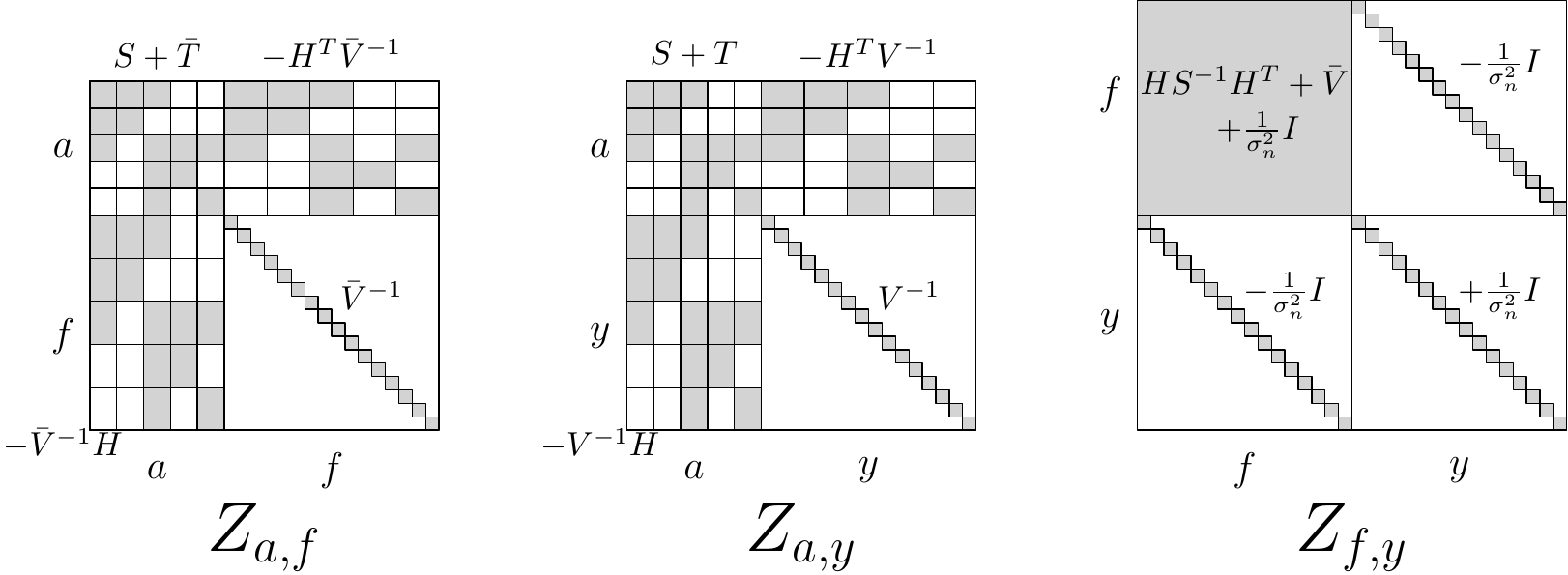}
\caption{Marginalized precision corresponding to 
$q_{c,\gamma}(\ba,\bff) = \NNo{\bO, \bZ_{\ba,\bff}^{-1}}$,
$q_{c,\gamma}(\ba,\by) = \NNo{\bO, \bZ_{\ba,\by}^{-1}}$
and
$q_{c,\gamma}(\bff,\ba) =  \NNo{\bO, \bZ_{\bff,\by}^{-1}}$, respectively.
Thereby we used the notation $\bV = \bar{\bV} + \sigma_n^2 \II$, 
$\bar{\bT}= \bH^T \bar{\bV}^{-1} \bH$
and
$\bT= \bH^T \bV^{-1} \bH$.
Note that the corresponding dense covariance matrices are directly obtained from $\bW$ in Fig.\ \ref{fig:joint_prior_wo_star} by selecting the corresponding entries.
}
\end{figure}

\begin{proposition}[Marginal Likelihood; Proof \ref{proof:margLik}]
\label{prop:margLik}
The 
marginal likelihood 
is   
\begin{align*}
q_{C,\gamma}(\by\vert \bthe)
= q_{C,\gamma}(\by)
 &=
\int q_{C,\gamma}(\by,\ba) \diff \ba
=
  \NNo{\bO,\bP}
\end{align*}
with
$\bP=\bH\bS^{-1}\bH^T +\bV \in 
\RR^{N\times N}$
where all dependencies on $\bthe$  of the matrices are omitted.
\end{proposition}
\begin{proposition}[Prior Approximation II; Proof \ref{proof:priorApprox2}]
\label{prop:priorApprox2}
Alternatively to Proposition
\ref{prop:priorApprox},
 the prior approximation $\q{\ba}=\NN{\ba}{\bs{0},\bS^{-1}}$ 
can be equivalently written as
\begin{align*}
\q{\ba}
=
\prod_{j=1}^J
\pc{\ba_{j} }{ \ba_{\bs{\pi}(j)} }
=
\prod_{j=1}^J
\NN{\ba_{\bs{\pi}^+(j)}}{\bO, \bS_{(j)}^{-1}}
\end{align*}
with
$\bS_{(j)} = \tilde{\bF}_{j}^T \bQ_j^{-1} \tilde{\bF}_{j}$
,
$\tilde{\bF}_{j} = \begin{bmatrix}
-\bF_j & \II 
\end{bmatrix}$
and
$\bs{\pi}^+(j) = \bs{\pi}(j) \cup j$.
Further, the prior precision matrix can also be written as 
$$
\bS
= 
\sum_{j=1}^J \overline{\bS}_{(j)}
$$
where
$\overline{\bS}_{(j)} \in \RR^{M\times M}$
is the augmented matrix consisting of 
$\bS_{(j)} \in \RR^{LC\times LC}$
at the entries $[\bs{\pi^+(j)},\bs{\pi^+(j)}]$ and $0$ otherwise.
\end{proposition}
\begin{proposition}[Prior Approximation III; Proof \ref{proof:priorApprox3}]
\label{prop:priorApprox3}
Alternatively to Prop.\
\ref{prop:priorApprox} and Prop.\ 
\ref{prop:priorApprox2}
 the prior approximation $\q{\ba}$ 
can be equivalently written as
\begin{align*}
\q{\ba}
=
\prod_{j=1}^J
\frac{
\pp{\ba_{j}, \ba_{\bs{\pi}(j)} } 
}{
\pp{\ba_{\prB{j}}}
}
=
\prod_{j=C}^J
\frac{
\pp{ \ba_{\bs{\pi}^+(j)} } 
}{
\pp{\ba_{\prB{j}}}
}
\end{align*}
which is a Gaussian
$\NN{\ba}{\bs{0},\bS^{-1}}$ with
prior precision 
$$
\bS
= 
\sum_{j=1}^J \overline{\bK}^{-1}_{\bA_{\bs{\pi}^+(j)}\bA_{\bs{\pi}^+(j)}}
-
\overline{\bK}^{-1}_{\bA_{\prB{j}}\bA_{\prB{j}}}
$$
where
$\overline{\bK}^{-1}_{\bA_{\bs{\phi}} \bA_{\bs{\phi}}} \in \RR^{M\times M}$
is the augmented matrix consisting of 
$\bK^{-1}_{\bA_{\bs{\phi}} \bA_{\bs{\phi}}} \in \RR^{T\times T}$
at the entries $[\bs{\phi},\bs{\phi}]$ and $0$ otherwise with $T=\vert \bs{\phi}\vert$. 
\end{proposition}
\begin{proposition}[Exact Diagonal of Prior; Proof \ref{proof:traceKaa}]
\label{prop:traceKaa}
The precision matrix $\bS_C$ of the prior approximation 
$q_C(\ba)$ is exact on the diagonal, that is,
\begin{align*}
tr(\bS_C \bK_{\bA\bA})=JL
\end{align*}
where $JL$ is the dimension of the matrices.
\end{proposition}
\begin{proposition}[Band-Diagonal Approximation]
\label{prop:prop11}
In the \textbf{consecutive} case, i.e.\ $\psB{j}=\{j-C+1,\ldots,j\}$, 
the block-entries 
\begin{align*}
\bS^{-1}_{\left[\psB{j},\psB{j}\right]}
=
{\Kab{\bA}{\bA}}_{\left[\psB{j},\psB{j}\right]},
\end{align*}
are equal which means that the block-band-diagonals $-C+1,\ldots,0,\ldots,C-1$ of the both matrices are the same.
For the case $C=J$ it holds
$\bS^{-1}
=
\Kab{\bA}{\bA}$.
\end{proposition}
\begin{proposition}[Decreasing Prior Entropy; Proof \ref{proof:decPriorEntropy}]
\label{prop:decPriorEntropy}
For any predecessor structure $\bs{\pi}_C$ as in Def.\ \ref{def:predSets},
the 
entropy 
$H$ 
of the approximate prior
$q_C(\ba)$ is decreasing for $C \rightarrow J$, in particular 
\begin{align*}
H\left[  q_1(\ba) \right]
\geq \cdots \geq 
 H\left[  q_j(\ba) \right]
\geq \cdots \geq 
 H\left[  q_J(\ba) \right]
\end{align*}
where it holds
$
H\left[ q_J(\ba) \right]
=  H\left[  p(\ba) \right]$
and
$$H[q_j(\ba)] = \frac{1}{2}\log\vert\bQ_C \vert +  \frac{M}{2}(1+\log 2\pi)
.$$
Similar results can be obtained for the joint prior $q_C(\ba,\bff,\by)$.
\end{proposition}
From the last proposition we know that increasing the degree of correlation $C$ add always more information to the prior. In particular,  the prior of complete independent PoEs (i.e.\ $C=1$) encodes the least of information since all correlations between the experts are missing, whereas 
the prior of full GP 
incorporates the most information since all correlations are modeled. 
\begin{proposition}[Prior Quality II]
\label{prop:priorImp2}
The prior approximation quality improvement 
$
\mathbb{D}_{(C,T)}[\ba]$ in Prop.\ \ref{prop:convJointPrior} can be equivalently written as
\begin{align*}
\begin{split}
\mathbb{D}_{(C,T)}
=\frac{1}{2}
\log
\frac{\vert\bS_{C+T}\vert}{\vert\bS_{C}\vert}
=
\frac{1}{2}
\sum_{j=1}^J
\log
\frac{
\vert \bQ_{j\vert \bs{\pi}_j}\vert  ~
\vert \bQ_{\bs{\phi}_j\vert \bs{\pi}_j }\vert 
}{
\vert \bQ_{j\cup\bs{\phi}_j \vert \bs{\pi}_j }\vert 
}
\\
=
\frac{1}{2}
\sum_{j=1}^J
\log
\frac{
\vert \bK_{\bA_{j\cup\pi_j}\bA_{j\cup\pi_j}} \vert  
~
\vert \bK_{\bA_{\phi_j\cup\pi_j}\bA_{\phi_j\cup\pi_j}} \vert 
}{
\vert \bK_{\bA_{j\cup\phi_j\cup\pi_j}\bA_{j\cup\phi_j\cup\pi_j}}  \vert 
~
\vert \bK_{\bA_{\pi_j}\bA_{\pi_j}} \vert 
}
\end{split}
\end{align*}
where
$\bs{\pi}_j=\bs{\pi}_C(j)$, 
$\bs{\phi}_j = \cup_{i=C+1}^{C+T}\bs{\phi}_{i}(j)$
and
$
\bQ_{\varphi_1\vert \varphi_2}
=
\bK_{\bA_{\varphi_1}\bA_{\varphi_1}}  -
\bK_{\bA_{\varphi_1}\bA_{\varphi_2}} \bK_{\bA_{\varphi_2}\bA_{\varphi_2}}^{-1} \bK_{\bA_{\varphi_2}\bA_{\varphi_1}}$.
\end{proposition}

\subsection{Proofs}
\label{se:proofs}
\begin{proofII}[Proof of Prop.\ \ref{prop:graphicalModel}; Joint Distribution]
\label{proof:graphicalModel}.
The matrices in the conditional distributions \eqref{eq:gaussDef} and \eqref{eq:gaussDef2} 
in 
Prop.\ \ref{prop:graphicalModel}
 can be obtained via Gaussian conditioning  
 \eqref{eq:condGauss}
%
%
from the assumed joint densities
\begin{align*}
\pp{
%
\bff_j^i, \ba_{\psB{j}}
} 
&=
\NNo
{\bO,\Kab{
[
\bX_j^i ;
\bA_{\psB{j}}
]
}{
[
\bX_j^i ;
\bA_{\psB{j}}
]
}
}
;\\
\pp{
\ba_j, \ba_{\prB{j}}
} 
&=
\NNo
{\bO,\Kab{
[
\bA_j ,
\bA_{\prB{j}}
]
}{
[
\bA_j ,
\bA_{\prB{j}}
]
}
},
\end{align*}
%
%
resulting in 
\begin{align*}
\bH_{j}
&=
\bK_{\bX_{j}\bA_{\psB{j}}}  \bK_{\bA_{\psB{j}}\bA_{\psB{j}}}^{-1}
;\\
\overline{\bV}_{j}
&=
Diag[
\bK_{\bX_j\bX_j}  -
\bK_{\bX_j\bA_{\psB{j}}} \bK_{\bA_{\psB{j}}\bA_{\psB{j}}}^{-1} \bK_{\bA_{\psB{j}}\bX_j}
]
;\\
\bF_{j}
&=
\bK_{\bA_{j}\bA_{\prB{j}}}  \bK_{\bA_{\prB{j}}\bA_{\prB{j}}}^{-1}
;\\
\bQ_{j}
&=
\bK_{\bA_j\bA_j}  -
\bK_{\bA_j\bA_{\prB{j}}} \bK_{\bA_{\prB{j}}\bA_{\prB{j}}}^{-1} \bK_{\bA_{\prB{j}}\bA_j}
\end{align*}
with $\bF_1 = \bs{0}$ and $\bQ_1=\bK_{\bA_1\bA_1}$.
\end{proofII}
\begin{proofII}[Proof used in  Def.\ \ref{def:jointDistr}; Joint Distribution II]
\label{proof:jointDistr_full}
In the case $\gamma=1$, thus
$\ba_j=\bff_j$ and $\ba=\bff$, the  joint distribution can be written as
$\q{\bff, \ba,\by} = 
\q{\bff, \bff,\by} = 
\q{\bff,\by}$ is
\begin{align*}
\q{\bff, \by}
&=
\prod_{j=1}^J
\pc{\by_{j}}{\bff_{j}}
\pc{\bff_{j}}{ \ba_{\psB{j}} }
\pc{\ba_{j}}{ \ba_{\prB{j}} }
\\
&=
\prod_{j=1}^J
\pc{\by_{j}}{\bff_{j}}
\pc{\bff_{j}}{ \bff_{\psB{j}} }
\pc{\bff_{j}}{ \bff_{\prB{j}} }
\\
&=
\prod_{j=1}^J
\pc{\by_{j}}{\bff_{j}}
\frac{
\pp{\bff_{j}   \bff_{\psB{j}}  }
}{
\pp{  \bff_{\psB{j}} }
}
\pc{\bff_{j}}{ \bff_{\prB{j}} }
\\
&=
\prod_{j=1}^J
\pc{\by_{j}}{\bff_{j}}
\pc{\bff_{j}}{ \bff_{\prB{j}} }
\end{align*}
since
\begin{align*}
\frac{
\pp{\bff_{j} ,  \bff_{\psB{j}}  }
}{
\pp{  \bff_{\psB{j}} }
}
&=
\frac{
\pp{\bff_{j},   \bff_{j },  \bff_{\psB{j} \setminus j }  }
}{
\pp{   \bff_{j },  \bff_{\psB{j} \setminus j }  }
}
= 1.
\end{align*}
\end{proofII}
\begin{proofII}[Proof of Prop.\ \ref{prop:equality}; Equality to Full GP ]
\label{proof:equality}
\textbf{Full GP:} 
For $\gamma=1$, the joint distribution of our model  is formulated 
in 
Def.\ 
\ref{def:jointDistr}
and 
Proof
\ref{proof:jointDistr_full}.
For $C=J$, we have
\begin{align*}
q_J(\bff,\by)
=
\prod_{j=1}^J
\pc{\by_{j}}{\bff_{j}}
\pc{\bff_{j}}{ \bff_{\bs{\pi}_J(j)} },
\end{align*}
where the predecessor set $\bs{\pi}_J(j)$ correspond to $\{1,\ldots,j-1\}$ and thus the conditional variables $\bff_{\prB{j}} = \bff_{1:j-1}$.
The posterior $q_J(\bff\vert\by)$ is proportional to the  
joint distribution $q_J(\bff,\by)$ (see Proof \ref{proof:posterior}), thus we have
\begin{align*}
q_J(\bff\vert \by)
\propto
\prod_{j=1}^J
\pc{\by_{j}}{\bff_{j}}
\pc{\bff_{j}}{ \bff_{1:j-1} }
\end{align*}
which is equal to the posterior distribution of full GP \eqref{eq:postFull}.
Also the hyperparameter optimization is the same since the marginal likelihood $q_J(\by)$ can be derived from the joint $q_C(\bff,\by)$ (see Proof  \ref{proof:margLik}).
Further, in the prediction step, for $C=J$ we have $J_2 = C-J+1 = 1$ predictive expert which is based on the full region $\psi(J)=\{1,\ldots,J\}$. 
 Therefore we conclude that the two models in considerations are the same.
\\
\quad
\\
\textbf{Sparse global GP:}
Similarly, for $C=J$ but $\gamma<1$, we have
\begin{align*}
q_J(\bff, \by)
&=
\prod_{j=1}^J
\pc{\by_{j}}{\bff_{j}}
\pc{\bff_{j}}{ \ba_{\psB{j}} }
\pc{\ba_{j}}{ \ba_{\prB{j}} }
\\
&=
\prod_{j=1}^J
\pc{\by_{j}}{\bff_{j}}
\pc{\bff_{j}}{ \ba_{ 1:J} }
\pc{\ba_{j}}{ \ba_{1:-j-1} }
\\
&=
\prod_{j=1}^J
\pc{\by_{j}}{\bff_{j}}
\pc{\bff_{j}}{ \ba }
\pp{\ba}
\\
&=
\pc{\by}{\bff}
\pc{\bff}{ \ba }
\pp{\ba}
\end{align*}
so that the posterior 
 correspond to that  of sparse GP in \eqref{se:sparse_GP}. The prediction simplifies also to 1 predictive expert based on the full region. Also the marginal likelihood is the same for $C=J$ and could be adapted as illustrated in Section \ref{se:variational}.
 \\
\quad
\\
\textbf{Independent local GP:}
For $C=\gamma=1$ we have
\begin{align*}
q_1(\bff,\by)
=
\prod_{j=1}^J
\pc{\by_{j}}{\bff_{j}}
\pc{\bff_{j}}{ \bff_{\oslash }}
=
\prod_{j=1}^J
\pc{\by_{j}}{\bff_{j}}
\pp{\bff_{j}}
\end{align*}
which is equal to \eqref{eq:PoE}. Prediction and hyperparameters similar as above. 
\end{proofII}

\begin{proofII}[Proof of Prop.\ \ref{prop:priorApprox};  Prior Approximation]
\label{proof:priorApprox}
Here we prove the first part for the prior over $\ba$, the second part is proved in 
Proof \ref{proof:likApprox}.

Using Prop.\ \ref{prop:priorApprox2} (with Proof \ref{proof:priorApprox2}), the prior $\q{\ba}$
can be equivalently written as
\begin{align*}
\prod_{j=1}^J
\NN{\ba_{\bs{\pi}^+(j)}}{\bO, \bS_{(j)}^{-1}}
%
\propto
-\frac{1}{2}
\ba_{\bs{\pi}^+(j)}^T
\tilde{\bF}_{j}^T
\bQ_{j}^{-1}
\tilde{\bF}_{j}
\ba_{\bs{\pi}^+(j)}
\end{align*}
with
$\bS_{(j)} = \tilde{\bF}_{j}^T \bQ_j^{-1} \tilde{\bF}_{j} \in \RR^{LC\times LC}$
and
$\tilde{\bF}_{j} = \begin{bmatrix}
-\bF_j & \II 
\end{bmatrix} \in \RR^{L\times LC} $.
This $LC$-dimensional Gaussian for $\ba_{\bs{\pi}^+(j)}$ can be augmented to a $M$-dimensional Gaussian for $\ba$ proportional to
\begin{align*}
&-\frac{1}{2}
\ba^T
\bar{\bF}_{j}^T
\bar{\bQ}_{j}^{-1}
\bar{\bF}_{j}
\ba
~~ \propto ~~
\NN{\ba}{\bO, \left( 
\bar{\bF}_{j}^T
\bar{\bQ}_{j}^{-1}
\bar{\bF}_{j}
\right)^{-1}}
\end{align*}
where
$\bar{\bQ}_{j}^{-1} \in \RR^{M\times M}$ a  zero matrix except $\bQ_j^{-1}\in \RR^{L\times L}$ at the entries
$[\bs{\pi}^+(j),\bs{\pi}^+(j)]$.
Further, the matrix $\bar{\bF}_j \in \RR^{M \times M}$ has one sparse row at $j$, that is,
\begin{align*}
\begin{bmatrix}
0 & 0 & 0 & 0 & 0 & 0 & 0 & 0 & 0 & 0 \\
\vdots & \vdots & \vdots & \vdots & \vdots & \vdots & \vdots & \vdots & \vdots & \vdots  \\
0
&
\cdots
&
-\bF_j^1
&
0
& 
-\bF_j^i
&
\cdots
& 
-\bF_j^{I_j}
& 
\II
&
\cdots
&
0
\\
\vdots & \vdots & \vdots & \vdots & \vdots & \vdots & \vdots & \vdots & \vdots & \vdots  \\
0 & 0 & 0 & 0 & 0 & 0 & 0 & 0 & 0 & 0
\end{bmatrix}
\end{align*}
where 
$\bF_j^i \in \RR^{L\times L} $ is the $i$th part of $\bF_j \in \RR^{L\times L(C-1)}$ which correspond to the contribution of the $i$th predecessor $\bs{\pi}^i(j)$.
\\
By using the property in \eqref{eq:prodGauss}, the original product
$q(\ba)$ is then
\begin{align*}
&~\prod_{j=1}^J \NN{\ba}{\bO, 
\left( 
\bar{\bF}_{j}^T
\bar{\bQ}_{j}^{-1}
\bar{\bF}_{j}
\right)^{-1}
}
=
\NN{\ba}{\bO,  \left( \sum_{j=1}^J 
\bar{\bF}_{j}^T
\bar{\bQ}_{j}^{-1}
\bar{\bF}_{j} \right)^{-1}}
\\
=
&~
\NN{\ba}{\bO, \left(
\bF^T \bQ^{-1} \bF
\right)^{-1}}
=
\NN{\ba}{\bO, \bS^{-1}}
\end{align*}
with
$\bQ^{-1} = Diag[\bQ_1^{-1},\ldots,\bQ_J^{-1}]$
and $\bF$ correspond then to the matrix depicted in Fig.\ \ref{fig:transF}.
Note that $\bS$ is positive definite since $\bQ^{-1}$  positive definite because each $\bQ_j^{-1}$ is positive definite
which concludes the proof. 
\end{proofII}
\begin{proofII}[(Sub)proof of Prop.\ \ref{prop:priorApprox} (Projection Approximation) ]
\label{proof:likApprox}
The 
 projection
$\qc{\bff}{\ba}
=
q_{C}(\bff\vert\ba)$ 
is
\begin{align*}
q_{C}(\bff\vert\ba)
=
\prod_{j=1}^J
\pc{\bff_{j}}{ \ba_{\psB{j}} }
=
\prod_{j=1}^J
\NN{\bff_j}{\bH_j\ba_{\psB{j}},\overline{\bV}_j},
\end{align*}
where
$\bH_j \in \RR^{B\times LC}$ and
$\overline{\bV}_j  \in \RR^{B\times B} $.
The log of this density  
in $\bff_j \in \RR^B$
is
proportional to
$$
\propto
-
\frac{1}{2}
( \bff_j - \bH_j \ba_{\bs{\psi}(j)})^T
\overline{\bV}_j^{-1}
( \bff_j - \bH_j \ba_{\bs{\psi}(j)})
$$
which can be equivalently written as
$$
-
\frac{1}{2}
(\II_j\bff - \bar{\bH}_j \ba)^T
\bar{\overline{\bV}}_j^{-1}
(\II_j\bff - \bar{\bH}_j \ba)
$$
 with $\bar{\overline{\bV}}_j \in \RR^{M \times M}$ with $\overline{\bV}_j$ at 
 $[\bs{\psi}(j),\bs{\psi}(j)]$ and
 $\bar{\bH}_j \in \RR^{BJ \times M}$ the following matrix 
$$
\begin{bmatrix}
0 & 0 & 0 & 0 & 0 & 0 & 0 & 0 & 0 & 0
\\
\vdots & \vdots & \vdots & \vdots & \vdots & \vdots & \vdots & \vdots & \vdots & \vdots  \\
0
&
\cdots
&
\bH_j^1
&
0
& 
\bH_j^i
&
\cdots
& 
\bH_j^{C-1}
& 
\bH_j^{C}
&
\cdots
&
0
\\
\vdots & \vdots & \vdots & \vdots & \vdots & \vdots & \vdots & \vdots & \vdots & \vdots  \\
0 & 0 & 0 & 0 & 0 & 0 & 0 & 0 & 0 & 0
\end{bmatrix}
$$
where $j$th row not empty
with $\bH_j^i \in \RR^{B \times L}$ the $i$th entry in $\bH_j$ which correspond to 
to $\bs{\psi}^i(j)$.
Further,
$\II_j\in \RR^{BJ \times BJ}$ a zero matrix with $\II \in \RR^{B\times B}$ at $[j,j]$.
For the original product of the projections
\begin{align*}
q_{C}(\bff\vert\ba)
&=
\prod_{j=1}^J
\NN{\bO}{\II_j^T\bff-\bar{\bH}_j\ba,\bar{\overline{\bV}_j}},
\end{align*}
using the product rule of Gaussians in \eqref{eq:prodGauss}, we obtain 
\begin{align*}
&\NN{\bO}{\overline{\bV} \sum_j^J 
\bar{\overline{\bV}}_j^{-1}
\left( \II_j\bff-\bar{\bH}_j\ba
\right)
,
\left(
\sum_j^J \bar{\overline{\bV}}_j^{-1}
\right)^{-1}
}
\\
=&~
\NN{\bO}{\overline{\bV} 
\overline{\bV}^{-1}
\sum_j^J 
\left( \II_j\bff-\bar{\bH}_j\ba
\right)
,
\overline{\bV}
}
=
\NN{\bO}{
\sum_j^J \left(
  \II_j \right)
\bff
-
\sum_j^J \left(
\bar{\bH}_j
\right)
\ba
,
\overline{\bV}
}
\\
=&~
\NN{\bO}{
  \II
\bff
-
\bH
\ba
,
\overline{\bV}
}
=
\NN{
\bff
}{
\bH
\ba
,
\overline{\bV}
}.
\end{align*}
Since $\overline{\bV}$ positive definite this concludes the statement.
\end{proofII}
\begin{proofII}[Proof of Prop.\ \ref{prop:convJointPrior}; Decreasing Prior KL]
\label{proof:convJointPrior}
We first show the decomposition 
which states 
\begin{align*}
&~\mathbb{D}_{(C,T)}[\bff, \ba, \by]
\\
=&~
\mathbb{D}_{(C,T)}[\ba]
+
\mathbb{D}_{(C,T)}[\bff\vert \ba]
+
\mathbb{D}_{(C,T)}[\by\vert \bff].
\end{align*}
Starting with the definition in Def.\ \ref{def:KL_diff} we get
\begin{align*}
&~\mathbb{D}_{(C,T)}[\bff, \ba, \by]
\\
=&~
\mathbb{E}_{p(\bff, \ba, \by)}\left[\log \frac{q_{C+T}(\bff, \ba, \by)}{q_C(\bff, \ba, \by)}\right]
\\
=&~
\mathbb{E}_{
p(\by\vert \bff) p(\bff\vert \ba) p(\ba)
}
\left[\log \frac{
q_{C+T}(\by\vert \bff) q_{C+T}(\bff\vert \ba) q_{C+T}(\ba)
}{
q_{C}(\by\vert \bff) q_{C}(\bff\vert \ba) q_{C}(\ba)
}\right]
\\
=&~
\int 
p(\by\vert \bff) p(\bff\vert \ba) p(\ba)
\log \frac{
q_{C+T}(\by\vert \bff) q_{C+T}(\bff\vert \ba) q_{C+T}(\ba)
}{
q_{C}(\by\vert \bff) q_{C}(\bff\vert \ba) q_{C}(\ba)
}
\diff \ba \diff \bff \diff \by
\\
=&~
\int  p(\ba) \Bigg( 
\int  p(\bff\vert \ba) \Bigg[
\int  p(\by\vert \bff) 
\log \frac{
q_{C+T}(\by\vert \bff)
}{
q_{C}(\by\vert \bff)
}\diff \by
~
\cdots 
\\
&
~ + ~
\log \frac{
q_{C+T}(\bff\vert \ba)
}{
q_{C}(\bff\vert \ba) 
}
\Bigg]
\diff \bff
+
\log \frac{
q_{C+T}(\ba)
}{
q_{C}(\ba) 
}
\Bigg)
\diff \ba
\\
=&~
\int  p(\ba)
\log \frac{
q_{C+T}(\ba)
}{
q_{C}(\ba) 
}
\diff \ba
+
\int  p(\ba)
\int  p(\bff\vert \ba) 
\log \frac{
q_{C+T}(\bff\vert \ba)
}{
q_{C}(\bff\vert \ba) 
}
\diff \bff
\\
&~
+
\int  p(\bff)
\int  p(\by\vert \bff) 
\log \frac{
q_{C+T}(\by\vert \bff)
}{
q_{C}(\by\vert \bff)
}\diff \by
\diff \bff
\\
=
&~
\mathbb{E}_{p(\ba)}
\left[
\log \frac{
q_{C+T}(\ba)
}{
q_{C}(\ba) 
}
\right]
+
\mathbb{E}_{p(\ba)}
\left[
\mathbb{E}_{p(\bff\vert\ba)}
\left[
\log \frac{
q_{C+T}(\bff\vert\ba)
}{
q_{C}(\bff\vert\ba) 
}
\right]
\right]
\\
&~
+
\mathbb{E}_{p(\bff)}
\left[
\mathbb{E}_{p(\by\vert\bff)}
\left[
\log \frac{
q_{C+T}(\by\vert\bff)
}{
q_{C}(\by\vert\bff) 
}
\right]
\right]
\\
=&~
\mathbb{D}_{(C,T)}[\ba]
+
\mathbb{D}_{(C,T)}[\bff\vert \ba]
+
\mathbb{D}_{(C,T)}[\by\vert \bff],
\end{align*}
where we used the definitions in Def.\ \ref{def:KL_diff}.
We also immediately see that $$\mathbb{D}_{(C,T)}[\by\vert \bff]=0$$ since $q_C(\by\vert\bff) = q_{C+T}(\by\vert\bff) = p(\by \vert\bff ) $ is exact.
The proofs for $\mathbb{D}_{(C,T)}[\ba]\geq 0$
and
$\mathbb{D}_{(C,T)}[\bff\vert \ba]\geq 0$
are given in Proof \ref{proof:pos1}, \ref{proof:pos2} and \ref{proof:pos2}, respectively.
\end{proofII}
\begin{proofII}[Proof of Subproof I of Proof \ref{proof:convJointPrior}]
\label{proof:pos1}
We prove
$$\mathbb{D}_{(C,T)}[\ba]
=
\mathbb{E}_{p(\ba)}\left[\log \frac{q_{C+T}(\ba)}{q_C(\ba)}\right]
\geq 0.$$
We abbreviate $q_1(\ba) = q_C(\ba)$ and $q_2(\ba) = q_{C+T}(\ba)$.
The difference 
 $\mathbb{D}_{(C,T)}[\ba]$ is
\begin{align*}
&
\quad\int 
p(\ba)
\log \frac{q_2(\ba)}{q_1(\ba)} \diff \ba
=\int 
p(\ba)
\log \frac{
\prod_{j=1}^J
\pc{\ba_{j}}{ \ba_{\bs{\pi}_2(j)} }
}{
\prod_{j=1}^J
\pc{\ba_{j}}{ \ba_{\bs{\pi}_1(j)} }
} \diff \ba
\\
&
=\int 
p(\ba)
\sum_{j=1}^J
\log \frac{
\pc{\ba_{j}}{ \ba_{\bs{\pi}_2(j)} }
}{
\pc{\ba_{j}}{ \ba_{\bs{\pi}_1(j)} }
} \diff \ba
=\sum_{j=1}^J
\int 
p(\ba)
\log \frac{
\pc{\ba_{j}}{ \ba_{\bs{\pi}_2(j)} }
}{
\pc{\ba_{j}}{ \ba_{\bs{\pi}_1(j)} }
} \diff \ba
\end{align*}
We recall property $(iii)$ in Def.\ \ref{def:predSets}, thus we have 
$\bs{\pi}_2(j) = \bs{\pi}_1(j) \cup \phi(j)$ where $\phi(j)$ is the additional predecessor of expert $j$ in the model $2$ compared to model $1$. In the following, we abbreviate $\bs{\pi}_1(j)=\bs{\pi}(j)$ yielding 
\begin{align*}
&
\sum_{j=1}^J
\int 
p(\ba)
\log \frac{
\pc{\ba_{j}}{ \ba_{\bs{\pi}(j)}, \ba_{\bs{\phi}(j)} }
}{
\pc{\ba_{j}}{ \ba_{\bs{\pi}(j)} }
} \diff \ba
\\
=&
\sum_{j=1}^J
\int 
p(\ba)
\log \frac{
\pc{\ba_{j}}{ \ba_{\bs{\pi}(j)}, \ba_{\bs{\phi}(j)} }
\pc{\ba_{\bs{\phi}(j)}}{ \ba_{\bs{\pi}(j)}  }
}{
\pc{\ba_{j}}{ \ba_{\bs{\pi}(j)} } 
\pc{\ba_{\bs{\phi}(j)}}{ \ba_{\bs{\pi}(j)}  }
} \diff \ba
\\
=&
\sum_{j=1}^J
\int 
p(\ba)
\log \frac{
\pc{\ba_{j}, \ba_{\bs{\phi}(j)}}{ \ba_{\bs{\pi}(j)}  }
}{
\pc{\ba_{j}}{ \ba_{\bs{\pi}(j)} } 
\pc{\ba_{\bs{\phi}(j)}}{ \ba_{\bs{\pi}(j)}  }
} \diff \ba
\\
=&
\sum_{j=1}^J
\int 
p(\tilde{\ba}_{j})
\log \frac{
\pc{\ba_{j}, \ba_{\bs{\phi}(j)}}{ \ba_{\bs{\pi}(j)}  }
}{
\pc{\ba_{j}}{ \ba_{\bs{\pi}(j)} } 
\pc{\ba_{\bs{\phi}(j)}}{ \ba_{\bs{\pi}(j)}  }
} \diff \tilde{\ba}_{j}
\\
=&
\sum_{j=1}^J
I\left( \ba_{j} , \ba_{\bs{\phi}(j)}\vert  \ba_{\bs{\pi}(j)} \right)
\geq 0
\end{align*}
where
$\tilde{\ba}_{j} = \ba_{j} \cup \ba_{\bs{\phi}(j)} \cup \ba_{\bs{\pi}(j)}$ and
$I\left( \ba_{j} , \ba_{\bs{\phi}(j)}\vert  \ba_{\bs{\pi}(j)} \right)$ the conditional mutual information which is always positive \cite[p.~30]{ajjanagadde2017lecture} and therefore concludes the first part of the proof.
\end{proofII}
\begin{proofII}[Subproof II of Proof \ref{proof:convJointPrior}]
\label{proof:pos2}
We prove
$$\mathbb{D}_{(C,T)}[\bff\vert \ba]
=
\mathbb{E}_{p(\ba)}\left[
\mathbb{E}_{p(\bff\vert\ba)}\left[
\log \frac{q_{C+T}(\bff\vert \ba)}{q_C(\bff\vert \ba)}
\right]\right]
\geq 0.$$
We abbreviate $q_1(\bff\vert \ba) = q_C(\bff\vert \ba)$ and $q_2(\bff\vert \ba) = q_{C+T}(\bff\vert \ba)$.
The difference 
 $\mathbb{D}_{(C,T)}[\bff\vert \ba]$ is
\begin{align*}
&
\quad\int 
p(\ba)
\int 
p(\bff\vert \ba)
\log \frac{q_2(\bff\vert \ba)}{q_1(\bff\vert \ba)} \diff \bff \diff \ba
=\int 
p(\ba,\bff)
\log \frac{
\prod_{j=1}^J
\pc{\bff_{j}}{ \ba_{\bs{\psi}_2(j)} }
}{
\prod_{j=1}^J
\pc{\bff_{j}}{ \ba_{\bs{\psi}_1(j)} }
} \diff \bff  \diff \ba
\\
&
=\int 
p(\ba,\bff)
\sum_{j=1}^J
\log \frac{
\pc{\bff_{j}}{ \ba_{\bs{\psi}_2(j)} }
}{
\pc{\bff_{j}}{ \ba_{\bs{\psi}_1(j)} }
} \diff \bff  \diff \ba
=\sum_{j=1}^J
\int 
p(\ba,\bff)
\log \frac{
\pc{\bff_{j}}{ \ba_{\bs{\psi}_2(j)} }
}{
\pc{\bff_{j}}{ \ba_{\bs{\psi}_1(j)} }
}\diff \bff  \diff \ba
\end{align*}
We recall  
the definition of $\bs{\psi}_C(j)$
in Def.\ \ref{def:corrSets} 
where we have $\bs{\psi}_C(j) = \bs{\pi}_C(j) ~\cup ~\{j,\ldots,C\}$ if $j < C$ and               $\bs{\psi}_C(j) = \bs{\pi}_C(j) ~\cup ~j $ otherwise. Further,
we have
$\bs{\pi}_2(j) = \bs{\pi}_1(j) \cup \phi(j)$ 
where $\phi(j)$ is the additional predecessor of expert $j$ in the model $2$ compared to model $1$.  Therefore, we have $\bs{\psi}_2(j) = \bs{\psi}_1(j)  ~\cup~ \phi(j)$ for all j.

[Proof:
If $j < C$, we have $\bs{\pi}_1(j) = \bs{\pi}_2(j)$ since $\phi(j)$ empty. Therefore, we have $\bs{\psi}_1(j) = \bs{\pi}_1(j) ~\cup ~\{j,\ldots,C\} =
 \bs{\pi}_2(j) ~\cup ~\{j,\ldots,C\} =\bs{\psi}_2(j)
$
for all $j = 1,\ldots,C-1$.
\\
If $j \geq C$, we have 
$\bs{\psi}_1(j) = \bs{\pi}_1(j) ~\cup ~j$ and
$\bs{\psi}_2(j) = \bs{\pi}_2(j) ~\cup ~j  
=\bs{\pi}_1(j) ~\cup~ \phi(j) ~\cup ~j 
= \bs{\psi}_1(j) ~\cup~ \phi(j)
$
for all $j = C,\ldots,J$.
]

We abbreviate $\bs{\psi}_1(j)=\bs{\psi}(j)$ and substitute $\bs{\psi}_2(j) = \bs{\psi}(j)  ~\cup~ \phi(j)$ yielding
\begin{align*}
&
\sum_{j=1}^J
\int 
p(\ba,\bff)
\log \frac{
\pc{\bff_{j}}{ \ba_{\bs{\psi}(j)}, \ba_{\bs{\phi}(j)} }
}{
\pc{\bff_{j}}{ \ba_{\bs{\psi}(j)} }
} \diff \bff \diff \ba
\\
=&
\sum_{j=1}^J
\int 
p(\ba,\bff)
\log \frac{
\pc{\bff_{j}}{ \ba_{\bs{\psi}(j)}, \ba_{\bs{\phi}(j)} }
\pc{\ba_{\bs{\phi}(j)}}{ \ba_{\bs{\psi}(j)}  }
}{
\pc{\bff_{j}}{ \ba_{\bs{\psi}(j)} } 
\pc{\ba_{\bs{\phi}(j)}}{ \ba_{\bs{\psi}(j)}  }
} \diff \ba
\\
=&
\sum_{j=1}^J
\int 
p(\ba,\bff)
\log \frac{
\pc{\bff_{j}, \ba_{\bs{\phi}(j)}}{ \ba_{\bs{\psi}(j)}  }
}{
\pc{\bff_{j}}{ \ba_{\bs{\psi}(j)} } 
\pc{\ba_{\bs{\phi}(j)}}{ \ba_{\bs{\psi}(j)}  }
} \diff \ba
\\
=&
\sum_{j=1}^J
\int 
p(\tilde{\ba}_{j},\bff_j)
\log \frac{
\pc{\bff_{j}, \ba_{\bs{\phi}(j)}}{ \ba_{\bs{\psi}(j)}  }
}{
\pc{\bff_{j}}{ \ba_{\bs{\psi}(j)} } 
\pc{\ba_{\bs{\phi}(j)}}{ \ba_{\bs{\psi}(j)}  }
} \diff \tilde{\ba}_{j}
\\
=&
\sum_{j=1}^J
I\left( \bff_{j} , \ba_{\bs{\phi}(j)}\vert  \ba_{\bs{\psi}(j)} \right)
\geq 0
\end{align*}
where
$\tilde{\ba}_{j} = \ba_{\bs{\phi}(j)} \cup \ba_{\bs{\psi}(j)}$ and
$I\left( \bff_{j} , \ba_{\bs{\phi}(j)}\vert  \ba_{\bs{\psi}(j)} \right)$ the conditional mutual information which is always positive \cite[p.~30]{ajjanagadde2017lecture} and therefore concludes the first part of the proof.
\end{proofII}

Moreover, the difference in the joint prior is
\begin{align*}
&~
\mathbb{D}_{(C,T)}[\bff, \ba, \by]
=
\mathbb{D}_{(C,T)}[\bff, \ba]
=
\mathbb{D}_{(C,T)}[\ba]
+
\mathbb{D}_{(C,T)}[\bff\vert \ba]
\\
=&~
\frac{1}{2}
\log
\frac{\vert\bar{\bV}_{C}\vert   \vert\bS_{C}^{-1}\vert
}{
\vert\bar{\bV}_{C+T} \vert\bS_{C+T}^{-1}\vert \vert
}
=
\frac{1}{2}
\log
\frac{\vert\bar{\bV}_{C}\vert   \vert\bQ_{C+T}\vert
}{
\vert\bar{\bV}_{C+T} \vert\bQ_{C}\vert \vert
}
\geq 0.
\end{align*}
\begin{proofII}[Subproof III of Proof \ref{proof:convJointPrior}; Prior KL]
\label{proof:convPrior}
%
For the second part, we use \eqref{eq:KLdiff} where the difference in KL of 2 Gaussians with zero mean and same base distribution is formulated. In our case we have
\begin{align*}
\frac{1}{2}\left(
tr((\bS_C-\bS_{C+T})\bK_{\bA\bA}) 
+ \log\frac{\vert\bS_{C+T}\vert}{\vert\bS_{C}\vert}
\right)
\end{align*}
where the trace is $0$ by Prop.\ \ref{prop:traceKaa} and thus 
$
\mathbb{D}_{(C,T)}=
\frac{1}{2}\log\frac{\vert\bS_{C+T}\vert}{\vert\bS_{C}\vert}
.$
Since $\bS_C = \bF^T \bQ^{-1} \bF$ and $\vert \bF\vert = 1$, we have
$
\mathbb{D}_{(C,T)}=
\frac{1}{2}\log\frac{\vert\bQ_{C}\vert}{\vert\bQ_{C+T}\vert}
$
which
concludes the proof.
\end{proofII}
\begin{proofII}[Proof of Prop.\ \ref{prop:decPriorEntropy}; Decreasing Prior Entropy]
\label{proof:decPriorEntropy}
For the third part of the statement, the entropy $H$ of $q_C(\ba)$ is 
\begin{align*}
H[q_C(\ba)]
=
\frac{1}{2}\left(
-\log \vert \bS_C \vert +
JL(1+\log 2\pi)
\right)
=
\frac{1}{2}\left(
\log \vert \bQ_C \vert +
JL(1+\log 2\pi)
\right)
\end{align*}
where we used Eq.\ \eqref{eq:entropy} and $\vert \bF \vert =1$.
The second part follows from Prop.\ \ref{prop:equality}.
Using 
Proof
\ref{proof:convPrior}
which states
\begin{align*}
\mathbb{D}_{(C,T)}=
\frac{1}{2}\log\frac{\vert\bQ_{C}\vert}{\vert\bQ_{C+T}\vert} \geq 0,
\end{align*}
it follows
\begin{align*}
\log \vert\bQ_{C}\vert \geq \log \vert\bQ_{C+T}\vert
\end{align*}
for any
$T\in\{1,\ldots,C-1\}$ and
therefore
\begin{align*}
H[ q_{C+T}(\ba)]  \leq H[q_{C}(\ba)]
\end{align*}
which concludes the proof.
\end{proofII}
\begin{proofII}[(Sub)Proof of Prop.\ \ref{prop:posterior}; Marginalized Joint Distribution]
\label{proof:jointDistr2}
From the joint distribution in Def.\ \ref{def:jointDistr} over all variables, the latent function values 
$\bff$ 
can be integrated out 
resulting in
\begin{align*}
\q{ \ba,\by}
&=
\int \q{\bff, \ba,\by}\diff \bff
=\int \pc{\by}{\bff}\qc{\bff}{\ba}  \diff \bff~ \q{\ba},
\end{align*}
where the integral can be computed  via \eqref{eq:int} yielding
\begin{align*}
\q{\ba, \by}
&=
\int \pc{\by}{\bff}\qc{\bff}{\ba}  \diff \bff
=
\int \NN{\by}{\bff,\sigma_n^2\II}\NN{\bff}{\bH\bff,\overline{\bV}} \diff \bff
=
\NN{\by}{\bH\ba,\bV}
\end{align*}
with
$\bV=\overline{\bV} + \sigma_n^2\II$
and thus
\begin{align*}
\q{ \ba,\by}
=\qc{\by}{\ba} \q{\ba}
=
\NN{\by}{\bH\ba,\bV} \NN{\ba}{\bO,\bS^{-1}}
\end{align*}
which concludes the proof.
\end{proofII}
\begin{proofII}[Proof of Prop.\ref{prop:posterior}; Posterior Approximation]
\label{proof:posterior}
The posterior approximation is
\begin{align*}
\qc{\ba}{\by}
&=
\frac{\q{ \ba,\by}}{\q{\by}}
\propto
\q{ \ba,\by}
=
\qc{\by}{\ba} q_C(\ba)
\end{align*}
where the first equality comes from the definition of conditional probabilities, the proportionality because the marginal likelihood $q(\by)$ is independent of $\ba$ and the last equality exploits Proof \ref{proof:jointDistr2}.
Since 
$$\qc{\by}{\ba} q_C(\ba)=
\NN{\by}{\bH\ba,\bV}
\NN{\ba}{\bO,\bS^{-1}},
$$
the desired posterior distribution can be analytically computed via \eqref{eq:GaB} 
yielding
$$
\qc{\ba}{\by} = \NN{\ba}{\bmu,\bSig},
$$
with
$\bSig
=\left(
\bH^T\bV^{-1}\bH
+
\bS
\right)^{-1}$,
$\bmu
= \bSig\bb$
and
$\bb = \bH^T\bV^{-1}\by$.
\end{proofII}
\begin{proofII}[Proof of Prop.\ \ref{prop:margLik}; Marginal Likelihood ]
\label{proof:margLik}
The 
marginal likelihood 
$q(\by) 
$ 
is   obtained 
by integrating \eqref{eq:int} over the 
joint
 distribution 
$\q{\by,\ba}$ in Prop.\ \ref{prop:jointDistr2}
leading to
\begin{align*}
  \q{\by}
 &=
\int \q{\by,\ba} \diff \ba
=
\int \qc{\by}{\ba} \q{\ba} \diff \ba
=
\int \NNo{\bH\ba,\bV}\NNo{\bO,\bS^{-1}}  \diff \ba
=
  \NNo{\bO,\bP}
\end{align*}
where
$\bP=\bH\bS^{-1}\bH^T +\bV$.
\end{proofII}
\begin{proofII}[Proof of Prop.\ \ref{prop:traceKaa}; Exact Diagonal of Prior ]
\label{proof:traceKaa}
Using Prop.\ \ref{prop:priorApprox3}, the trace can be written as
\begin{align*}
&tr\left(\bS \bK_{\bA\bA}\right)
=
~tr\left( \left(
\sum_{j=1}^J \overline{\bK}^{-1}_{\bA_{\bs{\pi}^+(j)}\bA_{\bs{\pi}^+(j)}}
-
\overline{\bK}^{-1}_{\bA_{\prB{j}}\bA_{\prB{j}}}
\right)  \bK_{\bA\bA} \right)
\\
=
&
\sum_{j=1}^J tr\left(
\overline{\bK}^{-1}_{\bA_{\bs{\pi}^+(j)}\bA_{\bs{\pi}^+(j)}}
\bK_{\bA\bA} 
\right)
-
 tr\left(
\overline{\bK}^{-1}_{\bA_{\prB{j}}\bA_{\prB{j}}}
\bK_{\bA\bA}
\right).
\end{align*}
By construction of the matrices 
$\overline{\bK}^{-1}_{\bA_{\phi},\bA_{\phi}}$, they contain the matrix $\bK^{-1}_{\bA_{\phi},\bA_{\phi}}$ at the entries $[\phi,\phi]$. Therefore, the resulting product when multiplying with $\bK_{\bA\bA}$ is
a matrix with identity $\II_{T}$ at the position $[\phi,\phi]$ with
$T = \vert \phi \vert$ and 0 at the diagonal where not $\phi$.
The quantity above is then
\begin{align*}
&
\sum_{j=1}^J L\vert\bs{\pi}^+(j) \vert 
-
L\vert \prB{j} \vert
=
\sum_{j=1}^J L \left( min(j,C) 
-
min(j-1,C-1) \right)
=
JL.
\end{align*}
\end{proofII}
\begin{proofII}[Proof of Prop.\ \ref{prop:priorApprox2}; Prior Approximation II]
\label{proof:priorApprox2}
The prior approximation is
\begin{align*}
q(\ba)
=
\prod_{j=1}^J
\pc{\ba_{j}}{ \ba_{\prB{j}} }
=
\prod_{j=1}^J
\NN{\ba_j}{\bF_j\ba_{\prB{j}},\bQ_j}
\end{align*}
for which
the quadratic term inside the exponential of the individual Gaussian can be written as
\begin{align*}
&-\frac{1}{2}
( \ba_j- \bF_j\ba_{\prB{j}} )^T \bQ_j^{-1} 
( \ba_j- \bF_j\ba_{\prB{j}} )
\\
&
-\frac{1}{2}
\begin{bmatrix}
\ba_{\prB{j}}^T & \ba_j^T
\end{bmatrix} 
\begin{bmatrix}
-\bF_j^T \\ \II 
\end{bmatrix}
\bQ_j^{-1}  
\begin{bmatrix}
-\bF_j & \II 
\end{bmatrix}
\begin{bmatrix}
\ba_{\prB{j}} \\ \ba_j
\end{bmatrix} 
\end{align*}
which correspond to a Gaussian
\begin{align*}
&\NN{\ba_{\bs{\pi}^+(j)}}{\bO, \bS_{(j)}^{-1}}
\end{align*}
with
$
\bS_{(j)} = \tilde{\bF}_{j}^T \bQ_j^{-1} \tilde{\bF}_{j}  \in \RR^{LC\times LC}
$
and
$
\tilde{\bF}_{j} = \begin{bmatrix}
-\bF_j & \II 
\end{bmatrix}
 \in \RR^{L\times LC}
$
which proves the first part.
We can augment this Gaussian for  $\ba_{\bs{\pi}^+(j)}\in \RR^{LC}$ to
\begin{align*}
&-\frac{1}{2}
\ba^T
\bar{\bS}_{(j)}^{-1}
\ba
\quad \propto \quad
\NN{\ba}{\bO, \bar{\bS}_{(j)}^{-1}}
\end{align*}
over $\ba\in \RR^{M}$
where
$\bar{\bS}_{(j)} \in \RR^{M\times M}$
is the augmented matrix consisting of 
$\bS_{(j)}$
at the entries $[\bs{\pi}^+(j),\bs{\pi}^+(j)]$ and $0$ otherwise.
Using \eqref{eq:prodGauss}, the original product $q(\ba)$ is then 
\begin{align*}
\prod_{j=1}^J
\NN{\ba}{\bO, \bar{\bS}_{(j)}^{-1}}
=
\NN{\ba}{\bO, \left( \sum_{j=1}^J \bar{\bS}_{(j)} \right)^{-1}}
\end{align*}
and thus $\bS = \sum_{j=1}^J \bar{\bS}_{(j)}$ positive definite
which concludes the proof.
\end{proofII}
\begin{proofII}[Proof of Prop.\ \ref{prop:priorApprox3}; Prior Approximation III]
\label{proof:priorApprox3}
The prior $q(\ba)$ can be written as
\begin{align*}
q(\ba)
&=
\prod_{j=1}^J
\pc{\ba_{j}}{ \ba_{\prB{j}} }
=\prod_{j=1}^J
\frac{
\pp{\ba_{j}, \ba_{\prB{j}} } 
}{
\pp{\ba_{\prB{j}}}
}
=
\prod_{j=1}^J
\frac{
\pp{ \ba_{\bs{\pi}^+(j)} } 
}{
\pp{\ba_{\prB{j}}}
}
\\
&
=\prod_{j=1}^J
\frac{
\NN{\ba_{\bs{\pi}^+(j)} }{\bs{0},
\bK_{\bA_{\bs{\pi}^+(j)} \bA_{\bs{\pi}^+(j)}}
}
}{
\NN{\ba_{\prB{j}} }{\bs{0},
\bK_{\bA_{\prB{j}} \bA_{\prB{j}}}
}
}
\end{align*}
Similarly to the Proof \ref{proof:priorApprox2}, we can augment the $CL$-dimensional and the $(C-1)L$-dimensional Gaussian in the nominator and denominator, respectively, to $M$-dimensional Gaussians with covariance
$\overline{\bK}^{-1}_{\bA_{\bs{\phi}} \bA_{\bs{\phi}}}$ consisting of 
$\bK^{-1}_{\bA_{\bs{\phi}} \bA_{\bs{\phi}}}$
at the entries $[\bs{\phi},\bs{\phi}]$ and $0$ otherwise.
This gives with \eqref{eq:prodGauss}
\begin{align*}
&\prod_{j=1}^J
\frac{
\NN{\ba }{\bs{0},
\overline{\bK}_{\bA_{\bs{\pi}^+(j)} \bA_{\bs{\pi}^+(j)}}
}
}{
\NN{\ba}{\bs{0},
\overline{\bK}_{\bA_{\prB{j}} \bA_{\prB{j}}}
}
}
\\
=&\NN{\ba }{\bs{0},
\left(
\sum_{j=1}^J
\overline{\bK}_{\bA_{\bs{\pi}^+(j)} \bA_{\bs{\pi}^+(j)}}^{-1}
-
\overline{\bK}_{\bA_{\prB{j}} \bA_{\prB{j}}}^{-1}
\right)^{-1}
},
\end{align*}
which concludes the proof with 
$\bS =
\sum_{j=1}^J
\overline{\bK}_{\bA_{\bs{\pi}^+(j)} \bA_{\bs{\pi}^+(j)}}^{-1}
-
\overline{\bK}_{\bA_{\prB{j}} \bA_{\prB{j}}}^{-1}$
which is positive definite.
\end{proofII}
\begin{proofII}[Proof of Prop.\ \ref{prop:predAgg}; Prediction Aggregation]
\label{proof:predAgg}
The predictive posterior distribution is defined as
\begin{align*}
\pc{f_*}{\by}
=
\prod_{j=C}^{J}
\pc{f_{*j}}{\by}
^{\beta_{*j}}.
\end{align*}
Since the local predictions
$\pc{f_{*j}}{\by}=\NNo{m_{*j},v_{*j}}$ 
are all univariate Gaussians, we obtain via
the product rule of Gaussians in \eqref{eq:prodGauss}
 directly
\begin{align*} 
\begin{split}
m_*
&
=
v_{*j}
\sum_{j=C}^{J}
\beta_{*j}
\frac{m_{*j}}{v_{*j}}
\quad\text{  and   }\quad
\frac{1}{v_*}
=
\sum_{j=C}^{J}
\frac{\beta_{*j}}{v_{*j}}.
\end{split}
\end{align*} 
Using the usual likelihood 
$\pc{y_*}{f_*} = \NNo{f_*,\sigma_n^2}$
yields with \eqref{eq:int}
 the  final noisy prediction 
 $\pc{y_*}{\by}=
 \int \pc{y_*}{f_*} \pc{f_*}{\by} \diff f_*
 =
 \NNo{m_*,v_* + \sigma_n^2}$.
\end{proofII}

\begin{proofII}[Proof of Prop.\ \ref{prop:locPred}; Local Predictions]
\label{proof:locPred}
The predictive conditional
$\pc{f_{*j}}{\ba_{\psB{j}}}$
can be again derived via \eqref{eq:condGauss} from the assumed joint 
$$\pp{f_{*j},\ba_{\psB{j}}}
=
\NNo{\bO,\Kab{[\bx_*,\bA_{\psB{j}}]}{[\bx_*,\bA_{\psB{j}}]}}$$
leading
to
$\NNo{\bh_*\ba_{\psB{j}},v_*}$ 
with 
$$\bh_*=\Kab{\bx_*}{\bA_{\psB{j}}}
\Kab{\bA_{\psB{j}}}{\bA_{\psB{j}}}^{-1}
$$
and
$$v_*=\Kab{\bx_*}{\bx_*}
-
\Kab{\bx_*}{\bA_{\psB{j}}}
\Kab{\bA_{\psB{j}}}{\bA_{\psB{j}}}^{-1}
\Kab{\bA_{\psB{j}}}{\bx_*}.
$$
%
%
Moreover, 
the local posteriors
$
\qc{\ba_{\psB{j}}}{\by}
=
\NNo{\bmu_{\psB{j}}, \bSig_{\psB{j}}}
$
are obtained 
from the corresponding entries $\bs{\psi}(j)$ of the mean $\bmu$ and covariance $\bSig$ (via partial inversion \ref{se:systemAndInv}) in Prop,\ \ref{prop:posterior}.
Finally, the local predictions $\pc{f_{*j}}{\by}$ in Prop.\ \eqref{prop:locPred} can then be computed with Gaussian integration \eqref{eq:int} yielding
$$
\qc{f_{*j}}{\by}
=
\int
\pc{f_{*j}}{\ba_{\psB{j}}}
\pc{\ba_{\psB{j}}}{\by}
\diff
\ba_{\psB{j}}
$$
which correspond to the
 desired quantities
$$\NNo{m_{*j},v_{*j}}=
\NNo{\bh_*\bmu_{\psB{j}},
\bh_*^T\bSig_{\psB{j}} \bh_*+v_*}.$$
\end{proofII}

\begin{proofII}[Proof for Figure \ref{fig:joint_prior_wo_star}; Joint Prior Covariance]
\label{proof:priorCovariance}
For the joint prior 
\begin{align*}
q_C(\ba,\bff,\by)
&=
\NN{[\ba;\bff;\by]~}{~\bO,\bW_{\gamma}}
\end{align*}
with covariance 
$$\bW_{\gamma} =
\begin{bmatrix}
\bSig_{\ba\ba} & \bSig_{\ba\bff} & \bSig_{\ba\by}\\
\bSig_{\bff\ba} & \bSig_{\bff\bff} & \bSig_{\bff\by}\\
\bSig_{\by\ba} & \bSig_{\by\bff} & \bSig_{\by\by}
\end{bmatrix}
$$ corresponding to Fig.\  \ref{fig:joint_prior_wo_star}, we show that we recover the marginal and conditional distributions $q_C(\ba)$, $q_C(\bff\vert\ba)$ and 
$p(\by\vert \bff)$.
For $q_C(\ba)$, the marginalization correspond to selecting the corresponding mean and covariance, i.e.
$\NN{\ba}{\bO,\bSig_{\ba\ba}}
=\NN{\ba}{\bO,\bS^{-1}}$.
For $q_C(\bff\vert\ba)$, we use Eq.\ \eqref{eq:condGauss} yielding
\begin{align*}
&\NN{\bff}{\bSig_{\bff\ba} \bSig_{\ba\ba}^{-1} \ba,
\bSig_{\bff\bff} - \bSig_{\bff\ba} \bSig_{\ba\ba}^{-1} \bSig_{\ba\bff}
 }
 \\
 &=\NN{\bff}{\bH \ba,
(\bH \bS^{-1} \bH^T + \overline{\bV}) - \bH (\bS^{-1} \bH^T }
 \\
 &=\NN{\bff}{\bH \ba,
 \overline{\bV} }
\end{align*}
since $\bSig_{\bff\ba} \bSig_{\ba\ba}^{-1}
= 
(\bH \bS^-1) \bS = \bH$.
Similarly for 
$p(\by\vert\bff)$, with 
Eq.\ \eqref{eq:condGauss} we get
\begin{align*}
&\NN{\by}{\bSig_{\by\bff} \bSig_{\bff\bff}^{-1} \bff,
\bSig_{\by\by} - \bSig_{\by\bff} \bSig_{\bff\bff}^{-1} \bSig_{\bff\by}
 }
 \\
 &=\NN{\by}{\II \bff,
(\bH \bS^{-1} \bH^T + \bV) - \II ( \bH \bS^{-1} \bH^T + \overline{ \bV })  }
 \\
 &=\NN{\by}{\bff,
 \sigma_n^2 \II }
\end{align*}
since $\bSig_{\by\bff} \bSig_{\bff\bff}^{-1} = \II.$ 
\end{proofII}
%
%

\subsection{Derivative of LML }
\label{se:derivatives}
The log marginal likelihood in Section \ref{se:hyperEst} in Eq.\ \eqref{eq:margLikDet}
is proportional to 
\begin{align*}
-\frac{1}{2}
\by^T\bV^{-1}\by 
+\frac{1}{2}\bmu^T\bSig^{-1}\bmu
-\frac{1}{2}\log
\vert\bSig^{-1}\vert
-\frac{1}{2}\log
  \vert\bV\vert
-\frac{1}{2}\log
\vert\bQ\vert
.
\end{align*}
In the following, we provide the partial derivative with respect to $\theta$ for  each additive term.
\begin{align*}
\frac{\partial}{\partial\theta}\left[
 -\frac{1}{2}
\by^T\bV^{-1}\by 
\right]
=
\frac{1}{2}
\by^T\bV^{-1} \frac{\partial \bV}{\partial\theta}\bV^{-1}\by 
\end{align*}
\begin{align*}
\frac{\partial}{\partial\theta}\left[
\frac{1}{2}\bmu^T\bSig^{-1}\bmu
 \right]
=
\frac{\partial \bb^T}{\partial\theta}
\bmu
-
\frac{1}{2}
\bmu^T  \frac{\partial \bSig^{-1}}{\partial\theta}\bmu
\end{align*}
\begin{align*}
\frac{\partial}{\partial\theta}\left[
-\frac{1}{2}\log \vert \bSig^{-1}\vert
 \right]
=
-
\frac{1}{2}
tr
\left\{ 
\bSig
\frac{\partial \bSig^{-1}}{\partial\theta}
\right\}
\end{align*}
In the last expression the whole posterior covariance is needed, however, it turns out that only the entries which are non-zero in the precision are needed. 
The right term in the last expression equals $
sum
\left\{ 
\bSig \odot
\frac{\partial \bSig^{-1}}{\partial\theta}
\right\}$, where $\odot$ denotes the pointwise multiplication. Therefore it is enough to only compute 
$sum\left\{ 
\overline{\bSig} \odot
\frac{\partial \bSig^{-1}}{\partial\theta}
\right\}$, where $\overline{\bSig}$ is the partial inversion (for more derails \ref{se:systemAndInv}) which is sparse as well and already computed for the local predictions in Prop.\ \ref{prop:locPred}.
\begin{align*}
\frac{\partial}{\partial\theta}\left[
-\frac{1}{2}\log \vert \bV\vert
 \right]
=
-
\frac{1}{2}
sum
\left\{ 
\bV^{-1} \odot
\frac{\partial \bV}{\partial\theta}
\right\}
\end{align*}
\begin{align*}
\frac{\partial}{\partial\theta}\left[
-\frac{1}{2}\log \vert \bQ\vert
 \right]
=
-\frac{1}{2}
sum
\left\{ 
\bQ^{-1} \odot
\frac{\partial \bQ}{\partial\theta}
\right\}
\end{align*}
The derivatives 
$\frac{\partial \bSig^{-1}}{\partial\theta}$
,
$\frac{\partial \bV}{\partial\theta}$ and
$\frac{\partial \bQ}{\partial\theta}$
can be computed via chain rule of derivatives.

    \section{Sequential  Algorithm}
  \label{se:stateSpace}
  
The probabilistic equations in Section \ref{se:CPoE} can be equivalently formulated as
\begin{align*}
\ba_j &= \bF_{j} \ba_{\prB{j}} +  \bga_j
;\\
\bff_j &= \bH_j \ba_{\psB{j}} + \bnu_j
;\\
\by_j &= \bff_j + \bs{\varepsilon}_j,
\end{align*}
with
$\bga_j \sim \NNo{0,\bQ_{j}}$,
$\bnu_j \sim \NNo{0,\overline{\bV}_j}$
and
$\bs{\varepsilon}_j \sim \NNo{0,\sigma^2_n\II}$.
Instead to the inference procedure described in Prop.\ \ref{prop:posterior}, the posterior could be alternatively computed with sequential algorithms.
Assuming $C=2$ and $\prB{j} = \{j-1\}$, the Kalman Filter and Smoother
(e.g. \cite{murphy2012machine})
 provide an  equivalent solution to the posterior distribution in Prop.\ \ref{prop:posterior}.
 For  $C>2$ and general neighbourhood set, the Gaussian loopy belief propagation algorithm or Gaussian expectation propagation (e.g. \cite{murphy2012machine}) might constitute an interesting approach for sequential/online and distributed learning procedures exploited in future work.

\section{Tables}
\label{se:tables}
Here we provide more results for the experiments in Section \ref{se:examples_evaluation} and the datasets in Table \ref{tab:datasets}.
In the following, we report different average quantities for  several test points $\bx_*,y_*$  corresponding to the predictive distributions $\pc{y_*}{\by}= \NNo{m_{*},v_*}$. 
The considered quantities are \textit{Kullback-Leibler-(KL)-divergence (KL)}  to full GP,  \textit{Continuous Ranked Probability Score (CRPS)}  and \textit{95\%-coverage (COV)},  \textit{root mean squared error (RMSE)}, \textit{absolut error (ABSE)}, \textit{negative log probability (NLP) }, root mean squared error to full GP (ERR) and
\textit{log marginal likelihood (LML)}.
\\
We use the KL to compare the closeness of predictive distributions of different GP approximation models  to the one of full GP $\NNo{m,v}$.
Since both are univariate Gaussians, the 
$KL\left(
 \NNo{m,v}
 \parallel
 \NNo{m_{*},v_*}
\right)$
can be computed as 
$\frac{1}{2}\left(
\log{\frac{v_*}{v}}
+ \frac{v}{v_*}
+
\frac{(m-m_*)^2}{v_*}
-1
\right)$.
\\
The CRPS can be used to assess the respective accuracy of two probabilistic forecasting models. In particular, it is a measure between the forecast CDF $F_*$ of $\NNo{m_{*},v_*}$  and the empirical CDF of the observation $\by_*$ and 
is defined as 
$CRPS(F_*, y_*)
\int \left( F(z) - 
1
_{z\geq y_* }\right)^2 \diff z $.
\\
The  \textit{95\%-confidence interval} can be computed as $c_{1,2}=m_* \pm 1.96 \sqrt{v_*}$. The 95\%-coverage is then defined as 
$COV = 
1
_{c_1 \leq y_* \leq c_2}$.
\\
The negative log probability is 
$-\pc{y_*}{\by} = \frac{1}{2}\log{(2\pi v_*)}+\frac{(y_*-m_*)^2}{2v_*}$. 
\\
For all quantities except LML (large values are better) and COV (should be close to 0.95), small values mean better predictions.

  \begin{landscape}

\begin{table*}
\begin{small}
\centering
\begin{tabular}{l | lllllllll}
\toprule
{} &            time &                LML &                KL &                ERR &               CRPS &               RMSE &               ABSE &              NLP &              COV \\
\midrule
fullGP   &   7.3 $\pm$ 0.6 &   -314.2 $\pm$ 5.1 &     0.0 $\pm$ 0.0 &      0.0 $\pm$ 0.0 &  0.162 $\pm$ 0.004 &  0.311 $\pm$ 0.011 &  0.218 $\pm$ 0.005 &  0.47 $\pm$ 0.12 &  0.92 $\pm$ 0.01  \\
\hline
SGP(25)  &   6.4 $\pm$ 0.6 &  -595.4 $\pm$ 10.7 &  440.3 $\pm$ 19.6 &  0.314 $\pm$ 0.008 &  0.234 $\pm$ 0.005 &   0.422 $\pm$ 0.01 &  0.324 $\pm$ 0.005 &  1.11 $\pm$ 0.04 &  0.96 $\pm$ 0.01 \\
SGP(50)  &  14.5 $\pm$ 2.6 &  -539.6 $\pm$ 10.2 &  405.0 $\pm$ 31.3 &  0.291 $\pm$ 0.012 &  0.222 $\pm$ 0.004 &  0.402 $\pm$ 0.008 &  0.308 $\pm$ 0.005 &  1.01 $\pm$ 0.03 &  0.95 $\pm$ 0.01 \\
SGP(100) &  36.4 $\pm$ 2.9 &   -494.6 $\pm$ 7.8 &  352.9 $\pm$ 29.5 &  0.264 $\pm$ 0.011 &  0.211 $\pm$ 0.004 &  0.384 $\pm$ 0.007 &  0.292 $\pm$ 0.006 &  0.92 $\pm$ 0.03 &  0.95 $\pm$ 0.01 \\
minVar   &   1.5 $\pm$ 0.1 &   -389.8 $\pm$ 2.9 &  122.2 $\pm$ 13.1 &  0.156 $\pm$ 0.012 &  0.175 $\pm$ 0.004 &  0.335 $\pm$ 0.011 &  0.236 $\pm$ 0.005 &  0.61 $\pm$ 0.09 &  0.92 $\pm$ 0.01 \\
GPoE     &   1.4 $\pm$ 0.1 &   -389.8 $\pm$ 2.9 &   174.4 $\pm$ 9.4 &   0.166 $\pm$ 0.01 &  0.186 $\pm$ 0.004 &   0.342 $\pm$ 0.01 &  0.255 $\pm$ 0.007 &  0.68 $\pm$ 0.05 &  0.96 $\pm$ 0.01 \\
BCM      &   1.4 $\pm$ 0.1 &   -389.8 $\pm$ 2.9 &  338.1 $\pm$ 32.7 &  0.185 $\pm$ 0.012 &  0.195 $\pm$ 0.005 &   0.354 $\pm$ 0.01 &  0.265 $\pm$ 0.007 &  1.16 $\pm$ 0.12 &  0.82 $\pm$ 0.01 \\
RBCM     &   1.4 $\pm$ 0.1 &   -389.8 $\pm$ 2.9 &  427.9 $\pm$ 35.0 &  0.166 $\pm$ 0.013 &  0.187 $\pm$ 0.005 &  0.342 $\pm$ 0.011 &  0.249 $\pm$ 0.006 &  1.43 $\pm$ 0.21 &  0.79 $\pm$ 0.01 \\
GRBCM    &   1.7 $\pm$ 0.1 &   -465.0 $\pm$ 3.1 &  224.6 $\pm$ 30.3 &  0.202 $\pm$ 0.011 &   0.19 $\pm$ 0.004 &   0.352 $\pm$ 0.01 &  0.262 $\pm$ 0.006 &  0.71 $\pm$ 0.05 &  0.92 $\pm$ 0.01 \\
\textbf{CPoE(1)}  &   1.5 $\pm$ 0.0 &   -397.0 $\pm$ 2.8 &  111.1 $\pm$ 12.5 &  0.146 $\pm$ 0.011 &  0.175 $\pm$ 0.004 &  0.333 $\pm$ 0.011 &  0.237 $\pm$ 0.006 &  \textbf{0.59 $\pm$ 0.09} &  0.93 $\pm$ 0.01 \\
\textbf{CPoE(2)}  &   2.1 $\pm$ 0.1 &   -345.1 $\pm$ 5.6 &   89.6 $\pm$ 14.3 &  0.124 $\pm$ 0.013 &  0.172 $\pm$ 0.004 &  0.326 $\pm$ 0.011 &  0.232 $\pm$ 0.006 &    0.6 $\pm$ 0.1 &  0.91 $\pm$ 0.01 \\
\textbf{CPoE(3)}  &   2.5 $\pm$ 0.1 &   -337.0 $\pm$ 5.5 &   82.2 $\pm$ 14.3 &  0.116 $\pm$ 0.013 &   \textbf{0.17 $\pm$ 0.004} &   \textbf{0.323 $\pm$ 0.01} &  0.231 $\pm$ 0.005 &   \textbf{0.59 $\pm$ 0.1} &  0.91 $\pm$ 0.01 \\
\textbf{CPoE(4)}  &   2.8 $\pm$ 0.1 &   -339.4 $\pm$ 5.0 &   \textbf{79.5 $\pm$ 13.9} & \textbf{ 0.111 $\pm$ 0.012} &  0.171 $\pm$ 0.004 &  0.324 $\pm$ 0.011 &  0.232 $\pm$ 0.005 &    0.6 $\pm$ 0.1 &  0.91 $\pm$ 0.01 \\
\bottomrule
\end{tabular}

\caption{Results for dataset \textit{concrete}.}
\end{small}
\end{table*}

\begin{table*}
\begin{small}
\centering
\begin{tabular}{l |  lllllllll}
\toprule
{} &            time &                LML &                 KL &                ERR &               CRPS &               RMSE &               ABSE &              NLP &             COV \\
\midrule
fullGP   &  25.5 $\pm$ 1.1 &   -994.2 $\pm$ 1.1 &      0.0 $\pm$ 0.0 &      0.0 $\pm$ 0.0 &  0.283 $\pm$ 0.002 &  0.511 $\pm$ 0.004 &   0.39 $\pm$ 0.005 &  1.49 $\pm$ 0.02 &  0.94 $\pm$ 0.0 \\ \hline
SGP(25)  &   7.5 $\pm$ 0.7 &  -1082.8 $\pm$ 0.9 &   93.49 $\pm$ 3.86 &  0.232 $\pm$ 0.005 &  0.316 $\pm$ 0.003 &  0.561 $\pm$ 0.004 &  0.445 $\pm$ 0.005 &  1.68 $\pm$ 0.02 &  0.94 $\pm$ 0.0 \\
SGP(50)  &   9.7 $\pm$ 1.4 &  -1042.7 $\pm$ 5.2 &    41.4 $\pm$ 5.59 &  0.146 $\pm$ 0.012 &  0.299 $\pm$ 0.003 &  0.537 $\pm$ 0.003 &  0.416 $\pm$ 0.006 &  1.59 $\pm$ 0.01 &  0.94 $\pm$ 0.0 \\
SGP(100) &  14.4 $\pm$ 0.8 &  -1009.6 $\pm$ 1.2 &    9.86 $\pm$ 1.73 &  0.069 $\pm$ 0.006 &  0.285 $\pm$ 0.002 &  0.514 $\pm$ 0.004 &  0.395 $\pm$ 0.005 &  1.51 $\pm$ 0.02 &  0.94 $\pm$ 0.0 \\
minVar   &   2.0 $\pm$ 0.2 &  -1025.8 $\pm$ 1.1 &   19.39 $\pm$ 1.78 &  0.101 $\pm$ 0.005 &  \textbf{0.282 $\pm$ 0.002} &  \textbf{0.508 $\pm$ 0.005} &   \textbf{0.39 $\pm$ 0.003} &  \textbf{1.48 $\pm$ 0.02} &  0.93 $\pm$ 0.0 \\
GPoE     &   1.9 $\pm$ 0.1 &  -1025.8 $\pm$ 1.1 &   54.22 $\pm$ 1.64 &  0.162 $\pm$ 0.003 &  0.301 $\pm$ 0.002 &  0.535 $\pm$ 0.004 &  0.424 $\pm$ 0.006 &   1.6 $\pm$ 0.01 &  0.96 $\pm$ 0.0 \\
BCM      &   1.9 $\pm$ 0.1 &  -1025.8 $\pm$ 1.1 &  257.61 $\pm$ 8.81 &  0.209 $\pm$ 0.005 &  0.313 $\pm$ 0.003 &  0.555 $\pm$ 0.006 &  0.422 $\pm$ 0.004 &  2.02 $\pm$ 0.04 &  0.82 $\pm$ 0.0 \\
RBCM     &   1.9 $\pm$ 0.1 &  -1025.8 $\pm$ 1.1 &   38.35 $\pm$ 1.56 &  0.132 $\pm$ 0.003 &  0.295 $\pm$ 0.003 &  0.528 $\pm$ 0.005 &  0.408 $\pm$ 0.005 &  1.56 $\pm$ 0.02 &  0.92 $\pm$ 0.0 \\
GRBCM    &   2.3 $\pm$ 0.2 &  -1048.9 $\pm$ 1.7 &   69.12 $\pm$ 6.48 &   0.196 $\pm$ 0.01 &  0.307 $\pm$ 0.004 &  0.551 $\pm$ 0.007 &  0.431 $\pm$ 0.006 &  1.64 $\pm$ 0.02 &  0.94 $\pm$ 0.0 \\
\textbf{CPoE(1)}  &   2.1 $\pm$ 0.1 &  -1025.8 $\pm$ 1.1 &   12.18 $\pm$ 0.92 &  0.079 $\pm$ 0.003 &  0.284 $\pm$ 0.002 &   0.51 $\pm$ 0.004 &  0.393 $\pm$ 0.003 &  1.49 $\pm$ 0.02 &  0.94 $\pm$ 0.0 \\
\textbf{CPoE(2)}  &   2.8 $\pm$ 0.1 &  -1010.1 $\pm$ 1.5 &    8.44 $\pm$ 0.66 &  0.066 $\pm$ 0.003 &  0.285 $\pm$ 0.002 &  0.512 $\pm$ 0.004 &  0.394 $\pm$ 0.004 &   1.5 $\pm$ 0.02 &  0.93 $\pm$ 0.0 \\
\textbf{CPoE(3)}  &   3.1 $\pm$ 0.1 &  -1007.0 $\pm$ 1.5 &    7.83 $\pm$ 0.58 &  0.064 $\pm$ 0.002 &  0.285 $\pm$ 0.002 &  0.513 $\pm$ 0.004 &  0.394 $\pm$ 0.004 &   1.5 $\pm$ 0.02 &  0.93 $\pm$ 0.0 \\
\textbf{CPoE(4)}  &   3.3 $\pm$ 0.1 &  -1004.8 $\pm$ 1.5 &   \textbf{ 7.59 $\pm$ 0.63} &  \textbf{0.062 $\pm$ 0.003} &  0.285 $\pm$ 0.002 &  0.513 $\pm$ 0.004 &  0.393 $\pm$ 0.004 &   1.5 $\pm$ 0.02 &  0.93 $\pm$ 0.0 \\
\bottomrule
\end{tabular}

\caption{Results for dataset \textit{mg}.}
\end{small}
\end{table*}

\end{landscape}

  \begin{landscape}

\begin{table*}
\begin{small}
\centering
\begin{tabular}{l | lllllllll}
\toprule
{} &             time &                LML &                  KL &                ERR &               CRPS &               RMSE &               ABSE &              NLP &              COV \\
\midrule
fullGP   &  114.8 $\pm$ 4.3 &  -2113.6 $\pm$ 5.6 &       0.0 $\pm$ 0.0 &      0.0 $\pm$ 0.0 &  0.255 $\pm$ 0.005 &   0.471 $\pm$ 0.01 &  0.348 $\pm$ 0.007 &   1.3 $\pm$ 0.04 &   0.95 $\pm$ 0.0 \\ \hline
SGP(50)  &   34.8 $\pm$ 4.8 &  -2319.6 $\pm$ 7.4 &   137.62 $\pm$ 7.41 &  0.259 $\pm$ 0.009 &  0.288 $\pm$ 0.005 &  0.531 $\pm$ 0.012 &  0.395 $\pm$ 0.007 &  1.57 $\pm$ 0.04 &   0.95 $\pm$ 0.0 \\
SGP(100) &   46.6 $\pm$ 6.1 &  -2242.4 $\pm$ 7.5 &   108.14 $\pm$ 6.24 &  0.229 $\pm$ 0.008 &  0.279 $\pm$ 0.005 &  0.514 $\pm$ 0.012 &  0.382 $\pm$ 0.007 &   1.5 $\pm$ 0.04 &   0.95 $\pm$ 0.0 \\
SGP(150) &   56.6 $\pm$ 6.8 &  -2205.9 $\pm$ 6.6 &    90.94 $\pm$ 6.01 &   0.21 $\pm$ 0.009 &  0.275 $\pm$ 0.005 &  0.508 $\pm$ 0.012 &  0.376 $\pm$ 0.007 &  1.47 $\pm$ 0.04 &   0.94 $\pm$ 0.0 \\
minVar   &    7.2 $\pm$ 0.2 &  -2312.6 $\pm$ 6.8 &    63.58 $\pm$ 2.93 &    0.19 $\pm$ 0.01 &  0.272 $\pm$ 0.006 &  0.508 $\pm$ 0.016 &  0.374 $\pm$ 0.008 &  1.41 $\pm$ 0.04 &   0.95 $\pm$ 0.0 \\
GPoE     &    7.2 $\pm$ 0.2 &  -2312.6 $\pm$ 6.8 &    98.01 $\pm$ 3.06 &    0.2 $\pm$ 0.013 &  0.279 $\pm$ 0.006 &   0.515 $\pm$ 0.02 &  0.378 $\pm$ 0.008 &  1.49 $\pm$ 0.03 &   0.97 $\pm$ 0.0 \\
BCM      &    7.2 $\pm$ 0.2 &  -2312.6 $\pm$ 6.8 &   222.78 $\pm$ 4.12 &    0.2 $\pm$ 0.008 &   0.28 $\pm$ 0.007 &  0.511 $\pm$ 0.016 &   0.38 $\pm$ 0.008 &   1.75 $\pm$ 0.1 &  0.87 $\pm$ 0.01 \\
RBCM     &    7.2 $\pm$ 0.2 &  -2312.6 $\pm$ 6.8 &  635.61 $\pm$ 21.61 &  0.194 $\pm$ 0.011 &  0.285 $\pm$ 0.007 &  0.513 $\pm$ 0.018 &  0.378 $\pm$ 0.008 &  2.54 $\pm$ 0.18 &  0.77 $\pm$ 0.01 \\
GRBCM    &    6.5 $\pm$ 0.2 &  -2397.3 $\pm$ 6.2 &   105.64 $\pm$ 5.13 &   0.24 $\pm$ 0.008 &  0.284 $\pm$ 0.005 &  0.525 $\pm$ 0.012 &  0.391 $\pm$ 0.007 &   1.5 $\pm$ 0.04 &  0.95 $\pm$ 0.01 \\
\textbf{CPoE(1) } &    7.8 $\pm$ 0.2 &  -2316.1 $\pm$ 6.8 &    62.99 $\pm$ 2.94 &  0.186 $\pm$ 0.011 &  0.272 $\pm$ 0.006 &  0.507 $\pm$ 0.018 &  0.372 $\pm$ 0.008 &  1.41 $\pm$ 0.04 &   0.96 $\pm$ 0.0 \\
\textbf{CPoE(2)}  &   10.6 $\pm$ 0.2 &  -2164.9 $\pm$ 6.7 &    36.45 $\pm$ 3.02 &  0.142 $\pm$ 0.011 &  0.264 $\pm$ 0.005 &  0.491 $\pm$ 0.015 &  \textbf{0.361 $\pm$ 0.008} & \textbf{ 1.36 $\pm$ 0.04} &   0.95 $\pm$ 0.0 \\
\textbf{CPoE(3) } &   12.9 $\pm$ 0.2 &  -2165.9 $\pm$ 6.7 &    36.27 $\pm$ 2.99 &   0.141 $\pm$ 0.01 & \textbf{ 0.263 $\pm$ 0.005 }&   0.49 $\pm$ 0.014 & \textbf{ 0.361 $\pm$ 0.008 }&  \textbf{1.36 $\pm$ 0.04} &   0.95 $\pm$ 0.0 \\
\textbf{CPoE(4)}  &   14.9 $\pm$ 0.2 &  -2166.2 $\pm$ 6.7 &    \textbf{ 36.03 $\pm$ 3.0} &    \textbf{0.14 $\pm$ 0.01 }&  \textbf{0.263 $\pm$ 0.005} & \textbf{ 0.489 $\pm$ 0.014} &  \textbf{0.361 $\pm$ 0.008 }&  \textbf{1.36 $\pm$ 0.04} &   0.95 $\pm$ 0.0 \\
\bottomrule
\end{tabular}

\caption{Results for dataset \textit{space}.}
\end{small}
\end{table*}

\begin{table*}
\begin{small}
\centering
\begin{tabular}{l | lllllllll}
\toprule
{} &              time &                LML &                 KL &              ERR &               CRPS &               RMSE &               ABSE &              NLP &              COV \\
\midrule
fullGP   &  237.9 $\pm$ 12.2 &  -3722.3 $\pm$ 7.4 &      0.0 $\pm$ 0.0 &    0.0 $\pm$ 0.0 &   0.34 $\pm$ 0.005 &  0.635 $\pm$ 0.012 &  0.459 $\pm$ 0.006 &  1.92 $\pm$ 0.04 &   0.94 $\pm$ 0.0 \\
\hline
SGP(20)  &    21.9 $\pm$ 2.5 &  -3785.3 $\pm$ 5.8 &     27.8 $\pm$ 4.1 &  0.15 $\pm$ 0.01 &  0.343 $\pm$ 0.004 &  0.635 $\pm$ 0.011 &  0.463 $\pm$ 0.005 &  1.93 $\pm$ 0.03 &   0.95 $\pm$ 0.0 \\
SGP(50)  &    26.4 $\pm$ 3.6 &  -3758.7 $\pm$ 7.6 &     22.4 $\pm$ 3.9 &  0.14 $\pm$ 0.01 &  0.342 $\pm$ 0.004 &  0.633 $\pm$ 0.011 &  0.461 $\pm$ 0.006 &  1.93 $\pm$ 0.03 &   0.94 $\pm$ 0.0 \\
SGP(100) &    58.9 $\pm$ 7.0 &  -3746.9 $\pm$ 7.4 &     15.6 $\pm$ 3.5 &  0.11 $\pm$ 0.01 &  \textbf{ 0.34 $\pm$ 0.005} & \textbf{ 0.631 $\pm$ 0.012} &  \textbf{0.457 $\pm$ 0.006} &  1.92 $\pm$ 0.04 &   0.94 $\pm$ 0.0 \\
minVar   &     6.4 $\pm$ 0.4 &  -3847.3 $\pm$ 7.2 &     25.1 $\pm$ 1.5 &   0.15 $\pm$ 0.0 &  0.346 $\pm$ 0.005 &  0.647 $\pm$ 0.013 &  0.466 $\pm$ 0.006 &  1.94 $\pm$ 0.04 &   0.94 $\pm$ 0.0 \\
GPoE     &     6.3 $\pm$ 0.4 &  -3847.3 $\pm$ 7.2 &     50.3 $\pm$ 1.0 &   0.19 $\pm$ 0.0 &  0.353 $\pm$ 0.004 &  0.652 $\pm$ 0.011 &  0.478 $\pm$ 0.006 &  1.99 $\pm$ 0.02 &   0.96 $\pm$ 0.0 \\
BCM      &     6.3 $\pm$ 0.3 &  -3847.3 $\pm$ 7.2 &  1838.2 $\pm$ 46.8 &   0.16 $\pm$ 0.0 &  0.373 $\pm$ 0.006 &  0.642 $\pm$ 0.011 &  0.473 $\pm$ 0.006 &  5.33 $\pm$ 0.24 &  0.67 $\pm$ 0.01 \\
RBCM     &     6.3 $\pm$ 0.3 &  -3847.3 $\pm$ 7.2 &  1147.4 $\pm$ 64.8 &   0.12 $\pm$ 0.0 &  0.362 $\pm$ 0.006 &  0.638 $\pm$ 0.012 &  0.466 $\pm$ 0.006 &  4.01 $\pm$ 0.21 &  0.73 $\pm$ 0.01 \\
GRBCM    &     7.6 $\pm$ 0.4 &  -3864.0 $\pm$ 7.6 &     36.4 $\pm$ 1.9 &   0.18 $\pm$ 0.0 &  0.353 $\pm$ 0.004 &  0.661 $\pm$ 0.011 &  0.477 $\pm$ 0.005 &  1.98 $\pm$ 0.03 &   0.94 $\pm$ 0.0 \\
\textbf{CPoE(1)}  &     6.4 $\pm$ 0.4 &  -3848.6 $\pm$ 7.3 &     16.8 $\pm$ 0.6 &   0.12 $\pm$ 0.0 &  0.342 $\pm$ 0.004 &  0.638 $\pm$ 0.012 &  0.463 $\pm$ 0.005 &  1.92 $\pm$ 0.03 &   0.95 $\pm$ 0.0 \\
\textbf{CPoE(2) } &     7.5 $\pm$ 0.3 &  -3737.3 $\pm$ 7.0 &      8.1 $\pm$ 0.5 &   0.08 $\pm$ 0.0 &  0.341 $\pm$ 0.005 &  0.636 $\pm$ 0.012 &  0.463 $\pm$ 0.006 &  1.92 $\pm$ 0.04 &   0.94 $\pm$ 0.0 \\
\textbf{CPoE(3)}  &     9.3 $\pm$ 0.5 &  -3736.5 $\pm$ 7.2 &      6.2 $\pm$ 0.6 &   0.07 $\pm$ 0.0 &  0.341 $\pm$ 0.005 &  0.636 $\pm$ 0.012 &  0.461 $\pm$ 0.006 &  1.92 $\pm$ 0.04 &   0.94 $\pm$ 0.0 \\
\textbf{CPoE(4) } &    10.4 $\pm$ 0.3 &  -3733.7 $\pm$ 7.0 &     \textbf{ 4.7 $\pm$ 0.5 }&   \textbf{0.06 $\pm$ 0.0} &   \textbf{0.34 $\pm$ 0.005} &  0.635 $\pm$ 0.012 &   0.46 $\pm$ 0.006 &  \textbf{1.91 $\pm$ 0.04 }&   0.94 $\pm$ 0.0 \\
\bottomrule
\end{tabular}

\caption{Results for dataset \textit{abalone}.}
\end{small}
\end{table*}

\end{landscape}

\begin{landscape}

\begin{table*}
\begin{small}
\centering
\begin{tabular}{l | lllllllll}
\toprule
{} &             time &                 LML &                 KL &             ERR &               CRPS &               RMSE &               ABSE &              NLP &             COV \\
\midrule
fullGP   &  161.5 $\pm$ 3.6 &   -1232.1 $\pm$ 7.4 &      0.0 $\pm$ 0.0 &   0.0 $\pm$ 0.0 &  0.148 $\pm$ 0.001 &  0.267 $\pm$ 0.001 &  0.207 $\pm$ 0.001 &  0.17 $\pm$ 0.01 &  0.94 $\pm$ 0.0 \\ \hline
SGP(100) &   42.2 $\pm$ 6.5 &  -4033.6 $\pm$ 27.1 &    603.7 $\pm$ 9.4 &  0.4 $\pm$ 0.01 &  0.265 $\pm$ 0.003 &  0.476 $\pm$ 0.005 &  0.369 $\pm$ 0.004 &  1.35 $\pm$ 0.02 &  0.96 $\pm$ 0.0 \\
SGP(200) &   49.8 $\pm$ 3.3 &  -3141.3 $\pm$ 17.8 &    408.4 $\pm$ 3.7 &  0.29 $\pm$ 0.0 &  0.218 $\pm$ 0.001 &  0.392 $\pm$ 0.001 &  0.303 $\pm$ 0.001 &   0.96 $\pm$ 0.0 &  0.96 $\pm$ 0.0 \\
SGP(300) &   54.8 $\pm$ 2.2 &  -2732.8 $\pm$ 13.5 &    323.1 $\pm$ 5.0 &  0.25 $\pm$ 0.0 &  0.201 $\pm$ 0.001 &  0.363 $\pm$ 0.001 &  0.281 $\pm$ 0.001 &   0.8 $\pm$ 0.01 &  0.96 $\pm$ 0.0 \\
minVar   &    9.3 $\pm$ 0.2 &   -2820.5 $\pm$ 9.0 &    211.0 $\pm$ 2.3 &   0.2 $\pm$ 0.0 &  0.183 $\pm$ 0.001 &  0.333 $\pm$ 0.001 &  0.256 $\pm$ 0.001 &  0.59 $\pm$ 0.01 &  0.94 $\pm$ 0.0 \\
GPoE     &    9.4 $\pm$ 0.1 &   -2820.5 $\pm$ 9.0 &    342.3 $\pm$ 2.6 &  0.23 $\pm$ 0.0 &  0.202 $\pm$ 0.001 &  0.354 $\pm$ 0.002 &  0.278 $\pm$ 0.002 &  0.84 $\pm$ 0.01 &  0.99 $\pm$ 0.0 \\
BCM      &    9.4 $\pm$ 0.1 &   -2820.5 $\pm$ 9.0 &  1629.2 $\pm$ 24.7 &  0.25 $\pm$ 0.0 &  0.218 $\pm$ 0.002 &  0.367 $\pm$ 0.002 &  0.278 $\pm$ 0.002 &  3.45 $\pm$ 0.07 &  0.64 $\pm$ 0.0 \\
RBCM     &    9.4 $\pm$ 0.2 &   -2820.5 $\pm$ 9.0 &   939.3 $\pm$ 17.4 &   0.2 $\pm$ 0.0 &  0.193 $\pm$ 0.001 &  0.331 $\pm$ 0.002 &  0.253 $\pm$ 0.001 &  2.06 $\pm$ 0.05 &  0.71 $\pm$ 0.0 \\
GRBCM    &   11.9 $\pm$ 0.2 &   -2981.3 $\pm$ 9.6 &    129.8 $\pm$ 3.0 &  0.14 $\pm$ 0.0 &  0.168 $\pm$ 0.001 &  0.303 $\pm$ 0.001 &  0.235 $\pm$ 0.001 &  0.43 $\pm$ 0.01 &  0.94 $\pm$ 0.0 \\
\textbf{CPoE(1)}  &    9.2 $\pm$ 0.1 &   -2822.7 $\pm$ 8.9 &    152.4 $\pm$ 1.7 &  0.15 $\pm$ 0.0 &   0.17 $\pm$ 0.001 &  0.307 $\pm$ 0.001 &  0.237 $\pm$ 0.001 &   0.46 $\pm$ 0.0 &  0.97 $\pm$ 0.0 \\
\textbf{CPoE(2) } &   12.9 $\pm$ 0.1 &  -1811.2 $\pm$ 11.1 &     79.9 $\pm$ 1.3 &  0.11 $\pm$ 0.0 &  0.161 $\pm$ 0.001 &   0.29 $\pm$ 0.001 &  0.225 $\pm$ 0.001 &  0.33 $\pm$ 0.01 &  0.95 $\pm$ 0.0 \\
\textbf{CPoE(3) } &   19.8 $\pm$ 0.3 &   -1466.0 $\pm$ 9.9 &     46.9 $\pm$ 1.0 &  0.09 $\pm$ 0.0 &  0.155 $\pm$ 0.001 &  0.279 $\pm$ 0.001 &  0.217 $\pm$ 0.001 &  0.26 $\pm$ 0.01 &  0.95 $\pm$ 0.0 \\
\textbf{CPoE(4) } &   27.8 $\pm$ 0.2 &   -1363.8 $\pm$ 9.2 &    \textbf{ 32.8 $\pm$ 1.0 }&  \textbf{0.07 $\pm$ 0.0} &  \textbf{0.153 $\pm$ 0.001} & \textbf{ 0.276 $\pm$ 0.001 }&  \textbf{0.215 $\pm$ 0.001 }&  \textbf{0.24 $\pm$ 0.01 }&  0.94 $\pm$ 0.0 \\
\bottomrule
\end{tabular}

\caption{Results for dataset \textit{kin}.}
\end{small}
\end{table*}

\begin{table*}
\begin{small}
\centering
\begin{tabular}{l | lllllll}
\toprule
{} &             time &                 LML &               CRPS &               RMSE &               ABSE &              NLP &             COV \\
\midrule
SGP(250)  &   77.7 $\pm$ 0.4 &  -4163.9 $\pm$ 23.7 &  0.207 $\pm$ 0.002 &  0.366 $\pm$ 0.004 &  0.282 $\pm$ 0.002 &  0.93 $\pm$ 0.01 &  0.98 $\pm$ 0.0 \\
SGP(500)  &  112.1 $\pm$ 1.2 &  -3242.2 $\pm$ 12.6 &  0.183 $\pm$ 0.001 &  0.324 $\pm$ 0.002 &  0.252 $\pm$ 0.001 &  0.67 $\pm$ 0.01 &  0.98 $\pm$ 0.0 \\
SGP(1000) &  244.1 $\pm$ 2.9 &   -2534.7 $\pm$ 9.0 &  0.166 $\pm$ 0.001 &  0.294 $\pm$ 0.002 &   0.23 $\pm$ 0.001 &  0.46 $\pm$ 0.01 &  0.98 $\pm$ 0.0 \\
minVar    &   14.4 $\pm$ 0.5 &   -3388.8 $\pm$ 7.9 &  0.173 $\pm$ 0.002 &  0.314 $\pm$ 0.004 &  0.242 $\pm$ 0.002 &  0.48 $\pm$ 0.02 &  0.94 $\pm$ 0.0 \\
GPoE      &   14.4 $\pm$ 0.5 &   -3388.8 $\pm$ 7.9 &  0.193 $\pm$ 0.001 &   0.34 $\pm$ 0.003 &  0.267 $\pm$ 0.002 &  0.76 $\pm$ 0.01 &  0.99 $\pm$ 0.0 \\
BCM       &   14.4 $\pm$ 0.5 &   -3388.8 $\pm$ 7.9 &   0.21 $\pm$ 0.001 &   0.35 $\pm$ 0.003 &  0.266 $\pm$ 0.002 &    3.6 $\pm$ 0.1 &  0.63 $\pm$ 0.0 \\
RBCM      &   14.4 $\pm$ 0.5 &   -3388.8 $\pm$ 7.9 &  0.188 $\pm$ 0.001 &  0.318 $\pm$ 0.003 &  0.244 $\pm$ 0.002 &  2.39 $\pm$ 0.09 &  0.69 $\pm$ 0.0 \\
GRBCM     &   16.5 $\pm$ 0.4 &   -3388.8 $\pm$ 7.9 &  0.164 $\pm$ 0.001 &  0.294 $\pm$ 0.003 &  0.229 $\pm$ 0.002 &  0.37 $\pm$ 0.02 &  0.94 $\pm$ 0.0 \\
\textbf{CPoE(1)}   &   13.8 $\pm$ 0.2 &   -3393.9 $\pm$ 8.0 &  0.163 $\pm$ 0.001 &  0.292 $\pm$ 0.003 &  0.226 $\pm$ 0.002 &  0.38 $\pm$ 0.01 &  0.97 $\pm$ 0.0 \\
\textbf{CPoE(2)}   &   18.9 $\pm$ 0.3 &  -2076.6 $\pm$ 12.9 &  0.155 $\pm$ 0.001 &  0.278 $\pm$ 0.002 &  0.217 $\pm$ 0.001 &  0.27 $\pm$ 0.01 &  0.95 $\pm$ 0.0 \\
\textbf{CPoE(3) }  &   31.7 $\pm$ 0.6 &   -1655.2 $\pm$ 8.7 &  \textbf{0.151 $\pm$ 0.001} &   \textbf{0.27 $\pm$ 0.002} & \textbf{ 0.211 $\pm$ 0.001 }&  \textbf{0.21 $\pm$ 0.01 }&  0.95 $\pm$ 0.0 \\
\bottomrule
\end{tabular}

\caption{Results for dataset \textit{kin2} for the stochastic versions.}
\end{small}
\end{table*}

\end{landscape}

\begin{landscape}

\begin{table*}
\begin{small}
\centering
\begin{tabular}{l | lllllll}
\toprule
{} &             time &                 LML &               CRPS &               RMSE &               ABSE &              NLP &             COV \\
\midrule
SGP(250)  &   70.9 $\pm$ 3.7 &  -3905.6 $\pm$ 23.3 &  0.207 $\pm$ 0.002 &  0.373 $\pm$ 0.004 &  0.287 $\pm$ 0.002 &  0.85 $\pm$ 0.02 &  0.96 $\pm$ 0.0 \\
SGP(500)  &   86.1 $\pm$ 1.8 &  -2968.6 $\pm$ 11.7 &  0.181 $\pm$ 0.001 &  0.325 $\pm$ 0.003 &  0.252 $\pm$ 0.001 &  0.57 $\pm$ 0.01 &  0.96 $\pm$ 0.0 \\
SGP(1000) &  143.6 $\pm$ 3.6 &   -2277.2 $\pm$ 8.6 &  0.162 $\pm$ 0.001 &  0.292 $\pm$ 0.002 &  0.225 $\pm$ 0.001 &  0.36 $\pm$ 0.01 &  0.96 $\pm$ 0.0 \\
minVar    &   13.8 $\pm$ 0.2 &   -3384.5 $\pm$ 7.8 &  0.173 $\pm$ 0.002 &  0.314 $\pm$ 0.004 &  0.241 $\pm$ 0.002 &  0.48 $\pm$ 0.02 &  0.94 $\pm$ 0.0 \\
GPoE      &   13.8 $\pm$ 0.2 &   -3384.5 $\pm$ 7.8 &  0.193 $\pm$ 0.001 &   0.34 $\pm$ 0.002 &  0.267 $\pm$ 0.002 &  0.75 $\pm$ 0.01 &  0.99 $\pm$ 0.0 \\
BCM       &   13.8 $\pm$ 0.2 &   -3384.5 $\pm$ 7.8 &  0.209 $\pm$ 0.001 &   0.35 $\pm$ 0.003 &  0.266 $\pm$ 0.002 &  3.63 $\pm$ 0.07 &  0.63 $\pm$ 0.0 \\
RBCM      &   13.8 $\pm$ 0.2 &   -3384.5 $\pm$ 7.8 &  0.187 $\pm$ 0.001 &  0.317 $\pm$ 0.003 &  0.243 $\pm$ 0.001 &  2.38 $\pm$ 0.06 &  0.69 $\pm$ 0.0 \\
GRBCM     &   18.8 $\pm$ 0.4 &   -3608.7 $\pm$ 8.4 &  0.164 $\pm$ 0.001 &  0.294 $\pm$ 0.002 &  0.229 $\pm$ 0.002 &  0.38 $\pm$ 0.02 &  0.94 $\pm$ 0.0 \\
\textbf{CPoE(1) }  &   16.2 $\pm$ 0.8 &   -3389.8 $\pm$ 8.0 &  0.162 $\pm$ 0.001 &  0.292 $\pm$ 0.003 &  0.225 $\pm$ 0.002 &  0.37 $\pm$ 0.01 &  0.97 $\pm$ 0.0 \\
\textbf{CPoE(2) }  &   21.5 $\pm$ 0.7 &  -2071.4 $\pm$ 13.0 &  0.155 $\pm$ 0.001 &  0.278 $\pm$ 0.002 &  0.217 $\pm$ 0.001 &  0.26 $\pm$ 0.01 &  0.95 $\pm$ 0.0 \\
\textbf{CPoE(3) }  &   34.3 $\pm$ 0.9 &   -1650.7 $\pm$ 8.3 &   \textbf{0.15 $\pm$ 0.001} &   \textbf{0.27 $\pm$ 0.002 }&  \textbf{0.211 $\pm$ 0.001 }&  \textbf{0.21 $\pm$ 0.01} &  0.94 $\pm$ 0.0 \\
\bottomrule
\end{tabular}

\caption{Results for dataset \textit{kin2} for the deterministic batch version.}
\end{small}
\end{table*}

\begin{table}
\begin{small}
\centering
\begin{tabular}{l | lllllll}
\toprule
{} &             time &                  LML &               CRPS &               RMSE &               ABSE &              NLP &              COV \\
\midrule
SGP(250)  &  248.6 $\pm$ 0.6 &  -15182.0 $\pm$ 35.7 &  0.254 $\pm$ 0.003 &   0.48 $\pm$ 0.009 &  0.335 $\pm$ 0.004 &  1.42 $\pm$ 0.03 &   0.95 $\pm$ 0.0 \\
SGP(500)  &  346.9 $\pm$ 3.4 &  -15074.6 $\pm$ 37.2 &  0.253 $\pm$ 0.003 &  0.478 $\pm$ 0.009 &  0.333 $\pm$ 0.004 &  1.41 $\pm$ 0.03 &   0.95 $\pm$ 0.0 \\
SGP(1000) &  727.6 $\pm$ 3.5 &  -14961.2 $\pm$ 31.4 &  0.252 $\pm$ 0.003 &  \textbf{0.476 $\pm$ 0.009 }&  0.332 $\pm$ 0.004 &   1.4 $\pm$ 0.03 &   0.95 $\pm$ 0.0 \\
minVar    &   28.2 $\pm$ 1.0 &  -15387.4 $\pm$ 17.5 &  0.257 $\pm$ 0.003 &  0.491 $\pm$ 0.009 &  0.337 $\pm$ 0.005 &  1.42 $\pm$ 0.04 &   0.94 $\pm$ 0.0 \\
GPoE      &   28.3 $\pm$ 1.0 &  -15387.4 $\pm$ 17.5 &  0.289 $\pm$ 0.003 &  0.534 $\pm$ 0.009 &  0.371 $\pm$ 0.004 &  1.64 $\pm$ 0.02 &   0.96 $\pm$ 0.0 \\
BCM       &   28.5 $\pm$ 0.9 &  -15387.4 $\pm$ 17.5 &  0.321 $\pm$ 0.004 &   0.536 $\pm$ 0.01 &  0.373 $\pm$ 0.004 &  20.72 $\pm$ 1.0 &   0.45 $\pm$ 0.0 \\
RBCM      &   28.5 $\pm$ 0.9 &  -15387.4 $\pm$ 17.5 &  0.303 $\pm$ 0.005 &   0.515 $\pm$ 0.01 &  0.358 $\pm$ 0.004 &  15.98 $\pm$ 0.9 &  0.51 $\pm$ 0.01 \\
GRBCM     &   33.5 $\pm$ 1.2 &  -15387.4 $\pm$ 17.5 &  0.262 $\pm$ 0.003 &  0.499 $\pm$ 0.009 &  0.346 $\pm$ 0.004 &  1.44 $\pm$ 0.03 &   0.94 $\pm$ 0.0 \\
\textbf{CPoE(1)}   &   24.5 $\pm$ 0.1 &  -15404.2 $\pm$ 17.8 &  0.259 $\pm$ 0.004 &   0.492 $\pm$ 0.01 &  0.335 $\pm$ 0.005 &  1.43 $\pm$ 0.04 &   0.95 $\pm$ 0.0 \\
\textbf{CPoE(2)}   &   33.4 $\pm$ 0.2 &  -13645.5 $\pm$ 19.8 &  0.251 $\pm$ 0.003 &  0.479 $\pm$ 0.009 &  0.328 $\pm$ 0.004 &  1.36 $\pm$ 0.04 &   0.94 $\pm$ 0.0 \\
\textbf{CPoE(3)}   &   52.0 $\pm$ 0.5 &  -13483.2 $\pm$ 15.6 &  \textbf{0.249 $\pm$ 0.004} &   \textbf{0.476 $\pm$ 0.01} &  \textbf{0.324 $\pm$ 0.004} &  \textbf{1.34 $\pm$ 0.04 }&   0.94 $\pm$ 0.0 \\
\bottomrule
\end{tabular}

\caption{Results for dataset \textit{cadata}.}
\end{small}
\end{table}

\end{landscape}

\begin{landscape}

\begin{table}
\begin{small}
\centering
\begin{tabular}{l | lllllll}
\toprule
{} &              time &                   LML &                 CRPS &                 RMSE &               ABSE &               NLP &              COV \\
\midrule
SGP(250)  &   473.4 $\pm$ 1.0 &     9370.3 $\pm$ 60.7 &  0.0746 $\pm$ 0.0005 &  0.1407 $\pm$ 0.0008 &  0.097 $\pm$ 0.001 &  -0.39 $\pm$ 0.01 &   0.95 $\pm$ 0.0 \\
SGP(500)  &   730.1 $\pm$ 1.1 &    12112.0 $\pm$ 68.2 &  0.0695 $\pm$ 0.0003 &   0.1304 $\pm$ 0.001 &   0.09 $\pm$ 0.001 &  -0.49 $\pm$ 0.01 &   0.95 $\pm$ 0.0 \\
SGP(1000) &  1718.5 $\pm$ 1.8 &    16034.0 $\pm$ 91.6 &  0.0628 $\pm$ 0.0003 &  0.1172 $\pm$ 0.0009 &    0.081 $\pm$ 0.0 &  -0.64 $\pm$ 0.01 &   0.96 $\pm$ 0.0 \\
minVar    &   71.3 $\pm$ 23.1 &    27128.2 $\pm$ 20.2 &  0.0516 $\pm$ 0.0008 &  0.1024 $\pm$ 0.0034 &  \textbf{0.067 $\pm$ 0.001} & \textbf{ -1.88 $\pm$ 0.04} &   0.93 $\pm$ 0.0 \\
GPoE      &   71.4 $\pm$ 23.2 &    27128.2 $\pm$ 20.2 &  0.0862 $\pm$ 0.0004 &  0.1322 $\pm$ 0.0013 &  0.096 $\pm$ 0.001 &  -0.57 $\pm$ 0.01 &    1.0 $\pm$ 0.0 \\
BCM       &   71.5 $\pm$ 23.2 &    27128.2 $\pm$ 20.2 &    0.095 $\pm$ 0.001 &   0.1544 $\pm$ 0.001 &  0.115 $\pm$ 0.001 &    7.86 $\pm$ 0.3 &  0.48 $\pm$ 0.01 \\
RBCM      &   71.6 $\pm$ 23.2 &    27128.2 $\pm$ 20.2 &  0.0726 $\pm$ 0.0009 &  0.1196 $\pm$ 0.0013 &  0.086 $\pm$ 0.001 &  11.45 $\pm$ 0.47 &   0.5 $\pm$ 0.01 \\
GRBCM     &   84.6 $\pm$ 23.0 &    27128.2 $\pm$ 20.2 &    0.06 $\pm$ 0.0007 &   0.1102 $\pm$ 0.001 &  0.079 $\pm$ 0.001 &  -0.52 $\pm$ 0.08 &  0.79 $\pm$ 0.01 \\
\textbf{CPoE(1)}   &    45.4 $\pm$ 0.2 &  -41213.2 $\pm$ 883.2 &  0.0516 $\pm$ 0.0005 &  0.0998 $\pm$ 0.0019 &  0.067 $\pm$ 0.001 &  -1.86 $\pm$ 0.02 &   0.96 $\pm$ 0.0 \\
\textbf{CPoE(2)}   &    67.3 $\pm$ 0.4 &  -37867.5 $\pm$ 911.8 &  0.0509 $\pm$ 0.0006 &  0.0977 $\pm$ 0.0015 &  0.067 $\pm$ 0.001 &   -1.8 $\pm$ 0.02 &   0.93 $\pm$ 0.0 \\
\textbf{CPoE(3) }  &   134.3 $\pm$ 1.2 &  -37204.6 $\pm$ 949.1 & \textbf{ 0.0507 $\pm$ 0.0005} &  \textbf{0.0975 $\pm$ 0.0011 }&  \textbf{0.067 $\pm$ 0.001 }&  -1.78 $\pm$ 0.02 &   0.92 $\pm$ 0.0 \\
\bottomrule
\end{tabular}

\caption{Results for dataset \textit{sarcos}.}
\end{small}
\end{table}

\begin{table}
\begin{small}
\centering
\begin{tabular}{ l | lllllll}
\toprule
{} &              time &                  LML &               CRPS &               RMSE &               ABSE &              NLP &              COV \\
\midrule
SGP(250)  &   443.2 $\pm$ 2.1 &  -53395.2 $\pm$ 80.2 &  0.334 $\pm$ 0.004 &   0.59 $\pm$ 0.008 &  0.475 $\pm$ 0.007 &  1.77 $\pm$ 0.02 &   0.96 $\pm$ 0.0 \\
SGP(500)  &   632.9 $\pm$ 2.7 &  -52988.7 $\pm$ 58.9 &  0.329 $\pm$ 0.005 &  0.582 $\pm$ 0.008 &  0.467 $\pm$ 0.007 &  1.75 $\pm$ 0.02 &   0.96 $\pm$ 0.0 \\
SGP(1000) &  1362.5 $\pm$ 4.8 &  -52592.1 $\pm$ 46.9 &  0.325 $\pm$ 0.005 &  \textbf{0.575 $\pm$ 0.008 }&  0.459 $\pm$ 0.007 &  1.74 $\pm$ 0.02 &   0.96 $\pm$ 0.0 \\
minVar    &    45.8 $\pm$ 1.0 &  -39976.0 $\pm$ 22.6 &  0.294 $\pm$ 0.003 &  0.607 $\pm$ 0.006 &  0.387 $\pm$ 0.003 &   1.4 $\pm$ 0.03 &   0.93 $\pm$ 0.0 \\
GPoE      &    45.6 $\pm$ 0.8 &  -39976.0 $\pm$ 22.6 &  0.302 $\pm$ 0.003 &    0.6 $\pm$ 0.006 &  0.409 $\pm$ 0.005 &  1.43 $\pm$ 0.02 &   0.97 $\pm$ 0.0 \\
BCM       &    45.7 $\pm$ 0.9 &  -39976.0 $\pm$ 22.6 &  0.316 $\pm$ 0.005 &  0.615 $\pm$ 0.009 &  0.416 $\pm$ 0.007 &   2.47 $\pm$ 0.1 &  0.82 $\pm$ 0.01 \\
RBCM      &    45.7 $\pm$ 0.9 &  -39976.0 $\pm$ 22.6 &  0.312 $\pm$ 0.004 &  0.647 $\pm$ 0.008 &  0.425 $\pm$ 0.006 &  1.61 $\pm$ 0.05 &  0.91 $\pm$ 0.01 \\
GRBCM     &    59.4 $\pm$ 1.1 &  -39976.0 $\pm$ 22.6 &   0.31 $\pm$ 0.004 &  0.642 $\pm$ 0.008 &  0.421 $\pm$ 0.005 &   1.5 $\pm$ 0.04 &  0.92 $\pm$ 0.01 \\
\textbf{CPoE(1)}   &    45.1 $\pm$ 0.3 &  -40075.2 $\pm$ 22.1 &  0.289 $\pm$ 0.003 &  0.596 $\pm$ 0.006 &   0.38 $\pm$ 0.004 &  \textbf{1.35 $\pm$ 0.03} &   0.94 $\pm$ 0.0 \\
\textbf{CPoE(2) }  &    70.3 $\pm$ 0.6 &  -39571.2 $\pm$ 65.5 &  0.287 $\pm$ 0.004 &  0.589 $\pm$ 0.007 &   0.38 $\pm$ 0.005 &  1.36 $\pm$ 0.03 &   0.93 $\pm$ 0.0 \\
\textbf{CPoE(3) }  &   123.8 $\pm$ 1.4 &  -39439.5 $\pm$ 98.8 &  \textbf{0.282 $\pm$ 0.004} &  \textbf{0.575 $\pm$ 0.008 }&  \textbf{0.372 $\pm$ 0.006 }&  1.37 $\pm$ 0.04 &  0.92 $\pm$ 0.01 \\
\bottomrule
\end{tabular}

\caption{Results for dataset \textit{casp}.}
\end{small}
\end{table}
\end{landscape}

%
%

\end{document}